\documentclass[twoside,11pt]{article}

\usepackage{amsmath}
\usepackage{amssymb}
\usepackage{jmlr2e}
\usepackage[ruled,vlined]{algorithm2e}
\usepackage{multirow}

\newtheorem{theorem*}{Theorem}

\ShortHeadings{Context Aware Machine Learning (CAML)}{Yun Zeng}
\firstpageno{1}

\begin{document}

\title{Context Aware Machine Learning}

\author{\name Yun Zeng \email xzeng@google.com \\
       \addr 1600 Amphitheatre Parkway, Mountain View\\
       CA 94043, United States}

\editor{}

\maketitle

\begin{abstract}
We propose a principle for exploring \emph{context} in machine learning models. 
Starting with a simple assumption that each observation (random variables) may or may not depend on its context (conditional variables), a conditional probability distribution is decomposed into two parts: \emph{context-free} and \emph{context-sensitive}. 
Then by employing the log-linear word production model for relating random variables to their embedding space representation and making use of the convexity of natural exponential function, we show that the embedding of an observation can also be decomposed into a weighted sum of two vectors, representing its context-free and context-sensitive parts, respectively. 
This simple treatment of context provides a unified view of many existing deep learning models, leading to revisions of these models able to achieve significant performance boost. 
Specifically, our upgraded version of a recent sentence embedding model~\citep{sem_arora2017} not only outperforms the original one by a large margin, but also leads to a new, principled approach for compositing the embeddings of bag-of-words features, as well as a new architecture for modeling attention in deep neural networks. 
More surprisingly, our new principle provides a novel understanding of the gates and equations defined by the long short term memory (LSTM) model, which also leads to a new model that is able to converge significantly faster and achieve much lower prediction errors. 
Furthermore, our principle also inspires a new type of generic neural network layer that better resembles real biological neurons than the traditional linear mapping plus nonlinear activation based architecture. Its multi-layer extension provides a new principle for deep neural networks which subsumes residual network (ResNet) as its special case, and its extension to convolutional neutral network model accounts for irrelevant input (\emph{e.g.}, background in an image) in addition to filtering.
Our models are validated through a series of benchmark datasets and we show that in many cases, simply replacing existing layers with our context-aware counterparts is sufficient to significantly improve the results.
\end{abstract}


\section{Introduction}

\small{``A brain state can be regarded as a transient equilibrium condition, which contains all aspects of past history that are useful for future use. This definition is related but not identical to the equally difficult concept of \emph{context}, which refers to a set of facts or circumstances that surround an event, situation, or object.''}
\\[3pt]
\rightline{{\rm --- Gyo\"rgy Buzs\'aki in \emph{Rhythms of the Brain}}}
\\[3pt]

Most machine learning tasks can be formulated into inferring the likelihood of certain outcome (\emph{e.g.}, labels) from given input or context~\citep{prml_book}. Also most models assume such a likelihood is completely determined by its context, and therefore their relations can be approximated by functions (deterministic or nondeterministic) of the context with various approximate, parameterized forms (\emph{e.g.}, neural networks), which in turn can be learned from the data. 
Nevertheless, such an assumption does not always hold true as real-world data are often full of irrelevant information and/or noises.
Explicitly addressing the relevance of the context to the output has been shown to be very effective and sometimes game-changing (\emph{e.g.}, ~\cite{lstm_original,attention_all_you_need}). 
Most notably, three circuit-like gates are defined in the LSTM model~\citep{lstm_original} to address the relevance of previous state/previous output/current input to current state/current output in each time step. Despite its tremendous success in the past two decades in modeling sequential data~\citep{lstm_space_odyssey}, LSTM's heuristic construction of the neural network keeps inspiring people to search for better architectures or more solid theoretical foundations. 
In this work, we continue with such an effort by showing that not only better architectures exist, the whole context-aware treatment also squarely fits a unified probabilistic framework that allows us to connect with and revise many existing models. 

Our work is primarily inspired by a sentence embedding algorithm proposed by~\cite{sem_arora2017}, where the embedding of a sentence (bag of words) is decomposed into a weighted sum of a \emph{global vector}, computed as the first principal component on the embeddings of all training sentences, and a \emph{local vector} that changes with the sentence input.
Its decomposition is achieved via adding a word frequency term into the word-production model~\citep{emb_mnih2007three}, connected by a weight (a global constant) that is chosen heuristically.
Word frequency can take into account those common words (\emph{e.g.}, ``the", ``and") that may appear anywhere, and~\cite{sem_arora2017} showed that in the embedding space, the impact of those common/uninformative words can be reduced thanks to the new decomposition.
Essentially, that work connects the idea of term frequency-inverse document frequency (TF-IDF) from information retrieval~\citep{info_retrieval_manning} to sentence embedding. 
Despite its promising results, the model is constructed based on a number of heuristic assumptions, such as assuming the weights that connects the global and local vectors be constant, as well as the use of principal component analysis for estimating the global vector.

To find a more solid theoretical ground for the above sentence embedding model, we start by revamping its heuristic assumptions using rigorous probability rules. 
Rather than directly adding a new term, \emph{i.e.}, word frequency, into the word-production model, we introduce an indicator variable to the conditional part of the probability distribution that takes into account the extent to which the observation (\emph{e.g.}, a word) is related to its given context (\emph{e.g.}, a sentence or a topic). This \emph{context-aware} variable allows us to decompose the original conditional probability into two terms: a context-free part and a context-sensitive part, connected by a weighting function (a probability distribution) determined by the input. 
Furthermore, by restricting the probability distribution be within the exponential family and exploiting its convexity property, a new equation is obtained, which we call the \emph{Embedding Decomposition Formula} or \textbf{EDF}.
It decomposes the embedding of an observation into two vectors: context-free and context-sensitive, connected by a weighting function, called the $\mathbf{\chi}$-\textbf{function} that denotes the level of context-freeness for each observation. 
EDF is powerful in its \emph{simplicity} in form, its \emph{universality} in covering all models that can be formulated with a conditional probability distribution and therefore its potential in revolutionizing many existing machine learning problems from this context-aware point of view.

Note that the concept of context-aware treatment for machine learning problems is hardly new. Besides language related models mentioned above, it has also been explored in image models both explicitly (\emph{e.g.}, \cite{image_visual_search_iccv09}) and implicitly (\emph{e.g.}, \cite{image_visual_search_pami12}).
However, our \emph{contribution} here is a new formulation of such a common sense in the embedding space, based on established statistics theory and probability rules, which leads to a new, unified principle that can be applied to improving many existing models. 
Specifically, we revisit and revise the following problems in this paper:

\begin{itemize}
\item \emph{Sentence embedding} (CA-SEM): The basic idea of the original paper by~\cite{sem_arora2017} is to explore global information in a corpus of training sentences, using any already learned word embeddings. Our EDF inspires a new formulation by casting it into an energy minimization problem, which solves for the optimal decomposition of the original word embeddings into a context free part and a context sensitive part, connected by the $\chi$-function that represents the broadness of each word. We show that it can be solved iteratively by a block-coordinate descent algorithm with a closed-form solution in each step. In our experiment, our new approach significantly outperforms the original model in all the benchmarks tasks we tested.

\item \emph{Bag of sparse features embedding} (CA-BSFE): In a neural network with sparse feature input, it is essential to first embed each input feature column to a vector before feeding it to the next layer.
Sometimes the values for a feature column may be represented by bag-of-words, \emph{e.g.}, words in a sentence or topics in a webpage. 
Prior art simply computes the embedding for that feature column by averaging the embeddings of its values. Our framework suggests a new architecture for computing the embeddings of bag-of-sparse-features in neural networks. Its setup automatically learns the importance of each feature values and out-of-vocabulary (OOV) values are automatically taken care of by a context-free vector learned during training. 

\item \emph{Attention model} (CA-ATT): If attention can be considered as sensitivity to its context/input, our framework is then directly applicable to computing the attention vector used in sequence or image models. Different from the CA-BSFE model above that focuses on training a word-to-embedding lookup dictionary, our CA-ATT model focuses on the training of the $\chi$-function (attention) from input context, corresponding to the softmax weighting function in existing attention models.
Our theory implies normalization of the weight is not necessary, since the number of salient objects from the input should vary (\emph{e.g.}, single face versus multiple faces in an image).

\item \emph{Sequence model} (CA-RNN): Our framework is able to provide a new explanation of the LSTM model. 
First of all, a sequence modeling problem can be formulated into inferring the current output and hidden state from previous output and hidden state, as well as current input. Then by applying standard probability rules the problem is divided into the inferences of the current output and hidden state, respectively. For each of the sub-problems, by applying our new framework, we find both the forget and the output gates defined in LSTM can find their correspondences as a $\chi$-function defined in our model.  
This alignment also allows us to unify various existing sequence models and find the ``correct" version that outperforms existing sequence models significantly.  

\item \emph{Generic neural network layer} (CA-NN): Traditional linear mapping of input $+$ nonlinear activation based NN layer fails to capture the change-of-fire-patterns phenomena, known as bifurcation, which is common in real neurons. Our framework partially addresses this limitation by allowing an NN layer to switch between two modes: context-free and context-sensitive, based on the contextual input. We show that the mappings learned by CA-NN layer possess different properties from a normal NN layer.

\item \emph{Convolutional neural network} (CA-CNN): As an application of the CA-NN architecture, it is straightforward to design a new convolutional neural network whose $\chi$-function explicitly classifies each input fragment (\emph{e.g.}, an image patch) into irrelevant (context free) background/noise  or meaningful (context sensitive) foreground. 
Qualitative results on popular datasets such as MNIST illustrate that our model can almost perfectly tell informative foreground from irrelevant background.

\item \emph{Residual network layer} (CA-RES): By carefully modeling CA-NN in the multi-layer case using probability theory, we have discovered a new principle for building deep neural networks, which also subsumes the currently popular residual network (ResNet) as our special case. Our new treatment regards a deep neural network as inter-connected neurons without any hierarchy before training, much like human brain in the early ages, and hierarchical information processing is only formed when sufficient data are learned, controlled by the $\chi$-function in each CA-NN sublayer that turns it on/off when information is passed from its input to output.
\end{itemize}

The rest of this paper is organized as follows. We first establish the theoretical foundation of our new approach in Sec.~\ref{section:math_foundation}. Then we connect our new framework to the problems listed above and also revise them into better forms in Sec.~\ref{section:new_models}. In Sec.~\ref{section:experiment} the performance of our models are evaluated on various datasets and models. 
Related works on diverse fields covered in this paper are reviewed in Sec.~\ref{section:related_work}, after which we conclude our work in Sec.~\ref{section:conclusion}.

\section{Mathematical Formulation}
\label{section:math_foundation}

We start by discussing the close relations among language modeling, exponential family and word embedding models in Sec.~\ref{subsection:exponential_family}.
Then in Sec.~\ref{subsection:context_aware_embedding_model} we propose a novel framework for context-aware treatment of probabilistic models and introduce our EDF.
In Sec.~\ref{subsection:bof_embedding} we discuss its extension to embedding bag-of-features.

\subsection{Language Modeling, Exponential Family and Embedding}
\label{subsection:exponential_family}
Generally speaking, the task of statistical language modeling is to find the distribution of a set of words $P(w_1, \ldots, w_n), w_i\in \mathcal{W}, i = 1, 2, \ldots, n$~\cite{info_retrieval_manning}.
In practice, it is usually more useful to model the distribution of a word $w\in \mathcal{W}$ from its given context $c\in \mathcal{C}$: $P(w|c)$,
where context can be a sequence of words preceding $w$ or any other relevant information that contributes to the production of $w$~\cite{emb_bengio2003neural}.
Note that the same formulation can be applied to any other problems involving probability distributions with discrete variables.

To obtain a paramterized representation of $P(\cdot)$, one usually considers the exponential family that covers a rich set of distributions, \emph{i.e.},  $P(c) \propto \exp(\langle \mathbf{\theta}_1, \mathbf{T}_1(c) \rangle)$ and $P(w, c) \propto \exp(\langle \mathbf{\theta}_2, \mathbf{T}_2(w, c) \rangle)$, where $\mathbf{\theta}_i \in \mathbb{R}^{d}, i \in \{1, 2\}$,  $\mathbf{T}_1:  \mathcal{C}\mapsto \mathbb{R}^{d}$ and $\mathbf{T}_2: \mathcal{W} \times \mathcal{C} \mapsto \mathbb{R}^{d}$ map the input variables to a $d$-dimensional Euclidean space.
By further assuming $\mathbf{T}_2(w, c) = \mathbf{T}_3(w) \odot \mathbf{T}_1 (c)$ where $\mathbf{T}_3:  \mathcal{W}\mapsto \mathbb{R}^{d}$ and $\odot: \mathbb{R}^{d}\times \mathbb{R}^{d}\mapsto \mathbb{R}^{d}$ is the pointwise multiplication, then from $P(w|c) = P(w, c) / P(c)$, we have:
\begin{equation}\label{equ:loglinearmodel}
P(w|c) \propto \exp(\langle \theta_2\odot\mathbf{T}_3(w) - \theta_1, \mathbf{T}_1(c) \rangle) \equiv \exp(\langle \mathbf{W}(w), \mathbf{C}(c) \rangle),
\end{equation}
which is known as the \emph{log-linear word production model}. Here $\mathbf{W}: \mathcal{W}\mapsto \mathbb{R}^{d}$ and $\mathbf{C}: \mathcal{C}\mapsto \mathbb{R}^{d}$ can be considered as the \emph{embeddings} for $w$ and $c$, respectively.
Note that Eq.~\ref{equ:loglinearmodel} is based on very restricted assumption on the forms of exponential family distribution to decouple $w$ and $c$.
One can certainly consider the more general case that  $\mathbf{T}_2(w, c)$ is not factorable, then $\mathbf{W}(\cdot)$ becomes a function on $\mathcal{W} \times \mathcal{C}$ and $\mathbf{C}(\cdot)$ remains the same.
In the rest of this paper, we will denote the embedding of a variable $v$ simply by $\vec{v}$, then Eq.~\ref{equ:loglinearmodel} becomes 
\begin{equation}\label{equ:loglinearmodel_simplified}
P(w|c) \propto \exp(\langle \vec{w}, \vec{c} \rangle).
\end{equation}

Also note that the same model of Eq.~\ref{equ:loglinearmodel_simplified} has been proposed under various interpretations in the past (\emph{e.g.}, ~\cite{exponential_family_harmoniums,emb_mnih2007three,emb_arora2016latent}).
Here we favor the exponential family point of view as it can be easily generalized to constructing probability distributions with more complex forms (\emph{e.g.}, ~\cite{emb_exp_family_embed}), or with more than one conditional variables.
For example, from $P(w|c, s) = P(w, c|s) / P(c|s)$, we may consider a general family of distributions:
\begin{equation}
P(w|c, s) \propto \exp(\langle \vec{w}(s), \vec{c}(s) \rangle),
\end{equation}
where we use $\vec{w}(s)$ and $\vec{c}(s)$ to denote that the embeddings of $w$ and $c$ also depends on the value of $s$. 
We will show that this is very instrumental for our new treatment of many existing models.

\subsection{Context Aware Embedding Model}
\label{subsection:context_aware_embedding_model}
Given the definition of the log-linear model, we are ready to introduce the core concept in this paper: \emph{context free} and its negation, \emph{context sensitive}.

Consider the case of a language model $P(w|c)$ when $w$ is a word and $c$ is a sentence as its context.
When $w$ is a uninformative word like \{``is", ``the"\}, it may appear in any sentence and its frequency is independent of its context (context free).
Therefore we can loosely define $P(w | c, w \text{ is context free}) = \tilde{P}(w)$, meaning it is independent of any context $c$.

Formally, if we introduce an indicator $CF(w)$ to denote if a variable $w$ is context free or context sensitive, \emph{i.e.},
 \begin{equation}
\text{CF}(w) = \begin{cases}
1, &\text{if } w \text{ is context free} \\
0, &\text{if } w \text{ is context sensitive}
\end{cases}
\end{equation}
then we can quantitatively define the concept of context-freeness as follows.
\begin{definition}\textbf{(Context free indicator)}.
A random indicator variable $CF: \mathcal{W} \mapsto \{0, 1\}$ is called \emph{context free indicator} if and only if the following equation holds for $\forall c$
\begin{align}\label{equ:context_free_basic}
P(w|CF(w) = 1, c) &= \tilde{P}(w)
\end{align}
Here $\tilde{P}(w)$ is a probability distribution that is not necessarily the same as $P(w)$ (word frequency).
\end{definition}

Given the flexibility of the exponential family, we can always construct a probability distribution that satisfies the above equation.
For example, any distribution with the form $P_{\theta}(w|CF(w), c)\propto \exp(CF(w)\langle \vec{w}_1, \theta \rangle + (1 - CF(w))\langle\vec{w}_2, \vec{c} \rangle ), \theta, \vec{w}_1, \vec{w}_2 \in\mathbb{R}^d$ would work.
Note that the conditional distribution $P(w | CF(w)=0, c)$ would become different from $P(w | c)$ in order for them to reside in a valid probability space.

To study the relation between $P(w|c)$ and $P(w | CF(w)=0, c)$, let us combine Eq.~\ref{equ:context_free_basic} with standard probability rules to obtain
\begin{align}
P(w|c) &= P(w, CF(w)=1 | c)  + P(w, CF(w)=0 | c) \nonumber \\
& = P(w | CF(w)=1, c) P(CF(w)=1|c) +  P(w | CF(w)=0, c) P(CF(w)=0 |c) \nonumber \\
& = \tilde{P}(w) P(CF(w)=1 | c) + P(w | CF(w)=0, c) (1 - P(CF(w)=1 | c)).
\end{align}
If we denote by $\chi(w, c) \equiv P(CF(w)=1 | c)$, which we call the $\mathbf{\chi}$-\textbf{function} throughout this paper, the above equation then becomes
\begin{equation}\label{equ:prob_decomposition}
P(w|c)  = \tilde{P}(w) \chi(w, c) +  P(w | CF(w)=0, c) (1 - \chi(w, c)), 
\end{equation}
where $\chi(w, c)\in [0, 1]$ indicates how likely $w$ is independent of its context $c$, and the conditional probability $P(w|c)$
is decomposed into a weighted sum of two distributions, one denoting the context free part (\emph{i.e.}, $\tilde{P}(w)$),
another the context sensitive part (\emph{i.e.}, $P(w | CF(w)=0, c)$).

We can further connect Eq.~\ref{equ:prob_decomposition} to its embedding space using the exponential family assumption discussed in Sec~\ref{subsection:exponential_family}.
First of all, let us assume $ P(w | CF(w)=0, c) =  \exp(\langle \vec{w'}, \vec{c} \rangle - \ln Z_c) $ where $\vec{w'}$ is the embedding of $w$ when it is context sensitive and $Z_c$ is the partition function depending only on $c$.
Then Eq.~\ref{equ:prob_decomposition} becomes
\begin{align}
P(w|c)  &= \tilde{P}(w) \chi(w, c) +  P(w | CF(w)=0, c) (1 - \chi(w, c)) \nonumber \\
              &= \exp(\ln (\tilde{P}(w)))\chi(w, c) + \exp(\langle \vec{w'}, \vec{c} \rangle - \ln Z_c) (1 - \chi(w, c)).
\end{align}

Also from the convexity of the exponential function ($\mu e^a + (1-\mu) e^b \geq e^{\mu a + (1 - \mu) b}, \mu\in [0, 1]$), we have (let $a=\ln (\tilde{P}(w))$, $b=\langle \vec{w'}, \vec{c} \rangle - \ln Z_c$ and $\mu=\chi(w, c)$)
\begin{align}\label{equ:decomposition}
P(w|c) &\geq  \exp(\chi(w, c) \ln \tilde{P}(w) + (1 - \chi(w, c)) (\langle \vec{w'}, \vec{c} \rangle - \ln Z_c) )  \nonumber \\
&= \frac{\exp(\chi(w, c) (1 + \frac{\ln \tilde{P}(w)}{\ln Z_c}) \langle \vec{v}_c,  \vec{c} \rangle + (1 - \chi(w, c))  \langle \vec{w'},  \vec{c} \rangle)}{Z_c} \nonumber \\
&= \frac{ \exp(\langle \chi(w, c) (1 + \frac{\ln \tilde{P}(w)}{\ln Z_c})  \vec{v}_c  + (1 - \chi(w, c)) \vec{w'},  \vec{c} \rangle)}{Z_c}.
\end{align}
Here $\vec{v}_c$ is any vector that satisfies
\begin{align}\label{equ:vc}
Z_c = \sum_{w}\exp{\langle \vec{w'}, \vec{c} \rangle} = \exp(\langle \vec{v}_c, \vec{c}  \rangle).
\end{align}
Intuitively, $\vec{v}_c$ represents the partition function in the embedding space which summarizes the information on all $\vec{w'}$ interacting with the the same context $c$.
The solution to Eq.~\ref{equ:vc} is not unique, which gives us the flexibility to impose additional constraints in algorithm design as we will elaborate in Sec.~\ref{section:new_models}.
 
By comparing Eq.~\ref{equ:decomposition} with the original log-linear model of Eq.~\ref{equ:loglinearmodel_simplified} in the general case that $\vec{w}$ depends on both $w$ and $c$, 
we can establish the following decomposition for the embedding of $w$:
\begin{align}
\vec{w} \approx \chi(w, c) (1 + \frac{\ln \tilde{P}(w)}{\ln Z_c})  \vec{v}_c  + (1 - \chi(w, c)) \vec{w'}
\end{align}
Furthermore, if we assume $\frac{\ln \tilde{P}(w)}{\ln Z_c}\approx 0$, a very simple embedding decomposition formula can be obtained as follows.

\begin{theorem*}\textbf{(Embedding Decomposition Formula, \emph{a.k.a.}, EDF)}. The log-linear model of a conditional probability $P(w|c)$ can 
be approximately decomposed into two parts: context sensitive and context free, connected by the $\chi(\cdot)$ function that 
indicates its level of context-freeness
\begin{align}\label{equ:magic_decomposition_formula}
\vec{w} \approx \chi(w, c)  \vec{v}_c  + (1 - \chi(w, c)) \vec{w'}.
\end{align}
\end{theorem*}

Here $\vec{v}_c$ represents the context-free part of an embedding, and $\vec{w'}$ represents the context-sensitive part. 
This is because $P(w|c)\propto \exp(\langle\vec{w}, \vec{c}\rangle)$, and when $\chi(w, c)\rightarrow 1$, $P(w|c) \rightarrow \text{constant}$ (Eq.~\ref{equ:vc}) which is insensitive to the observation $w$\footnote{This is related to how the out-of-vocabulary (oov) tokens are dealt with in neural network based language models, where an oov token may appear anywhere (context-free). The usual approach is to map them to a single vector, corresponding to $v_c$ in our case.}. 
On the other hand, when $\chi(w, c)\rightarrow 0$, $w$'s embedding $\vec{w'}$ represents something that could interact with its context, \emph{i.e.}, $\langle \vec{w}, \vec{c} \rangle$ changes with $w$, and therefore is sensitive to the context.

In practice, we may make reasonable assumptions such as $\chi(\cdot)$ is independent of $c$ (\emph{e.g.}, ``the" is likely to be context free no matter where it appears),
or $\vec{v}_c$ is orthogonal to $\vec{w'}$ for all $w$ (\emph{e.g.}, the oov tokens can appear in any sentence and their distribution should be independent of the context, \emph{i.e.}, $\langle\vec{w}, \vec{c}\rangle$ is a constant). We will discuss in details in Sec.~\ref{section:new_models} how these simplified assumptions can help reduce model complexity.

\subsection{Bag-of-words Embedding}
\label{subsection:bof_embedding}
In the above we have only discussed the embedding model for a single variable $w$.
It is straightforward to generalize it to the case when we want to find the embedding of a set of variables $s = \{w_1, \ldots, w_n\}$.
We can combine the naive independence assumption and Eq.~\ref{equ:decomposition} to obtain an approximation of $P(s|c)$ as follows:
\begin{align}
P(s|c) &= \prod_{i} P(w_i|c) \nonumber \\
          &\geq  \exp (\sum_i \langle \chi(w_i, c) \vec{v}_c (1 + \frac{\ln \tilde{P}(w_i)}{\ln Z_c})  + (1 - \chi(w_i, c)) \vec{w_i'},  \vec{c} \rangle) / Z_c^n \nonumber \\
          &\propto  \exp (\sum_i \langle \chi(w_i, c) \vec{v}_c (1 + \frac{\ln \tilde{P}(w_i)}{\ln Z_c})  + (1 - \chi(w_i, c)) \vec{w_i'},  \vec{c} \rangle) \nonumber \\
          &\approx \exp  (\langle \vec{v_c} \sum_i \chi(w_i, c)   + \sum_i (1 - \chi(w_i, c)) \vec{w_i'},  \vec{c} \rangle)
\end{align}
By comparing it with the the log-linear model $P(s|c) \propto \exp(\langle \vec{s}, \vec{c}\rangle)$, we obtain the \emph{bag-of-features embedding formula}:
\begin{align}\label{equ:bof_embedding}
\vec{s} \approx \vec{v}_c \sum_i \chi(w_i, c)   + \sum_i (1 - \chi(w_i, c)) \vec{w_i'},
\end{align}
which is simply the sum of single-variable embeddings.
This gives us a very straightforward way of compositing bag-of-feature embeddings beyond the averaging scheme $\vec{s} \approx \sum_i \vec{w_i} / n$ (\emph{e.g.}, \cite{sem_wieting2015towards}).


\section{Existing Models Revisited and Revised}
\label{section:new_models}
In this section, we explore the connections between our context-awareness based framework (Sec.~\ref{section:math_foundation}) and various existing neural network models.
Specifically, we demonstrate how our EDF (Eq.~\ref{equ:magic_decomposition_formula}) can fit into various machine learning models and therefore provide a unified principle.

\subsection{Sentence Re-embedding}
\label{subsection:casem}
The task of re-embedding is to retrain any existing (pre-trained) embedding mapping ($\mathbf{T}: \mathcal{W} \mapsto \mathbb{R}^d$) for a new task to obtain a new mapping $\tilde{\mathbf{T}}: \mathcal{W} \mapsto \mathbb{R}^d$ that better fits the distribution of new data.
This is useful when training new embeddings is infeasible due to lack of (labeled) data. \cite{sem_wieting2015towards} demonstrated that by simply averaging the (pretrained) embeddings of all the words  in a sentence can already achieve better performance than sequence models. 
Later, \cite{sem_arora2017} further improved this bag-of-words based approach by introducing an additional term in the log-linear model to reduce the impact of those frequent and uninformative words, which is the primary inspiration for our work.

Specifically, in the work of~\cite{sem_arora2017}, the probability that a word $w$ is produced in a sentence $c$ (represented by a discourse vector, or embedding $\vec{c}$) is modeled by,
\begin{align}\label{equ:arora_model}
P(w|c) &= \alpha P(w) + (1-\alpha) \frac{\exp(\langle  \vec{w}, \vec{c} \rangle)}{Z_c} \nonumber \\
           &\text{where } \vec{c} = \beta\vec{c}_0 + (1 - \beta)\vec{c'} \text{ and } \vec{c}_0 \perp \vec{c'}
\end{align}
Here $\alpha,\beta$ are scalar hyper-parameters that are globally constant, $\vec{c}_0$ is a global vector representing the common part of the embeddings (computed via PCA on all training sentences), and $\vec{c'}$ changes with each sentence $c$.

Surprisingly or not, Eq.~\ref{equ:prob_decomposition} resembles Eq.~\ref{equ:arora_model} in almost every way\footnote{Note that Eq.~\ref{equ:prob_decomposition} is derived using standard probability rules whereas Eq.~\ref{equ:arora_model} is heuristically defined.}. It is easy to identify the rough correspondences
 $\chi(w, c) \leftrightarrow \alpha$, $\tilde{P}(w) \leftrightarrow P(w)$ and $P (w| CF (w)=0, c) \leftrightarrow \frac{\exp(\langle \vec{w}, \vec{c} \rangle)}{Z_c}$ between them.
 However, despite their resemblance in appearance, the underlying assumptions are largely different as highlighted below.
\begin{enumerate}
	\item $\chi(w, c)$ has a clear mathematical meaning ($P(CF(w)|c)$) under our context-awareness setting which changes with both $w$ and $c$, whereas $\alpha$ is a global constant that is heuristically defined.
	\item $\tilde{P}(w)$ in Eq.~\ref{equ:prob_decomposition} is a different distribution (frequency of $w$ when it is context free) from $P(w)$ (overall frequency of $w$). Otherwise it can be proven that there is no valid measurable space that satisfies Eq.~\ref{equ:prob_decomposition}. Hence our model does not involve overall word frequency.
	\item Although $\vec{c}_0$ of Eq.~\ref{equ:arora_model} resembles $\vec{v}_c$ in our case, the former is defined for each sentence and the later is on the side of $w$. Besides, $\vec{v}_c$ is defined for each value of $c$ (Eq.~\ref{equ:vc}), whereas $\vec{c}_0$ is a global constant.
	\item There is no correspondence of $\beta$ in our new model.
	\item The log-linear model $\frac{\exp(\langle \vec{w}, \vec{c} \rangle)}{Z_c} $ in Eq.~\ref{equ:arora_model} is defined for the original distribution $P(w|c)$, which can lead to a contradictory equation $P(w|c) = P(w)$. In our case, we assume $P (w|CF (w)=0, c) \propto \exp(\langle \vec{w'}, \vec{c} \rangle)$ such that everything remains in a valid probability space.
	\item The context $c$ in~\cite{sem_arora2017} represents the sentence that a word appears. This is confusing because if a word is already shown in its context, the probability $P(w|c)$ should be $1$. In contrast, our treatment of context is flexible to represent any high-level topic underlying the sentence, or even a whole corpus of sentences.
\end{enumerate}

Eq.~\ref{equ:arora_model} is solved via computing the first principle components of all sentence embeddings as $\vec{c}_0$ then remove them from each sentence to obtain $\vec{w'}$.
In contrast, in the rest of this section, we show that the sentence re-embedding problem can be formulated as an energy minimization problem which no longer requires PCA to estimate the global information $\vec{v}_c$. 

\subsubsection{Context-aware sentence re-embedding (CA-SEM)}
To find the ``correct" solution for sentence re-embedding, we resort to the EDF (Eq.~\ref{equ:magic_decomposition_formula})
and solve for a new embedding $\vec{w'}$ for each $w$ from a given corpus of sentences $S = \{s_1, \ldots, s_n\}$ and pre-trained embedding $\vec{w}$.
In other words, we would like to minimize the distance (\emph{e.g.}, L2-norm) between $\vec{w}$ and $\chi(w, c) \vec{v}_c + (1 - \chi(w, c)) \vec{w'}$ over all words $w\in s_i, i = 1, \ldots, n$.
Once we have found the new embeddings $\vec{w'}, w\in\mathcal{W}$, the embedding of any sentence can be computed based on the bag-of-features embedding composition of Eq.~\ref{equ:bof_embedding}. 

However, there are a number of ambiguities that needs to be clarified to make the problem well-defined.
First, both $\chi(w, c)$ and $\vec{v}_c$ change with the context $c$, so what exactly is the context for the re-embedding problem?
The answer largely depends on the problem we want to solve. 
Regardless, we can always assume that all the sentences in $\mathcal{S}$ share some common topic (\emph{e.g.}, about conversation, or movie review).
This means we can be spared from having to explicitly model $\vec{c}$ and just assume that $\chi(w, c)$ and $\vec{v}_c$ do not change with $c$.
Hence we only need to solve for a global context vector, denoted as $\vec{v}_0$, and a weight $\tilde{\chi}(w)\in [0, 1]$ for each $w$\footnote{Here we justified the use of a single global vector $\vec{c}_0$ in Eq.~\ref{equ:arora_model} with a new explanation. Moreover, our framework invites designs of new algorithms such as automatic discovery of hierarchical topics from given corpus: first compute the reembedding based on global context; then group the new embeddings, and discover the sub-context in each group, etc.}. 

Second, the minimization problem is ill-posed if we do not impose constraints on $\vec{w'}$. 
For instance, a trivial solution $\vec{w'} = \vec{w}$ and $\tilde{\chi}(w) = 0$, $\forall w$ can always achieve a perfect but useless decomposition, in which case the sentence embedding (Eq.~\ref{equ:bof_embedding}) becomes equivalent to summing the embeddings of words.
To obtain more meaningful embeddings, we find the assumption $\vec{v}_0 \perp \vec{w'}, \forall w$ used in~\ref{equ:arora_model} makes good theoretically sense under our new context-awareness framework.
In fact, this assumption essentially normalizes the partition function of Eq.~\ref{equ:vc} to be a constant.
With it, we can always guarantee that the solution is non-trivial.

In sum, the problem of sentence embedding can be cast as solving the following
constrained energy minimization problem:
\begin{align}\label{equ:sem_reembedding}
\min_{\vec{w'}, \vec{v}_0, \tilde{\chi}(w)\in [0, 1]} &\:\:\sum_{s\in \mathcal{S}} \sum_{w\in s} (\vec{w} - \tilde{\chi}(w) \vec{v}_0 - (1 - \tilde{\chi}(w)) \vec{w'})^2 \nonumber \\
\text{s.t.} &\:\:\:\:\vec{v}_0 \perp \vec{w'}, \tilde{\chi}(w)\in [0, 1], \forall w
\end{align}
However, question remains how to efficiently optimize the above energy. 
It turns our that there exists an efficient block-coordinate descent algorithm such that each step can be solved in a closed form.

\subsubsection{Optimization algorithm for CA-SEM}
\begin{figure}[t]
\centering
\begin{tabular}{c}
    \mbox{\epsfig{figure= 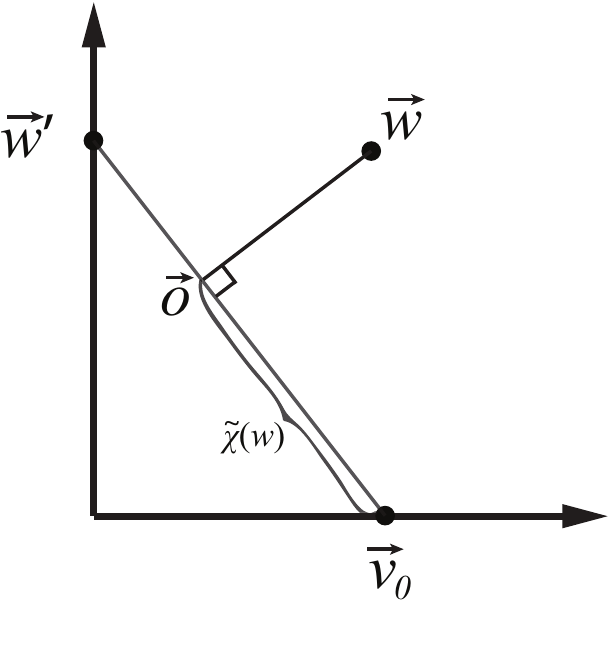, height = 5cm}}
\end{tabular}
\caption{\small Illustration of solving the optimal $\tilde{\chi}(w)$ from $\vec{w}$, $\vec{w'}$ and $\vec{v}_0$ in the $2$D case. It first projects $\vec{w}$ to the segment connecting $\vec{w'}$ and  $\vec{v}_0$ at $\vec{o}$, then the value of $\tilde{\chi}(w)$ can be computed as the ratio $|\vec{v}_0 - \vec{o}| / |\vec{w'} - \vec{v}_0|$.} \label{fig:compute_chi}
\end{figure}

Let us first rearrange Eq.~\ref{equ:sem_reembedding} by grouping similar terms. If we denote by $n_w$ the number of times that $w$ appears among sentences in $\mathcal{S}$,
then Eq.~\ref{equ:sem_reembedding}  becomes
\begin{align}\label{equ:sem_reembedding_regroup}
\min_{\vec{w'}, \vec{v}_0, \tilde{\chi}(w)\in [0, 1]} &\:\:\sum_{w\in \mathcal{S}} n_w (\vec{w} - \tilde{\chi}(w) \vec{v}_0 - (1 - \tilde{\chi}(w)) \vec{w'})^2 \nonumber \\
\text{s.t.} &\:\:\:\:\vec{v}_0 \perp \vec{w'}, \tilde{\chi}(w)\in [0, 1], \forall w
\end{align}
This is a nonlinear optimization problem and if we solve it using gradient descent it can easily stuck at local minima.
Fortunately, if we solve each of $\vec{v}_0$, $\{\vec{w'} | w \in \mathcal{W}\}$ and $\{\tilde{\chi}(w) | w \in \mathcal{W}\}$ by fixing the other values in a block-coordinate descent algorithm,
it turns out that each sub-problem can be solved efficiently in a closed form:

\begin{enumerate}
	\item Solve $\vec{v}_0$: If we fix the values of $\tilde{\chi}(w)$ and $\vec{w'}$, $\forall w$, the optimal value of $\vec{v}_0$ can be solved via a standard linear least squares minimization problem.
	\item Solve $\vec{w'}$:  If we fix the value of $\vec{v}_0$, we can also obtain $\vec{w'}$ by removing the component along $\vec{v}_0$ from $\vec{w}$ to respect the orthogonal constraint in Eq.~\ref{equ:sem_reembedding}, namely $\vec{w'} = (I - \vec{v}_0 \vec{v}_0^T / |\vec{v}_0|^2) \vec{w}$. This gives us a closed form solution for each $\vec{w'}$ once we know the value of $\vec{v}_0$.
	\item Solve $\tilde{\chi}(w)$: The optimization problem of Eq.~\ref{equ:sem_reembedding_regroup} can be divided into independent sub-problems if we fix the values of $\vec{v}_0$ and $\vec{w'}$, meaning we only need to solve for the optimal value of $\tilde{\chi}(w)\in [0, 1]$ for each $w$ by minimizing $(\vec{w} - \tilde{\chi}(w) \vec{v}_0 - (1 - \tilde{\chi}(w)) \vec{w'})^2$. Geometrically, this is equivalent to finding the closest point on the line segment connecting $\vec{v}_0$ and $\vec{w'}$ to the point $\vec{w}$ (Fig.~\ref{fig:compute_chi}), which can be easily solved in a closed form.
\end{enumerate}

Note that both $\tilde{\chi}(w)$ and $\vec{w'}$ can be computed in a closed form if we know the value of $\vec{v}_0$, which is the only thing we need to store for new word/sentence embeddings.
Alg.~\ref{alg:ca_sem_algorithm} summarizes our CA-SEM algorithm for re-embedding.
After the re-embedding of each word $w$ (\emph{i.e.}, $\vec{w'}$) is computed, a sentence embedding can be computed using the following equation:
\begin{align}
\vec{s} = \vec{v}_0 \sum_{w\in s} \tilde{\chi}(w) + \sum_{w\in s} (1 - \tilde{\chi}(w)) \vec{w'},
\end{align}
which represents the context-sensitive part of Eq.~\ref{equ:bof_embedding}.

One practical concern is the out-of-vocabulary (oov) words that are in the training/inference sentences but not in the pre-trained word embedding dictionary.
These oov tokens can appear anywhere and hence context-free, so we only need to map them to $\vec{v}_0$ and let $\tilde{\chi}(\text{oov}) = 1$.
In an extreme case, when all the words in a sentence are oov, the embedding becomes $|s|\vec{v}_0$, whose direction implies its level of context-freeness and its norm implies its extent (number of oov's in the sentence).

\begin{algorithm}[!]
\SetKwInOut{Input}{Input}
\SetKwInOut{Output}{Output}
\caption{\small{Block-coordinate Algorithm for Sentence Reembedding.}}\label{alg:ca_sem_algorithm}
\small
\Input{A set of sentences $\mathcal{S}$ and a pre-trained embedding $\{\vec{w} | w \in \mathcal{W}\}$.}
\Output{The context vector $\vec{v}_0\in  \mathbb{R}^n$.}
\vspace{2mm}
\textbf{Initialization:} Initialize $\vec{v}_0$ with the first principle component of all the points $\{\vec{w} | w\in s, s\in \mathcal{S} \}$.\\
\textbf{Step 1:} Fix $\vec{v}_0$, solve for $\vec{w'}$ and $\chi(w)$ for each $w$. \\
\textbf{Step 2:} Fix $\vec{w'}$ and $\chi(w)$ for each $w$, solve for $\vec{v}_0$. \\
\textbf{Step 3:} Evaluate the energy of Eq.~\ref{equ:sem_reembedding_regroup} with new values, if it drops from previous iteration and maximal iteration is not reached, repeat Step 1; otherwise, stop. 
\normalsize
\end{algorithm}

\subsection{Bag of Sparse Features Embedding (CA-BSFE)}
\label{subsection:cabsfe}
Representing a set of sparse features  (\emph{e.g.}, words or categorical features) in a vector space is a prerequisite in neural network models involving sparse input. 
When the number of sparse feature varies with input (\emph{i.e.}, bag of features), it is necessary to composite them into a single, fixed length vector to make sure that the architecture of an NN would remain fixed, a problem known as bag of sparse features embedding (BSFE) (or sentence/phrase embedding if the feature is a word). Compared to all the progresses made on the theoretical basis for embedding individual features (\emph{e.g.}, ~\citep{emb_bengio2003neural,emb_mnih2007three,emb_pennington2014glove,emb_arora2016latent}), fewer work has been done on the theoretical aspect of embedding bag of features (\emph{e.g.}, ~\citep{sem_mitchell2010composition,sem_arora2018}. Previously, a de facto approach for compositing BSFE in NN is to map each sparse feature $w$ to its embedding $\vec{w}$ and the embedding is computed by their average $\sum_{w\in s} \vec{w} / |s|$, which is usually fed into the next layer in a deep network.

Our EDF (Eq.~\ref{equ:magic_decomposition_formula}) provides a sound basis for compositing bag of sparse features embeddings that can be implemented in a neural network.
We can design an NN layer that explicitly solves for $\chi(w, c)$, $\vec{w^e}$, and $\vec{v}_c$ for each sparse feature $w$ in the MED, then a BSFE can be computed by Eq.~\ref{equ:bof_embedding}.
Because neural networks are highly modularized such that model definition and training are well separated, here we only need to focus on the design of the BSFE layer and let back propagation algorithm take care of the training. 

To start with, let us look at the meaning of context in BSFE. Similar to CA-SEM (Sec.~\ref{subsection:casem}), we can assume a global context $\vec{v}_0$ is shared among all the input. 
Therefore in the architecture of BSFE we need to solve for a global vector $\vec{v}_0\in \mathbb{R}^n$ during training, where $n$ is the embedding dimension.

The next thing to look at is the $\chi$-function in our EDF which depends on both input feature $w$ and context $\vec{v}_0$. Because now $\vec{v}_0$ is a global variable that does not change with each input, the $\chi$-function essentially becomes a function of $w$, namely $\tilde{\chi}(w)$.
Also the role of $\chi$-function is essentially a soft gating function that switches between $\vec{v}_0$ and $\vec{w'}$, similar to those gates defined in LSTM model~\cite{lstm_original}.
Hence we simply need to employ the standard sigmoid function for its definition, which automatically guarantees that the value of $\chi$-function are always within $[0, 1]$.

However, there is still one more obstacle to overcome before we can write down the equation for $\tilde{\chi}(w)$: the sigmoid function takes a real vector as its input whereas $w$ is a sparse feature and it needs to be mapped to an embedding first. Here we have two choices: i) reuse the embedding vector $\vec{w'}$ or ii) define a new embedding $\vec{w}_{\sigma} \in\mathbb{R}^m$ specifically for computing $\tilde{\chi}(w)$, where $m\in\mathbb{N}$ is not necessarily the same as the embedding dimension $n$. Theoretically speaking, ii) makes better sense since it decouples the embedding $\vec{w^e}$ and the $\chi$-function. 
Otherwise, EDF is just a nonlinear function of $\vec{w'}$ and the number of parameters of the sigmoid function would be constrained by $n$. 

\begin{figure}[t]
\centering
\begin{tabular}{cc}
    \mbox{\epsfig{figure= 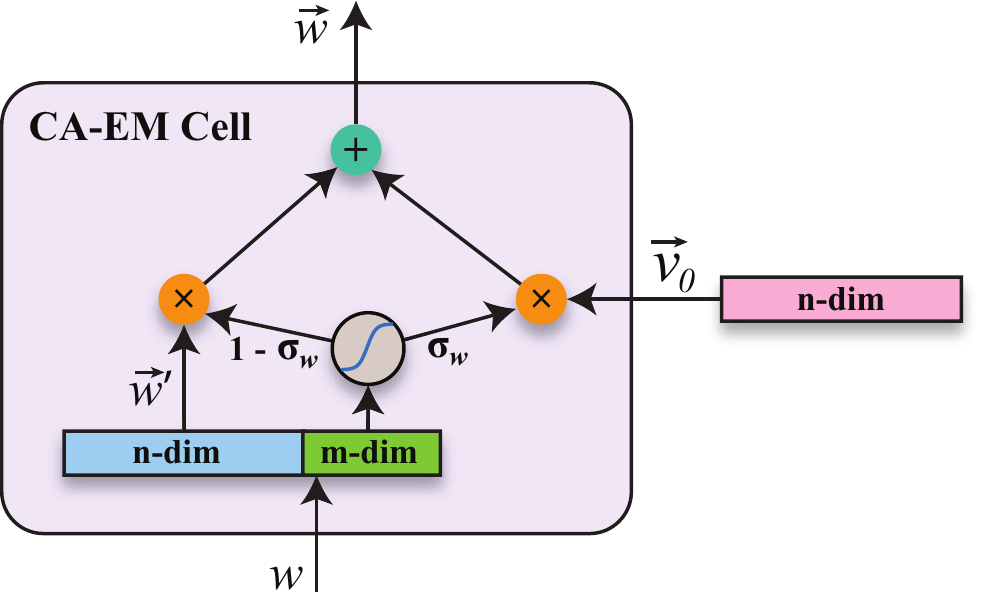, height = 3.2cm}} & \mbox{\epsfig{figure= 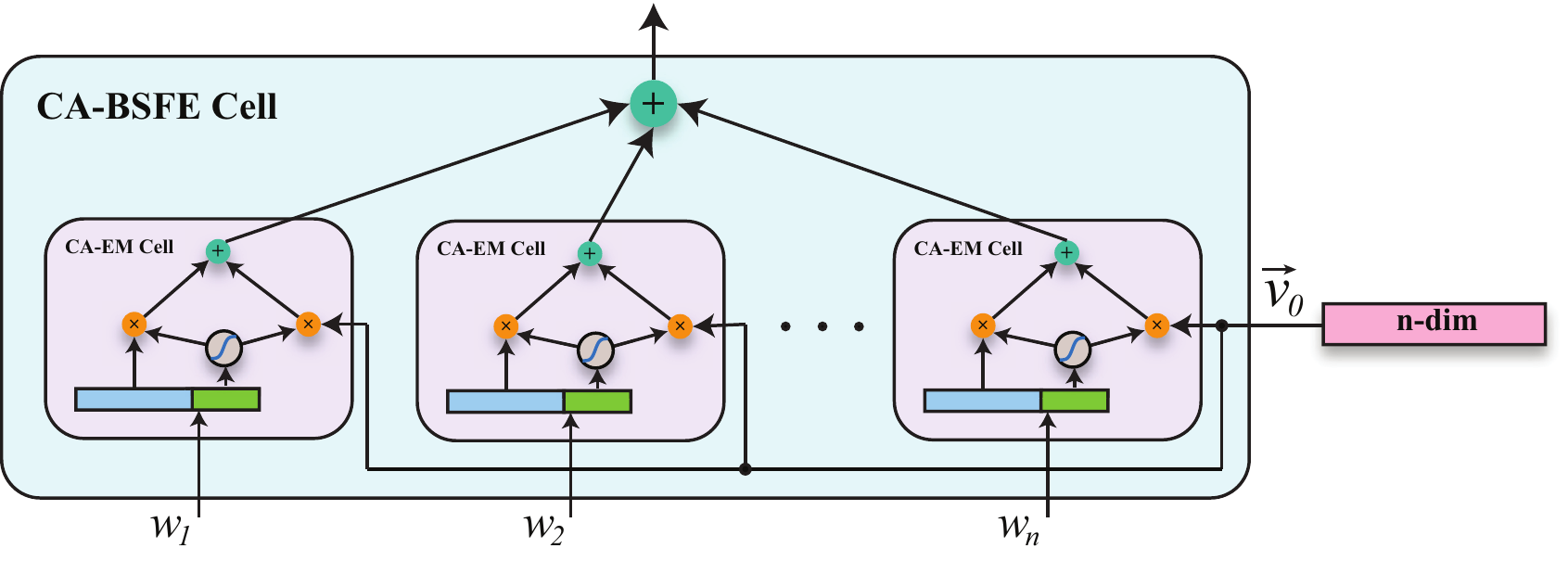, height = 3.2cm}} \\
    {\scriptsize (a) Single feature embedding.} & {\scriptsize (b) Bag of features embedding.}
\end{tabular}
\caption{\small The architecture for single feature embedding (CA-EM) and multiple feature embedding (CA-BSFE). Here $\sigma_w$ in (a) denotes the output of the $\chi$-function defined in Eq.~\ref{equ:sigmaforbsfe} and a global vector $\vec{v}_0$ is shared by all the input.} \label{fig:ca_bsfe}
\end{figure}

In sum, the $\chi$-function can be implemented by the following sigmoid function:
\begin{align}\label{equ:sigmaforbsfe}
\tilde{\chi}(w;\theta) = \frac{1}{1 + \exp(-\theta^T \vec{w}_{\sigma} )},
\end{align}
where $\vec{w^{\sigma}}$ is an $m$-dimensional vector associated to each $w$ and $\theta$ is a global m-dimensional vector shared by all $\chi$-functions.
From Eq.~\ref{equ:bof_embedding},  the embedding of a bag of features $s = \{w_1, w_2, \ldots\}$ then becomes
\begin{align}\label{equ:bsfe}
\vec{s} = \vec{v}_0 \sum_{w \in s} \tilde{\chi}(w; \theta) +  \sum_{w\in s} (1 - \tilde{\chi}(w; \theta)) \vec{w}.
\end{align}
Fig.~\ref{fig:ca_bsfe} illustrates the architecture for single feature embedding (CA-EM) and BSFE embedding (CA-BSFE).
Note that now the input embedding's dimension is $n + m$ and the model split it into two parts for $\vec{w'}$ and $\vec{w}_{\sigma}$, respectively.

Finally, let us connect our approach with the de facto, embedding lookup based network layer.
If we define
\begin{align}
\sigma(w)=\begin{cases}
0, \text{if } w \in \mathcal{W} \\
1, \text{otherwise, or OOV}
\end{cases}
\end{align}
and denote the embedding of OOV tokens as $\vec{v}_0$, then the embedding of a sparse feature can be represented as
\begin{align}
\vec{w} = \vec{v}_0 \sigma(w) + (1 - \sigma(w))\vec{w'},
\end{align}
which is similar to our EDF in form, except that in our case, $\sigma(w)\in[0, 1]$.
Hence our new model only provides a soft version of existing approaches, though it fundamentally changed the meaning of its variables.

\subsection{Attention Model (CA-ATT)}
\label{subsection:caatt}
The concept of attention used in neural networks (\emph{e..g}, ~\cite{attention_nmt14}) is closely related to that of context, where attention decides which contextual input should be ignored or not.
In a sequence model, the conditional distribution $P(w|c, s)$ is often considered, where $w$ is the output (\emph{e.g.}, $y_t$ in a sequence-to-sequence model~\cite{seq_seq2seq} at stage $t$), 
$c$ and $s$ are different contexts for $w$ (\emph{e.g.},  in a sequence-to-sequence model $c = \{y_1, \ldots, y_{t - 1}\}$ is the output sequence before $t$ and $s = \{x_1, \ldots, x_T\}$ is the whole input sequence).
In addition, attention is often applied when both the embeddings of $c$ (\emph{e.g.}, the hidden state of an RNN from previous steps, also known as the \emph{query}) and each elements of $s$ (\emph{e.g.}, the hidden states of the whole input sequence, also known as the \emph{memory}) are given.
Intuitively, attention models the importance of input/memory $s = \{s_i| i = 1, \ldots, T\}$ from given context/query $c$.

\begin{figure}[t]
\centering
\begin{tabular}{c}
    \mbox{\epsfig{figure= 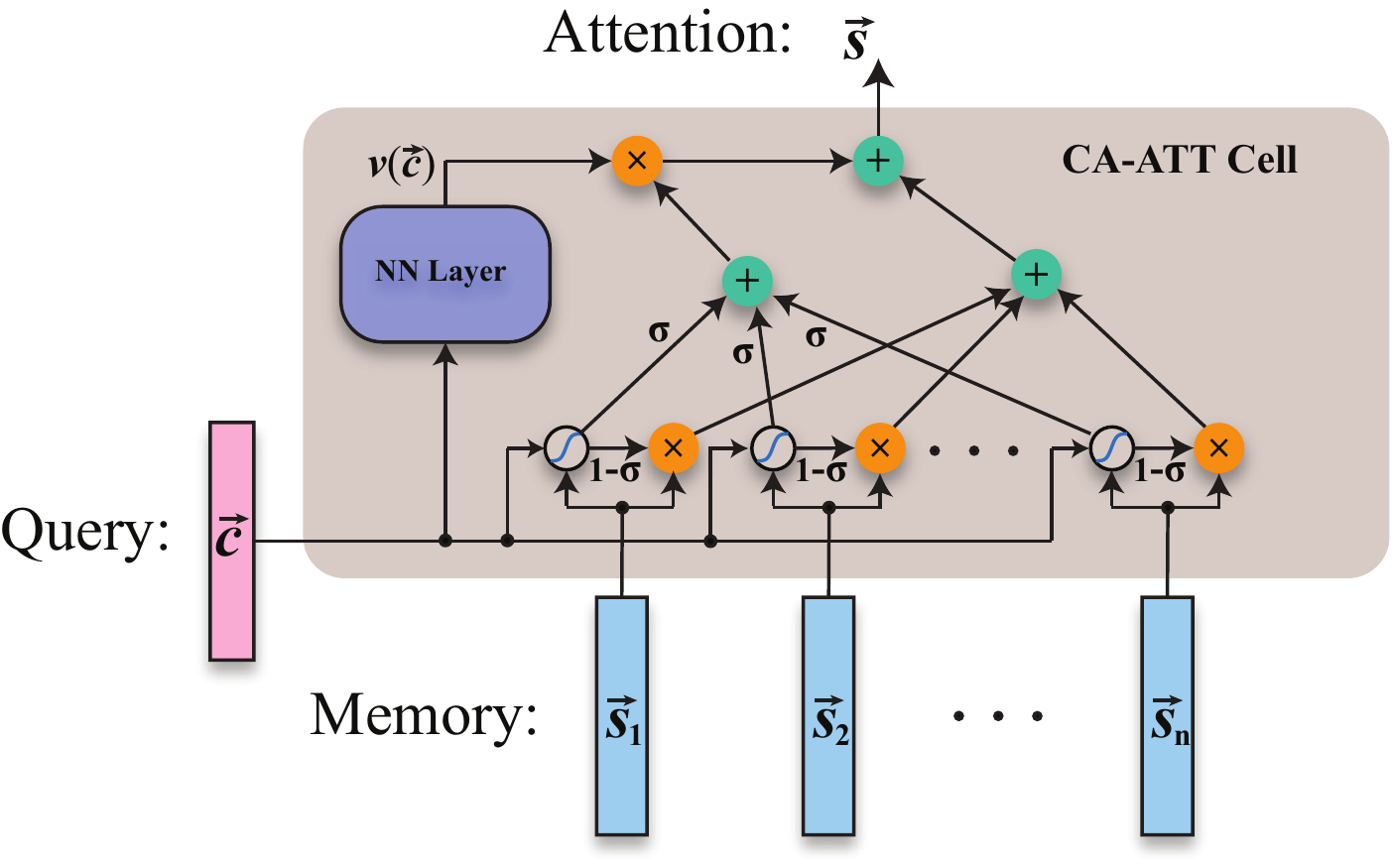, height = 6cm}} 
\end{tabular}
\caption{The architecture for context aware attention network (CA-ATT).} \label{fig:ca_att}
\end{figure}

Therefore, we only need to draw our attention (pun intended) to the distribution $P(s|c)$\footnote{The full context-aware treatment of the the conditional probability $P(w|s, c)$ will be given in the next section.}, which is often represented by a vector $\vec{s}$ (\emph{a.k.a.}, attention vector) in a sequence model. From the EDF for bag of features (Eq.~\ref{equ:bof_embedding}), the embedding of $s$ can be represented as
\begin{align}\label{equ:ca-att}
\vec{s} = \vec{v}(\vec{c}) \sum_{i = 1}^n \chi(\vec{s}_i, \vec{c}) + \sum_{i = 1}^n (1 - \chi(\vec{s}_i, \vec{c})) \vec{s_i},
\end{align}
where $\vec{c}$ and $\vec{s_i}, i \in \{1, \ldots, n\}$ are given as input, and we only need to define the forms for $\vec{v}(\cdot)$ and $\chi(\cdot, \cdot)$ to obtain $\vec{s}$.
Here the value $1 -  \chi(\vec{s}_i, \vec{c}), i = 1, \ldots, n$ represents the importance of input $s_i$ given the context $c$, which translates to context-sensitive-ness in our context-aware framework.
Compared to the embedding models in Sec.~\ref{subsection:casem} and~\ref{subsection:cabsfe}, our attention model no longer assumes a global context and therefore the $\chi$-function also depends on the context input $\vec{c}$.
Our new architecture for attention model can be illustrated in Fig.~\ref{fig:ca_att}.

Before discussing the definitions of $\vec{v}(\cdot)$ and $\chi(\cdot, \cdot)$, let us compare our context aware attention model (CA-ATT) with the original one (\cite{attention_nmt14}).
\begin{enumerate}
	\item The original model composites $\vec{s}$ with an equation $\vec{s} = \sum_i \alpha(\vec{s_i}, \vec{c})\vec{s_i}$ where 
	\begin{align}\label{equ:attention_original}
	\alpha(\vec{s_i}, \vec{c}) = \frac{\exp(f(\vec{s_i}, \vec{c}))}{\sum_i \exp(f(\vec{s_i}, \vec{c}))}.
	\end{align}
	In contrast, we replace $\alpha(\vec{s_i}, \vec{c})$ with $1 - \chi(\vec{s_i}, \vec{c})$, which has a clear meaning: the context-sensitive-ness of $s_i$ given the context $c$.
	However, one major difference is $\alpha(\cdot, \cdot)$ is normalized as a softmax distribution, which implicitly assumes only one of the input $\{s_i | i = 1, \ldots, n\}$ should be
	attended to. Whereas in our case, we employ the more general assumption that the number of important input can be arbitrary.
        \item Another major difference between our model and the original one is the introduction of $\vec{v}(\cdot)$ to account for the situation when none of the input is relevant.
        In such a case, for the original model, $f(\vec{s_i}, \vec{c})\rightarrow 0$ and $\alpha(\vec{s_i}, \vec{c})$ would approach $1/ |s|$, and $\vec{s}$ is simply the average of the input embeddings, which still allows the information of $s$ to be contributable to the final outcome of $w$. In contrast, we use a separate vector $v(\vec{c})$ to represent the state when the output is independent of the input (context-free). Hence when $1 - \chi(\vec{s_i}, \vec{c}) \rightarrow 0, i = 1, ..., T$, the influence of $s$ is eliminated (Eq.~\ref{equ:ca-att}). 
\end{enumerate}

Now let us elaborate on the details of $\vec{v}(\vec{c})$, which is equivalent to defining a new layer of activation from previous layer's input $\vec{c}$. 
A general form of $\vec{v}(\cdot)$ can be represented as
\begin{align}
\vec{v}(\vec{c}; \mathbf{W}_h, \mathbf{b}_h) = \sigma_h (\mathbf{W}_h \vec{c} + \mathbf{b}_h),
\end{align}
where $\sigma_h$ can be any activation function such as hyperbolic tangent $\tanh(\cdot)$, $\mathbf{W}_h\in\mathbb{R}^{dim(\vec{v})\times dim(\vec{c})}$ and $\mathbf{b}_h\in\mathbb{R}^{dim(\vec{v})}$ are the kernel and bias, respectively, where $dim(\cdot)$ represents the dimension of an input vector.

As for $\chi(\vec{s_i}, \vec{c})$, a straightforward choice is the sigmoid function:
\begin{align}\label{equ:sigma_for_caatt}
\chi(\vec{s_i}, \vec{c}; \mathbf{w}_g, \mathbf{u}_g, b_g) = \sigma_g (\mathbf{w}_g \vec{s_i} + \mathbf{u}_g \vec{c} + b_g),
\end{align}
where $\sigma_g$ is the sigmoid function, $\mathbf{w}_g\in\mathbb{R}^{dim(\vec{s})}$, $\mathbf{u}_g\in\mathbb{R}^{dim(\vec{c})}$ and $b_g\in\mathbb{R}$. Note that here we map each $(\vec{s_i}, \vec{c})$ to a scalar value within $[0, 1]$, which denotes the opposite of attention, \emph{i.e.}, context-freeness of each input $s_i$.

Finally, we would like to add a note that our framework also implies many possibilities of implementing a neural network with attention mechanism.
For example, in the CA-BSFE model (Sec.~\ref{subsection:cabsfe}), we have proposed to use another $m$-dim embedding for the input to the $\chi$-function, which allows the attention to be learned independent of the input embedding $\vec{s_i}$. We can certainly employ the same scheme as an alternative implementation.

\subsection{Recurrent Neural Network (CA-RNN)}
\label{subsection:carnn}
Now let us deal with some relatively complex sequence models. 
Generally speaking, in a RNN problem, the task is to infer the hidden state $c_t$ and output $y_t$ from
previous hidden state $c_{t - 1}$, previous output $y_{t - 1}$ and current input $x_t$, which can also be represented as a conditional probability $P(c_t, y_t | c_{t - 1}, y_{t - 1}, x_t) , t = 1, 2, \ldots$.

The first step to solve this problem under our context-aware setting is to factorize the conditional probability into two parts: \emph{cell state} and \emph{output}, namely
\begin{align}
P(c_t, y_t | c_{t - 1}, y_{t - 1}, x_t) = \underbrace{P(c_t | c_{t - 1}, y_{t - 1}, x_t)}_\text{cell state} \underbrace{P(y_t | c_t, c_{t - 1}, y_{t - 1}, x_t)}_\text{output}.
\end{align}
As shown in Table~\ref{table:lstm_gru}, it turns out that both the  LSTM model (\cite{lstm_original}) and the GRU model (\cite{seq_gru}) fall into solving the above sub-problems in the embedding space using different assumptions (circuits).

\begin{table}[ht]
\centering
\begin{tabular}{p{10mm} | p{60mm} | p{60mm}}
\hline
 & Cell State: $P(c_t | c_{t - 1}, y_{t - 1}, x_t)$ & Output: $P(y_t | c_t, c_{t - 1}, y_{t - 1}, x_t)$ \\ [0.5ex]
\hline
LSTM & $\bar{z}_t = \mathbf{W}_z x_t + \mathbf{R}_z y_{t - 1} + \mathbf{b}_z$ \newline
              $z_t = g(\bar{z}_t)$ \:\:\:\:\:\:\:\:\:\:\:\:\:\:\:\:\:\:\:\:\:\:\:\small{- \textbf{block input}} \newline
              $\bar{i}_t = \mathbf{W}_i x_t + \mathbf{R}_i y_{t - 1} + \mathbf{p}_i \odot c_{t - 1} + \mathbf{b}_i$ \newline
              $i_t = \sigma(\bar{i}_t)$ \:\:\:\:\:\:\:\:\:\:\:\:\:\:\:\:\:\:\:\:\:\:\:\:\:\:\:\:\:\:\small{- \textbf{input gate}} \newline
              $\bar{f}_t = \mathbf{W}_f x_t + \mathbf{R}_f y_{t - 1} + \mathbf{p}_f \odot c_{t - 1} + \mathbf{b}_f$ \newline
              $f_t = \sigma(\bar{f}_t)$ \:\:\:\:\:\:\:\:\:\:\:\:\:\:\:\:\:\:\:\:\:\:\:\:\:\:\:\small{- \textbf{forget gate}} \newline
              $c_t = z_t \odot i_t + c_{t - 1} \odot f_t$
          & $\bar{o}_t = \mathbf{W}_o x_t + \mathbf{R}_o y_{t - 1} + \mathbf{p}_o\odot c_t + \mathbf{b}_o $  \newline 
             $o_t = \sigma(\bar{o}_t)$ \:\:\:\:\:\:\:\:\:\:\:\:\:\:\:\:\:\:\:\:\:\small{- \textbf{output gate}} \newline 
             $y_t = h(c_t) \odot o_t$ \\
\hline
GRU &  $\bar{r}_t = \mathbf{W}_r x_t + \mathbf{U}_r c_{t - 1}$ \newline
             $r_t = \sigma(\bar{r}_t)$ \:\:\:\:\:\:\:\:\:\:\:\:\:\:\:\:\:\:\:\:\:\:\:\:\:\:\small{- \textbf{reset gate}}\newline
             $\bar{z}_t = \mathbf{W}_z x_t + \mathbf{U}_z c_{t - 1}$ \newline
             $z_t = \sigma(\bar{z}_t)$ \:\:\:\:\:\:\:\:\:\:\:\:\:\:\:\:\:\:\:\:\:\:\:\:\:\small{- \textbf{update gate}} \newline
             $\bar{c}_t = \tanh(\mathbf{W} x_t + \mathbf{U}(r \odot c_{t - 1}))$ \newline 
             $c_t = z_t \odot c_{t - 1} + (1 - z_t) \bar{c}_t$ 
        &   $y_t = h(\mathbf{W}_y c_t + \mathbf{b}_y)$ \\
\hline
\end{tabular}
\caption{A comparison between LSTM and GRU. Here $\sigma(\cdot)$ is the sigmoid function and $g(\cdot)$ and $h(\cdot)$ are activation functions such as $\tanh(\cdot)$.}
\label{table:lstm_gru}
\end{table}

To reformulate the RNN problem under our context-aware framework, let us first recall the EDF under an additional conditional variable $s$ in a distribution
$P(w|c, s)$ (Sec.~\ref{subsection:exponential_family}). 
In such a case, we can still consider $c$ as the context of interest, and EDF simply decomposes an embedding into one part that depends both on $c$ and $s$ (\emph{i.e.}, $\vec{v}(c, s)$) and one that does not depend on $c$ (\emph{i.e.}, $\vec{w'}(s)$):
\begin{align}
\vec{w} = \chi(w, c, s) \vec{v}(c, s) + (1 - \chi(w, c, s)) \vec{w'}(s)
\end{align}
Because $w$ is conditioned on $s$\footnote{Note that here we only consider $c$ as the context, and $s$ is a variable that both $w$ and $c$ are conditioned on.}, $\chi(w, c, s)$ actually depends only on $c$ and $s$. Hence we can simplify the above formula into
\begin{align}\label{equ:med_for_rnn}
\vec{w} = \chi(c, s) \vec{v}(c, s) + (1 - \chi(c, s)) \vec{w'}(s)
\end{align}
To shed a light on how Eq.~\ref{equ:med_for_rnn} is connected to the existing models listed in Table~\ref{table:lstm_gru}, one can consider the following correspondences
between $P(w|c, s)$ and the cell state distribution $P(c_t | c_{t - 1}, y_{t - 1}, x_t)$.
\begin{align}\label{equ:correspondence}
w &\leftrightarrow c_t  \nonumber\\
c &\leftrightarrow c_{t - 1} \nonumber\\
s &\leftrightarrow \{y_{t - 1}, x_t\}
\end{align}
Then it is not difficult to verify that all the gating equations defined in Table~\ref{table:lstm_gru} (\emph{i.e.}, input and forget gates for LSTM and reset and upgrade gates for GRU) fall into the same form $\chi(c_{t - 1}, y_{t - 1}, x_t)$, corresponding to $\chi(c, s)$ in Eq.~\ref{equ:med_for_rnn}. In the following, we explore the deep connections between Eq.~\ref{equ:med_for_rnn} and those RNN models while searching for the ``correct" version of RNN model based on the powerful EDF.

\subsubsection{Model Cell State}
Now let us consider the representation of $P(c_t | c_{t - 1}, y_{t - 1}, x_t)$ using the EDF of Eq.~\ref{equ:med_for_rnn} from the correspondences given by Eq.~\ref{equ:correspondence}.
By substituting Eq..~\ref{equ:correspondence} into Eq.~\ref{equ:med_for_rnn}, we obtain the following embedding decomposition:
\begin{align}\label{equ:carnn_cellstate}
\vec{c}_t = \chi(c_{t - 1}, y_{t - 1}, x_t) \vec{v}(c_{t - 1}, y_{t - 1}, x_t) + (1 - \chi(c_{t - 1}, y_{t - 1}, x_t)) \vec{c'}_t (y_{t - 1}, x_t),
\end{align}
where $\vec{v}(c_{t - 1}, y_{t - 1}, x_t)$ denotes the context-free part of $\vec{c}_t$ and $\vec{c'}_t (y_{t - 1}, x_t)$ denotes the context-aware part of $\vec{c}_t$.
Comparing it with the cell state equation of LSTM (\emph{i.e.}, $c_t = z_t \odot i_t + c_{t - 1} \odot f_t$), one may find a rough correspondence below
\begin{align}\label{equ:lstm_carnn_cellstate_corres}
\chi(c_{t - 1}, y_{t - 1}, x_t) &\leftrightarrow f_t \nonumber\\
\vec{v}(c_{t - 1}, y_{t - 1}, x_t) &\leftrightarrow c_{t - 1} \nonumber \\
1 - \chi(c_{t - 1}, y_{t - 1}, x_t) &\leftrightarrow i_t \nonumber \\
\vec{c'}_t (y_{t - 1}, x_t) &\leftrightarrow z_t
\end{align}
Furthermore, Eq.~\ref{equ:carnn_cellstate} provides us with some insights into the original LSTM equation.
\begin{enumerate}
    \item Under our framework, the gate functions $f_t$ and $i_t$ no longer map to a vector (same dimension as $c_t$), but rather to a scalar, which has a clear meaning: the probability that $c_t$ is independent of its context $c_{t - 1}$ conditioned on $\{y_{t - 1}, x_t\}$.
    \item The correspondences established in Eq.~\ref{equ:lstm_carnn_cellstate_corres} suggest that the forget gate and the input gate are coupled and should sum to one ($f_t + i_t = 1$). This may not make good sense in the LSTM model since it amounts to choosing either $c_{t - 1}$ (previous state) or $z_t$ (current input). But if one looks at the counterpart of $c_{t - 1}$, namely $\vec{v}(c_{t - 1}, y_{t - 1}, x_t)$, it already includes all the current ($x_t$) and past ($c_{t - 1}, y_{t - 1}$) information. Therefore, we can safely drop one gate in LSTM as well as reduce their dimension to one.
    \item Both $\vec{c'}_t (y_{t - 1}, x_t)$ in our framework and $z_t$ in LSTM share the same form: the embedding of $c_t$ without the information from $c_{t - 1}$. However, we call $\vec{c'}_t (y_{t - 1}, x_t)$ the context-sensitive vector as when interacted with the context vector in the log-linear model, it changes with $c_{t - 1}$. Hence our framework endows new meanings to the terms in LSTM that makes mathematical sense.
\end{enumerate}

Similarly for the GRU model listed in Table~\ref{table:lstm_gru}, we can also connect it to our context-aware framework by considering the following correspondences:
\begin{align}
\chi(c_{t - 1}, y_{t - 1}, x_t) &\leftrightarrow  z_t  \nonumber\\
\vec{v}(c_{t - 1}, y_{t - 1}, x_t) &\leftrightarrow c_{t - 1} \nonumber \\
\vec{c'}_t (y_{t - 1}, x_t) &\leftrightarrow \bar{c}_t
\end{align}
At a high level, the GRU equation $c_t = z_t \odot c_{t - 1} + (1 - z_t) \bar{c}_t$ resembles our EDF in appearance. 
However, $\bar{c}_t$ is not independent of $c_{t - 1}$, contrary to our assumption.
To remedy this, the GRU model employs the reset gate $r_t$ to potentially filter out information from $c_{t - 1}$, which is not necessary in our case.
Also GRU model assume the gate functions are vectors rather than scalars.

Given the general form of context-aware cell state equation~\ref{equ:carnn_cellstate}, one can choose any activation functions for its implementation.
Following similar notations used in LSTM equations, we can write down a new implementation of the RNN model, namely CA-RNN, as follows.
\begin{align}\label{equ:carnn_cell_state_impl}
\chi(c_{t - 1}, y_{t - 1}, x_t): &\:\:\:\:\:\:\: f_t = \sigma (\mathbf{v}_f^T  \vec{c}_{t - 1} + \mathbf{w}_f^T  \vec{x}_t + \mathbf{u}_f^T \vec{y}_{t - 1} + b_f) \nonumber \\
\vec{v}(c_{t - 1}, y_{t - 1}, x_t): &\:\:\:\:\:\:\: \vec{v}_t = h_v (\mathbf{W}_v \vec{x}_t + \mathbf{U}_v \vec{y}_{t - 1} + \mathbf{p}_v \odot \vec{c}_{t - 1} + \mathbf{b}_v)  \nonumber \\
\vec{c'}_t(y_{t - 1}, x_t): &\:\:\:\:\:\:\: \vec{c'}_t = h_c (\mathbf{W}_c \vec{x}_t + \mathbf{U}_c \vec{y}_{t - 1} + \mathbf{b}_c) \nonumber \\
\vec{c}_t: &\:\:\:\:\:\:\:  \vec{c}_t = f_t \vec{v}_t + (1 - f_t) \vec{c'}_t
\end{align}
Here $\sigma$ is the sigmoid function with parameters $\{\mathbf{v}_f\in \mathbb{R}^{dim(\vec{c}_t)}, \mathbf{w}_f\in \mathbb{R}^{dim(\vec{x}_t)}, \mathbf{u}_f\in \mathbb{R}^{dim(\vec{y}_t)}, b_f\in \mathbb{R}\}$, $h_v$ and $h_g$ are activation functions (\emph{e.g.}, $\tanh$, relu), with parameters $\{\mathbf{W}_v\in  \mathbb{R}^{dim(\vec{c}_t)\times dim(\vec{x}_t)}, \\ \mathbf{U}_v\in \mathbb{R}^{dim(\vec{c}_t)\times dim(\vec{y}_t)}, \mathbf{b}_v \in \mathbb{R}^{dim(\vec{c}_t)}\}$
and $\{\mathbf{W}_c\in  \mathbb{R}^{dim(\vec{c}_t)\times dim(\vec{x}_t)}, \mathbf{U}_c\in \mathbb{R}^{dim(\vec{c}_t)\times dim(\vec{y}_t)}, \mathbf{b}_c \in \mathbb{R}^{dim(\vec{c}_t)}\}$.

\subsubsection{Model Output}
In Table~\ref{table:lstm_gru}, the equations for the output $y_t$ look much simpler than those for the cell state for both LSTM and GRU models.
However, in our case, it is actually more complex as it models a distribution involving more random variables: $P(y_t | c_t, c_{t - 1}, y_{t - 1}, x_t)$.
Let us first write down the corresponding EDF, assuming the context of interest is $c_t$:
\begin{align}\label{equ:carnn_output}
\vec{y}_t = \chi(c_t, c_{t - 1}, y_{t - 1}, x_t) v(c_t, c_{t - 1}, y_{t - 1}, x_t)  + (1 - \chi(c_t, c_{t - 1}, y_{t - 1}, x_t)) \vec{y'}_t (c_{t - 1}, y_{t - 1}, x_t)
\end{align}
From the output equations for LSTM ($y_t = h(c_t) \odot o_t$) and  GRU ($y_t = h(\mathbf{W}_y c_t + \mathbf{b}_y)$), one can see that both only consider the context-free part of Eq.~\ref{equ:carnn_output}.
The LSTM model uses a single output gate $o_t$, meaning the output would be turned off (\emph{i.e.}, $y_t \rightarrow 0$), if $o_t \rightarrow 0$.
However, when LSTM is trained (usually $y_t$ is used as the activation of a softmax layer), the output can hardly be matched to a near zero vector target.
Hence $o_t$ as an output gate may not really have any real affect.
In contrast, our formulation of Eq.~\ref{equ:carnn_output} allows the (soft) switch between two different modes: context free and context sensitive, which models a more realistic situation.
Note that Eq.~\ref{equ:carnn_output} also subsumes the GRU output equation as its special case if we set $\chi(c_t, c_{t - 1}, y_{t - 1}, x_t) = 1$ and  $v(c_t, c_{t - 1}, y_{t - 1}, x_t) = h(\mathbf{W}_y c_t + \mathbf{b}_y)$.

Again, following the notation of LSTM equations, we can write down one implementation of our CA-RNN model for the output as follows.
\begin{align}\label{equ:carnn_output_impl}
\chi(c_t, c_{t - 1}, y_{t - 1}, x_t): &\:\:\:\:\:\:\: o_t = \sigma(\mathbf{z}_o^T \vec{c}_{t} + \mathbf{v}_o^T \vec{c}_{t - 1} + \mathbf{w}_o^T \vec{x}_t + \mathbf{u}_o^T \vec{y}_{t - 1} + b_o) \nonumber \\
v(c_t, c_{t - 1}, y_{t - 1}, x_t): &\:\:\:\:\:\:\:  \vec{v_t^o} = h_v^o (\mathbf{Z}_v \vec{c}_t + \mathbf{V}_v^o \vec{c}_{t - 1} + \mathbf{W}_v^o \vec{x}_t + \mathbf{U}_v^o \vec{y}_{t - 1} + \mathbf{b}_v^o) \nonumber \\
\vec{y'}_t (c_{t - 1}, y_{t - 1}, x_t): &\:\:\:\:\:\:\: \vec{c_t^o} = h_c^o (\mathbf{V}_c^o \vec{c}_{t - 1} + \mathbf{W}_c^o \vec{x}_t + \mathbf{U}_c^o \vec{y}_{t - 1} + \mathbf{b}_c^o) \nonumber \\
\vec{y}_t: &\:\:\:\:\:\:\: \vec{y}_t = o_t \vec{v_t^o} + (1 - o_t) \vec{c_t^o}
\end{align}
Here the superscript $^o$ indicates that the variables are for the output, and
$\{\mathbf{z}_o\in \mathbb{R}^{dim(\vec{c}_t)}, \mathbf{v}_o\in \mathbb{R}^{dim(\vec{c}_t)}, \mathbf{w}_o\in \mathbb{R}^{dim(\vec{x}_t)}, \mathbf{u}_o\in \mathbb{R}^{dim(\vec{y}_t)}, b_o\in \mathbb{R}\}$, 
 $\{\mathbf{Z}_v\in \mathbb{R}^{dim(\vec{c}_t)\times dim(\vec{c}_t)},  \mathbf{V}_v^o \in \mathbb{R}^{dim(\vec{c}_t)\times dim(\vec{c}_t)}, \mathbf{W}_v^o\in  \mathbb{R}^{|c|\times dim(\vec{x}_t)}, \mathbf{U}_v^o\in \mathbb{R}^{dim(\vec{c}_t)\times dim(\vec{y}_t)}, \mathbf{b}_v^o \in \mathbb{R}^{dim(\vec{c}_t)}\}$,
 $\{\mathbf{V}_c^o \in \mathbb{R}^{dim(\vec{c}_t)\times dim(\vec{c}_t)}, \mathbf{W}_c^o\in  \mathbb{R}^{dim(\vec{c}_t)\times dim(\vec{x}_t)}, \\ \mathbf{U}_c^o\in \mathbb{R}^{dim(\vec{c}_t)\times dim(\vec{y}_t)}, \mathbf{b}_v^o \in \mathbb{R}^{dim(\vec{c}_t)}\}$.

\begin{figure}[t]
\centering
\begin{tabular}{c}
    \mbox{\epsfig{figure= 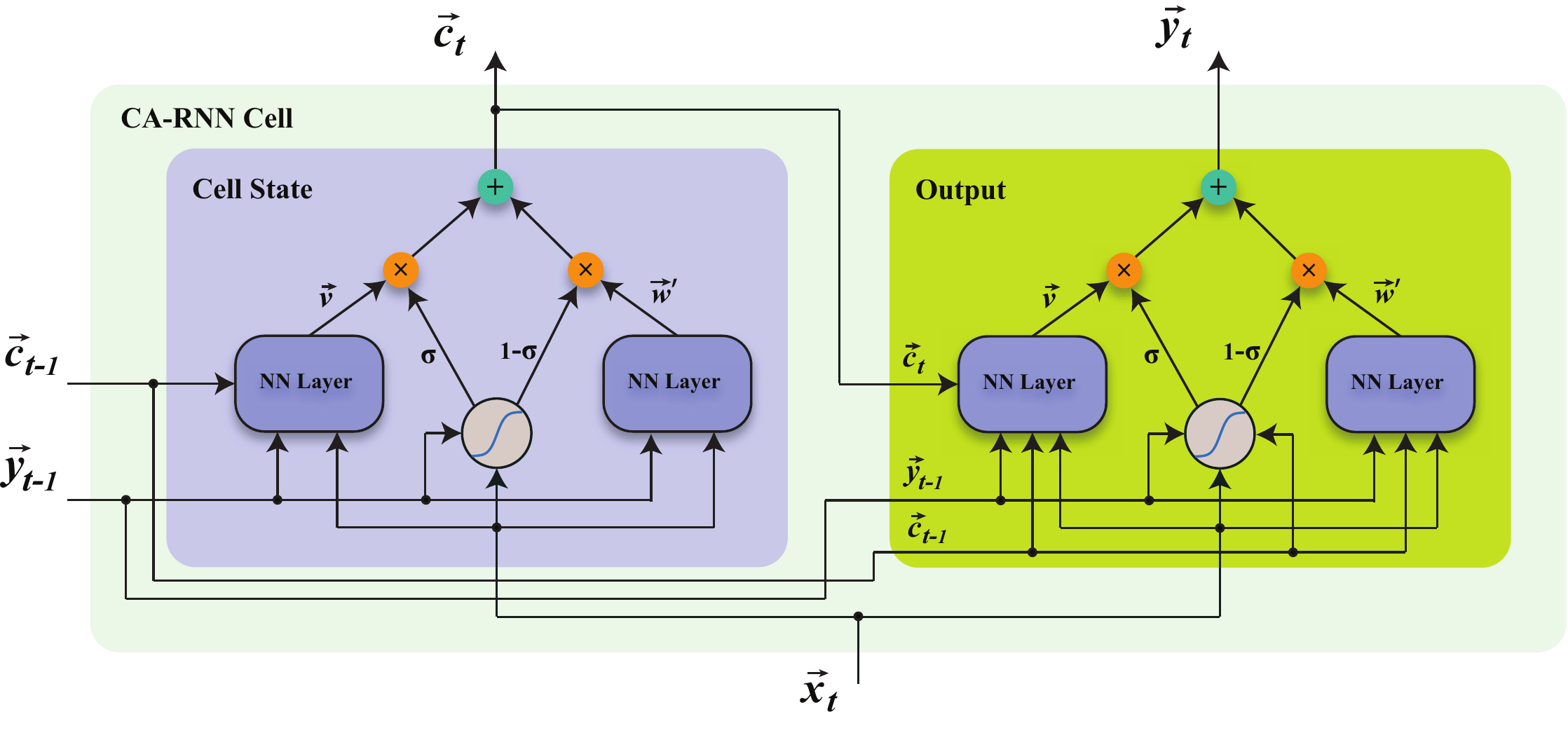, height = 6.8cm}}
\end{tabular}
\caption{The architecture of CA-RNN model.} \label{fig:ca_rnn}
\end{figure}

In sum, the architecture of our CA-RNN model can be illustrated in Fig.~\ref{fig:ca_rnn}.
Comparing it with the circuits of the LSTM model~\cite{lstm_space_odyssey}, the number of gates is reduced from three to two.
Its symmetric structure also makes the model more comprehensible.

Finally, note that our model invites many possible implementations beyond Eq.~\ref{equ:carnn_cell_state_impl} and Eq.~\ref{equ:carnn_output_impl}.
For example, one can make a simpler assumption that $P(y_t | c_t, c_{t-1},y_{t-1},x_t) = P(y_t | c_t, y_{t-1}, x_t)$ as $\vec{y}_{t-1}$ is a function of $\vec{c}_{t - 1}$ and other variables.

The model complexity (\emph{i.e.}, number of parameters) of CA-RNN depends on the implementation so there is no fair comparison with other models.
Nevertheless, the gate functions (\emph{i.e.}, $\chi$-function) of CA-RNN reduces the output dimension from $n$ to $1$, where $n$ is the dimension of cell state.

\subsection{Generic Neural Network Layer (CA-NN)}
\label{subsection:cann}
Next, let us design a generic neural network layer based on the EDF.
An NN layer simply takes the output from a lower level layer as input, and output an n-dimensional embedding.
If we let $c$ denote the input and $w$ denote the output, an NN layer can be consider as modeling the probability $P(w|c)$.
Hence, from our EDF of Eq.~\ref{equ:magic_decomposition_formula}, the embedding (output) $\vec{w}$ can be represented as
\begin{align}\label{equ:cann}
\vec{w} = \chi(\vec{w}_0, \vec{c}) v(\vec{c})  + (1 - \chi(\vec{w}_0, \vec{c})) \vec{w}_0.
\end{align}
Here $v(\vec{c})$ is a function of the input signal $\vec{c}$ that maps to the context-free part of the output;
$\vec{w}_0$ is a vector representing the ``default value" of the output when the input is irrelevant;
The $\chi$-function is determined both by the context and the default value, and because the default value $\vec{w}_0$ is a constant after training, it is just a function of $\vec{c}$.
Hence we can simplify Eq.~\ref{equ:cann} as
\begin{align}\label{equ:cann2}
\vec{w} = \chi(\vec{c}) v(\vec{c})  + (1 - \chi(\vec{c})) \vec{w}_0.
\end{align}
Compared to a traditional NN layer, which simply assumes $\chi(\vec{c}) = 1$, our model allows the input to be conditionally blocked based on its value\footnote{Note that this model is different from our single feature embedding model (Fig.~\ref{fig:ca_bsfe} (a)) which seeks to computes an embedding with a global constant context.}.

\begin{figure}[t]
\centering
\begin{tabular}{ccc}
    \mbox{\epsfig{figure= 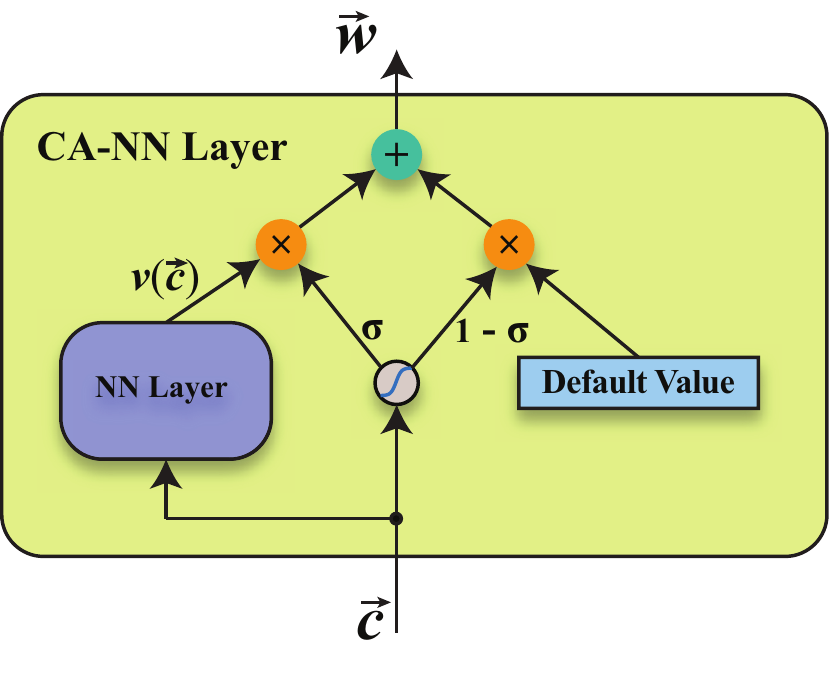, height = 5cm}}  
    & \hspace{5mm} \mbox{\epsfig{figure= 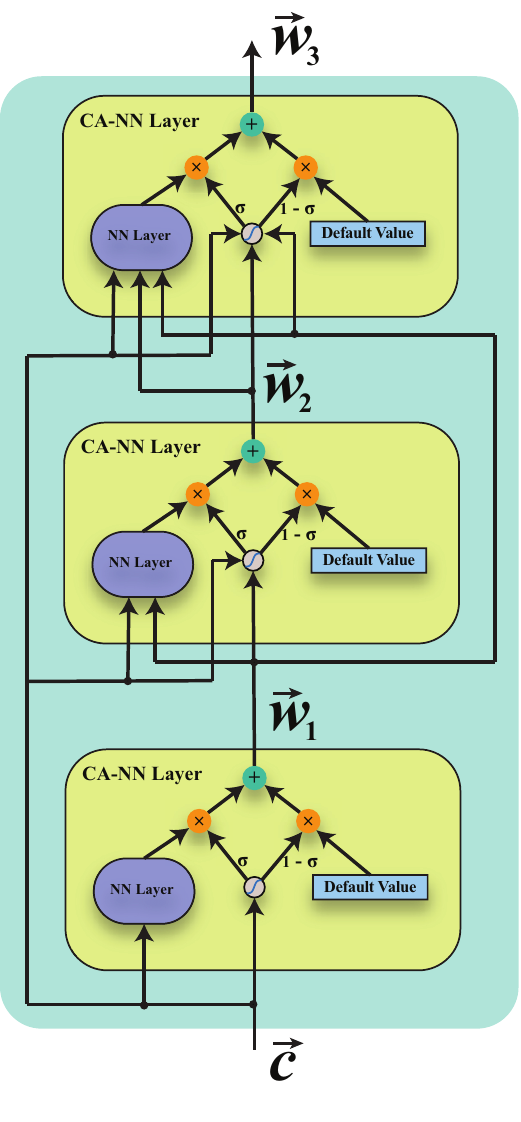, height = 5cm}} 
    & \hspace{5mm} \mbox{\epsfig{figure= 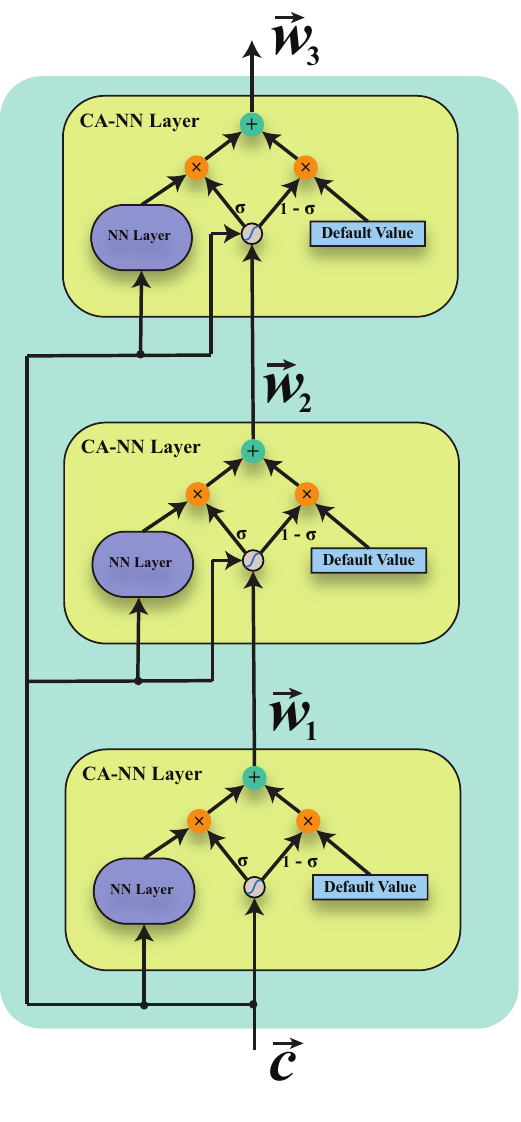, height = 5cm}}\\
    {\scriptsize (a) Single CA-NN layer.} & \hspace{5mm}  {\scriptsize (b) Three layer CA-RES.} & \hspace{5mm}  {\scriptsize (c) Simplified version.}
\end{tabular}
\caption{\small The architecture for a single layer CA-NN (a), its multi-layer extension CA-RES (b) and a simplified version of $3$-layer CA-RES (c) that resembles ResNet.} \label{fig:ca_nn}
\end{figure}

We argue that our new structure of an NN layer make better neurological sense than the traditional NN layers.
A biological neuron can be considered as a nonlinear dynamical system on its membrane potential~\cite{book_dynamical_systems_neuro}.
The communication among neurons are either based on their fire patterns (resonate-and-fire) or on a single spike (integrate-and-fire).
Moreover, a neuron's fire pattern may change with the input current or context, \emph{e.g.}, from excitable to oscillatory, a phenomena known as \emph{bifurcation}.
Such a change of fire pattern is partially enabled by the \emph{gates} on the neuron's membrane which control the conductances of the surrounding ions (\emph{e.g.}, Na$^+$).
Traditional NN layers under the perceptron framework (\emph{i.e.}, linear mapping + nonlinear activation) fail to address such bistable states of neurons, and assumes a neuron is a simple excitable system with a fixed threshold, which is disputed by evidences in neuroscience (Chapers 1.1.2 and 7.2.4 in~\cite{book_dynamical_systems_neuro}). Our CA-NN model (Fig.~\ref{fig:ca_nn}), in contrast, shares more resemblance with a real neuron and is able to model its bistable property.
For instance, the probability of the ion channel gates to be open or close are determined by the current potential of the membrane represented by a neuron's extracellular state, which resembles our $\chi$-function.
Here the extracellular (external) state is modeled by  $\vec{c}$, \emph{i.e.} the contextual input.

To implement a CA-NN model, one can choose any existing NN layer for $v(\vec{c})$, and the $\chi$-function can just be modeled using the sigmoid function, \emph{i.e.},
\begin{align}\label{equ:sigma_function}
\chi(\vec{c}) = \sigma( \mathbf{v}^T\vec{c} + b),
\end{align}
where $\{\mathbf{v} \in \mathbb{R}^{dim(\vec{c})}, b \in \mathbb{R}\}$ and $\sigma(\cdot)$ is the sigmoid function.
Fig.~\ref{fig:ca_nn}(a) shows the architecture of our CA-NN model.

Our CA-NN framework invites many possible implementations, depending on how the context is chosen.
In the rest of this paper, we discuss its two straightforward extensions,  one resembles the residual network (ResNet)~\cite{resnet_2015} and another the convolutional neural network (CNN)~\cite{cnn_nips1990}.

\subsubsection{Context Aware Convolutional Neural Network (CA-CNN)}
\label{subsubsection:cacnn}

\begin{figure}[t]
\centering
\begin{tabular}{c}
    \mbox{\epsfig{figure= 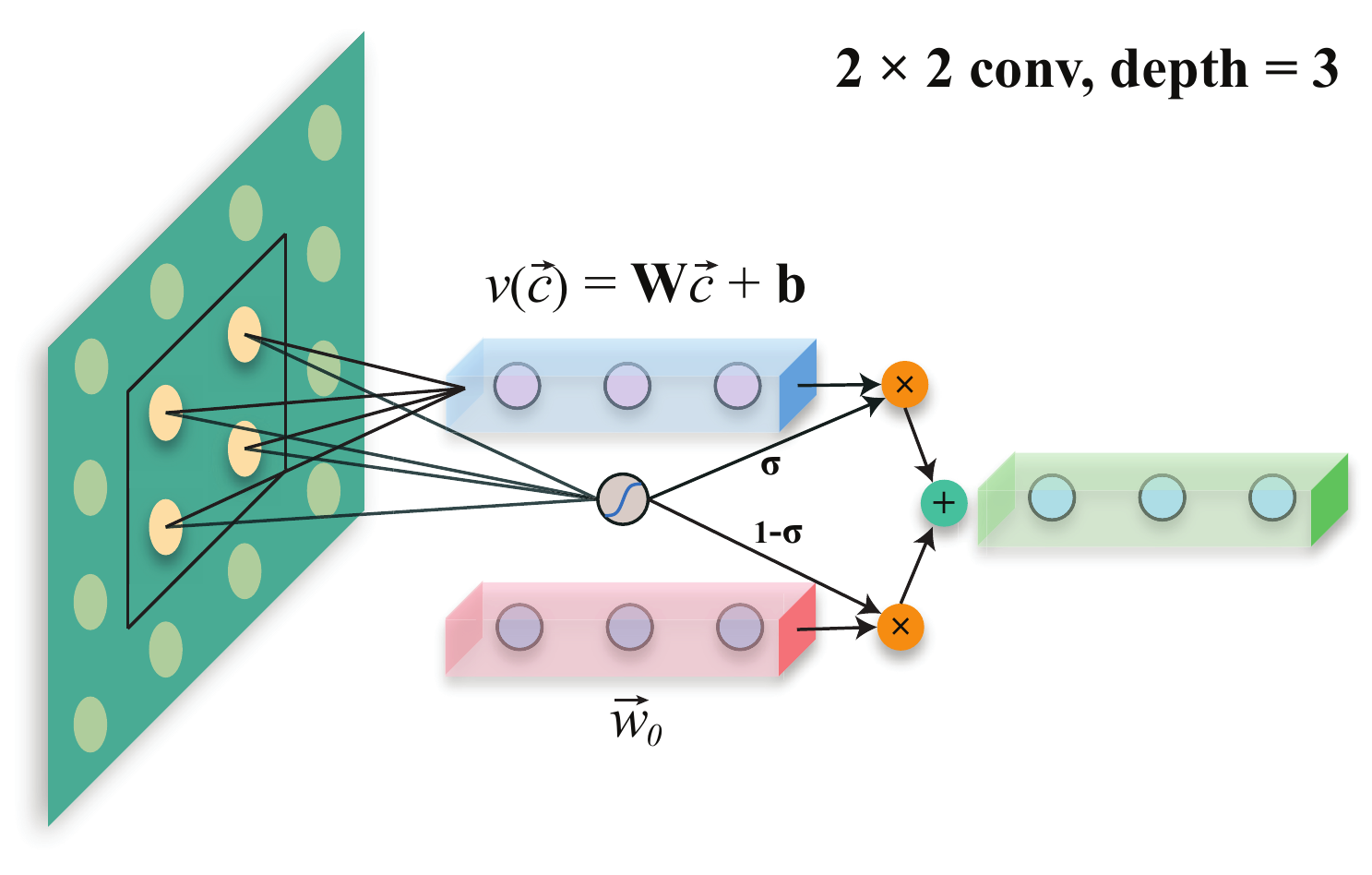, height = 4.8cm}} 
\end{tabular}
\caption{The architecture for context aware convolutional neural network (CA-CNN).} \label{fig:ca_cnn}
\end{figure}

It is straightforward to modify our CA-NN model to serve as a CNN model~\cite{cnn_nips1990}, by just applying it on a patch of the input as filters.
For example, in the linear case, we can define $v(\vec{c})$ in Eq.~\ref{equ:cann2} as a linear transformation
\begin{align}
v(\vec{c}) = \mathbf{W} \vec{c} + \mathbf{b},
\end{align}
where $ \mathbf{W} \in \mathbb{R}^{n\times dim(\vec{c})}, \mathbf{b}\in \mathbb{R}^n$ and $n$ is the depth of the output.
The presence of $\chi$-function and $\vec{w}_0\in\mathbb{R}^n$ in Eq.~\ref{equ:cann2} allows irrelevant input (\emph{e.g.}, background in an image) to be detected and replaced with a constant value.
Therefore, CA-CNN model can be considered as convolution (CNN) with attention.
Fig.~\ref{fig:ca_cnn} illustrates the architecture of a basic CA-CNN layer from a single layer CA-NN model.
In a multilayer convolutional network, one can certainly employ the CA-RES architecture (Sec.~\ref{subsubsection:cares}) as the convolution kernel.

\subsubsection{Context Aware Residual Network (CA-RES)}
\label{subsubsection:cares}
Lastly, let us look at how we can build a multilayer network using CA-NN layers as basic building blocks.
Simply stacking multiple CA-NN layers on top of one another and using the output from lower layer as context may not be a good idea because once the input to one layer is blocked, the information can no longer be passed to deeper layers~\footnote{Although this can be overcome by allowing the default value vector to change with input $\vec{c}$ as proposed in~\cite{image_training_very_deep_networks}, it falls out of our framework and we are unable to find a sound theoretical explanation for such a heuristic.}. 
However, if we look at the multiple layer network problem in a probabilistic setting, it can be actually formulated as solving the distribution
$P(w_1, w_2, \ldots, w_n | c)$ where $w_i, i = 1, 2, \ldots, n$ are the output from different layers and $n > 1$ is the total layer number, which can be factorized as
\begin{align}
P(w_1, w_2, \ldots, w_n | c) = P(w_1|c) P(w_2 | c, w_1) \ldots P(w_n | c, w_{n-1}, \ldots, w_{1}) , n > 1
\end{align}
A traditional muti-layer network simply assumes $P(w_l | c, w_{l-1}, \ldots, w_{1}) = P(w_l | w_{l-1}), l > 1$, \emph{i.e.}, the output at layer $l$ only depends on the output of lower layer $l - 1$\footnote{For comparison, a deep belief network~\cite{theory_dbn} rather considers a deep network in the opposite direction: from deepest layer $n$ to the observation $c$, which is represented by $P(c, w_1, w_2, \ldots, w_n) = P(c|w^1)\prod_{k=1}^{n - 2} P(w^k|w^{k+1})P(w^{n-1}, w^n)$. It assumes the states of $c$ and $w_i, i = 1,\ldots, n$ are represented by the binary states of several neurons, whereas we assume the state of each layer (a single neuron) can be represented in a continuous embedding space.}.
To construct a multi-layer CA-NN network, we have to use the full model $P(w_l | c, w_{l-1}, \ldots, w_{1})$ for each layer $l$ and by assuming the context to be $\{c, w_{l-1}, \ldots, w_{1}\}$ in Eq.~\ref{equ:cann2}, we have the following decomposition:
\begin{align}\label{equ:cares_decomposition}
\vec{w}_l = \chi(\vec{c}, \vec{w}_{l-1}, \ldots, \vec{w}_1) \vec{v}_l(\vec{c}, \vec{w}_{l-1}, \ldots, \vec{w}_1)  + (1 - \chi(\vec{c}, \vec{w}_{l-1}, \ldots, \vec{w}_1)) \vec{w}_{l0}, l > 1.
\end{align}
Here $\vec{w}_{l0}$ is the default value vector at level $l$. $\vec{w}_{l-1}, \ldots, \vec{w}_1$ are the output from previous layers.
$\vec{c}, \vec{w}_{l-1}, \ldots, \vec{w}_1)$ can be any activation function.
Fig.~\ref{fig:ca_nn}(b) shows an example of $3$-layer CA-NN network.

The number of parameters for the full version of our multi-layer NN model would grow squarely with the number of layers, which may not be desirable in a deep neural network.
One can certainly make simplified assumptions, \emph{e.g.}, $\chi(\vec{c}, \vec{w}_{l-1}, \ldots, \vec{w}_1) = \chi(\vec{c}, \vec{w}_{l-1}, \ldots, \vec{w}_{l - k}), k > 1$.
In an overly simplified case, if we assume $\vec{v}_l(\vec{c}, \vec{w}_{l-1}, \ldots, \vec{w}_1) =  \vec{v}_l(\vec{c})$ (Fig.~\ref{fig:ca_nn}(c)) and $\chi(\vec{c}, \vec{w}_{l-1}, \ldots, \vec{w}_1) = 1$,it becomes the popular residual network (ResNet)~\cite{resnet_2015}.
In essence, both our multi-layer CA-NN and ResNet are able to skip underlying nonlinear NN layers, hence we also name our muti-layer CA-NN models CA-RES\footnote{Despite their resemblance in structure, ResNet and CA-RES are derived from very different assumptions. We find the assumptions behind ResNet, decomposing a mapping $\mathcal{H}(x)$ into an identify mapping $x$ and a residual mapping $\mathcal{F} = \mathcal{H}(x) - x$, may not fit well to our context-aware framework.}.

In fact, our construction of multi-layer network provides some insights into existing models and we argue that it also better resembles a real neural network.
First, contrary to a superficial view that an artificial neural network (ANN) is simply a nonlinear mapping from input to output, it actually models the internal states $\{w_i | i = 1, \ldots, n\}$ from given context $c$. 
Also each internal state $w_i$ can be represented in the embedding space if we assume the distribution $P(w_l | c, w_{l-1}, \ldots, w_{1})$ belongs to the exponential family.

Second, it is known that a real neural network evolves from almost fully connected to sparsely connected as a human grows from childhood to adulthood.
Our CA-RES architecture is actually able to model this dynamic process in principle.
The role of default value in each layer has the effect of blocking certain input from lower layers to all its upper layers, therefore it can reduce the effective connections among different neurons as more training data are seen.
Note that the ResNet also achieves similar effect, but without a gate to explicitly switch on/off a layer therefore not adaptive.

\subsection{Connections to other models}
Before moving on to the experimental results, we would like to conclude this section by discussing how our context-aware framework can be further connected to other machine learning models. 

It is possible to extend our framework to multiple contexts, in which case the $\chi$-function can be modeled as a softmax function.
For example, the mixture of experts model~\cite{moe_iclr2017} can be improved by accounting for the situation that no expert's output is relevant.
However, this is out of the scope of this paper.

Our CA-RES architecture can also be a good candidate for multitask learning (\emph{e.g.}, ~\cite{multitask_unified,multitask_survey}). By wiring any of the output $w_l$ from different layers (Sec.~\ref{subsubsection:cares}) to fit different targets (\emph{e.g.}, \cite{cnn_lenet15}), in principle, the $\chi$-functions in each layer could play the role of switching to different networks for different tasks.

\section{Experimental Results}
\label{section:experiment}
Before delving into the details of the experiments for each of our context-aware models proposed in Sec.~\ref{section:new_models}, it is helpful to compare them side-by-side to highlight their differences and relationships (Table.~\ref{table:allmodels}). 
\begin{itemize}
\item Except for the CA-SEM model, all the other ones can be implemented as a neural network layer such that the optimization can be done via back propagation algorithm.
\item Both CA-SEM and CA-BSFE assume a global context. Therefore the trained result is only an embedding dictionary.
\item CA-RES is the multi-layer extension of CA-NN and CA-CNN is the single layer extension of CA-NN to convolutional network. CA-RES can also be extended to convolutional network.
\end{itemize}

\begin{table}[ht]
\centering
\begin{tabular}{p{17mm} | p{75mm} | p{45mm}}
\hline
Model & Form of EDF applied & Variables/functions to be learned \\ [0.5ex]
\hline
CA-SEM (Sec.~\ref{subsection:casem}) & $\vec{w} = \tilde{\chi}(w) \vec{v}_0 + (1 - \tilde{\chi}(w)) \vec{w'}$ & $\vec{v}_0, \tilde{\chi}(w), \vec{w'}, w\in \mathcal{W}$\\
\hline
CA-BSFE (Sec.~\ref{subsection:cabsfe}) & $\vec{s} = \vec{v}_0 \sum_{w \in s} \tilde{\chi}(w; \theta) +  \sum_{w\in s} (1 - \tilde{\chi}(w; \theta)) \vec{w}$  &  $\vec{v}_0, \tilde{\chi}(\cdot; \theta),  \vec{w}, w\in\mathcal{W}$\\
\hline
CA-ATT (Sec.~\ref{subsection:caatt}) & $\vec{s} = \vec{v}(\vec{c}) \sum_{i = 1}^n \chi(\vec{s}_i, \vec{c}) + \sum_{i = 1}^n (1 - \chi(\vec{s}_i, \vec{c})) \vec{s_i}$  
& $v(\cdot), \chi(\cdot, \cdot)$ \\
\hline
CA-RNN (Sec.~\ref{subsection:carnn})& $\vec{w} = \chi(c, s) \vec{v}(c, s) + (1 - \chi(c, s)) \vec{w'}(s)$  & $\chi(\cdot, \cdot), \vec{v}(\cdot, \cdot), \vec{w'}(\cdot)$ \\
\hline
CA-NN (Sec.~\ref{subsection:cann}) & $\vec{w} = \chi(\vec{c}) v(\vec{c})  + (1 - \chi(\vec{c})) \vec{w}_0$  & $\chi(\cdot), v(\cdot), \vec{w}_0$ \\
\hline
CA-CNN (Sec.~\ref{subsubsection:cacnn}) & $\vec{w} = \chi(\vec{c}) v(\vec{c})  + (1 - \chi(\vec{c})) \vec{w}_0$   & $\chi(\cdot), v(\cdot), \vec{w}_0$ \\
\hline
CA-RES (Sec.~\ref{subsubsection:cares})& $\vec{w}_l = \chi(\vec{c}, \vec{w}_{l-1}, \ldots, \vec{w}_1) \vec{v}_l(\vec{c}, \vec{w}_{l-1}, \ldots, \vec{w}_1)  + (1 - \chi(\vec{c}, \vec{w}_{l-1}, \ldots, \vec{w}_1)) \vec{w}_{l0}$  & $\chi(\cdot, \ldots), v(\cdot, \ldots), \vec{w}_{l0}$ \\
\hline
\end{tabular}
\caption{Comparison among all the models proposed in Sec.~\ref{section:new_models}.}
\label{table:allmodels}
\end{table}

A neural network model usually involves many layers of various types, and its overall performance is determined by not only the choice of individual layers, but also other hyper-parameters like initialization, batch size, learning rate, etc.
The focus of this paper is not on the design of complex neural network architectures to achieve state-of-the-art performance on benchmark datasets (\emph{e.g.}~\cite{cnn_imagenet,attention_all_you_need}), but rather on the superiority of our proposed layers over their ``traditional" counterparts, which can be validated by replacing the layers in various existing models with ours while keeping all other parameters unchanged to see the performance gain. 

\textbf{Dictionary-free embedding}: another practical concern that can affect the performance of a model involving sparse features is the selection of a dictionary/vocabulary for embedding and all other sparse values are mapped to a special ``unknown" or ``oov" embedding. Such a dictionary is often selected based on the frequencies of the sparse features shown in the training data. We can eliminate this model tuning option simply by allowing every input features be mapped to its own embedding.
Therefore in our experiments there is no such concept of ``dictionary" for us to process.

All the neural network models are implemented using the Tensorflow library~\cite{tensorflow} that supports distributed computation. 
For the public dataset tested in this paper, we usually deploy $2 - 100$ workers without GPU acceleration.
Below are general descriptions of the datasets that we use to evaluate our models.

The \textbf{SemEval} semantic textual similarity (STS) dataset~\cite{dataset_sts2017} consists of $5749$ training and $1379$ testing examples for comparing the semantic similarity (a number in $[0, 5]$) between two sentences, and the data is grouped by year from $2012$ to $2017$.
The data contains both a testing file and a training file labeled by year.

The \textbf{SICK-2014} dataset~\cite{dataset_sick14} is designed to capture semantic similarities among sentences based on common knowledge, as opposed to relying on specific domain knowledge.
Its annotations for each sentence pair include both a similarity score within $[0, 5]$ and an entailment label: neutral, entailment, contradiction.
The data consists of $4500$ sentence pairs for training and $4927$ pairs for testing.

The \textbf{SST} (Stanford Sentiment Treebank)~\cite{dataset_treebank2013} is a dataset of movie reviews which was annotated for 5 levels of sentiment: strong negative, negative, neutral, positive, and strong positive.  Training $67349$ and testing $1821$ examples.

The \textbf{IMDB} movie review dataset~\cite{imdb_dataset} contains 50,000 movie reviews with an even number of positive and negative labels.
The data is split into 25,000 for training and 25,000 for testing.

The \textbf{MNIST} dataset~\cite{cnn_maxpooling} contains 60,000 training and 10,000 testing grey-level images each normalized to size $28\times 28$.

The \textbf{CIFAR-10} dataset~\cite{dataset_cifar} contains 60,000 color images in $10$ classes, with 6000 images per class and each image is normalized to size $32\times 32$.
The data is divided into 50,000 for training and 10,000 for testing. 

The \textbf{WMT} 2016 dataset contains machine translation tasks among a wide range of languages. Its training data size is sufficient large to be considered useful for real-world applications. 
For example, its English-German dataset consists of 4,519,003 sentence pairs.
Its testing data is relatively small, around 2,000 for each year ($2009-2016$).

\subsection{Results for CA-SEM}
Recall that our CA-SEM model solves the sentence re-embedding problem by a block-coordinate algorithm that minimizes Eq.~\ref{equ:sem_reembedding_regroup}.
Once we have solved the global context vector $\vec{v}_0$, all other variables, namely $\sigma(w)$ and $\vec{w'}$, $\forall w \in \mathcal{W}$, can be obtained via a closed form solution (Sec.~\ref{subsection:casem}). Therefore $\vec{v}_0$ contains all the information we need for a re-embedding which makes model serving fast and simple.

One very useful application of CA-SEM is \emph{text search}: find closest text snippets in a pre-defined corpus (\emph{e.g.}, articles in news website or text in books) from given text query in a natural language form\footnote{One advantage of this bag-of-words embedding over sequence model based embedding is speed in serving, since computing an embedding vector for a sentence only requires a few lookups and fast re-embedding computation.}.
In such a case, one can pre-compute the embeddings of all the sentences in the given corpus from any pre-trained word embedding based on our re-embedding algorithm.
In our experiments, our algorithms easily outperform that of~\cite{sem_arora2017} on all datasets we tried (SemEval, SST, and several Google's internal datasets) by a significant large margin.

\subsubsection{Implementation details}
We implemented both the algorithm described in~\cite{sem_arora2017} and our method using C++. 
To handle large dataset (\emph{e.g.}, $>100$M sentences), we also implemented a distributed version of our algorithm using standard distributed programming model that is similar to MapReduce~\cite{mapreduce}. 
In addition, the original embedding dictionary may not be restricted to single words (\emph{e.g.}, ``united", ``states") but also contains phrases (\emph{e.g.}, ``united states").
In such a case, there exists multiple ways of segmenting a sentence into phrases and we select the segmentation with least number of phrases and OOV tokens (\emph{e.g.}, to segment ``united state" with a dictionary of \{``united", ``states", ``united state"\}, a segmentation of  \{``united state"\} is preferrable than  \{``united", ``states"\}), which can be implemented by a fast dynamic programming algorithm. 

In our implementation of Alg.~\ref{alg:ca_sem_algorithm}, $\vec{v}_0$ is initialized with the first principal component computed from all the word/phrase embeddings from all the sentences in the training corpus. Then the energy minimization can be optimized iteratively until the energy of Eq.~\ref{equ:sem_reembedding_regroup} no longer decreases between two iterations.
We observed that for small datasets like SemEval, the energy can keep decreasing after hundreds of iterations (Fig.~\ref{fig:sts_training}(a), but for large datasets with more than $10$M sentences, the algorithm usually stops within $10 - 20$ iterations.
As for runtime, it takes less than $2$ seconds to finish $100$ iterations for small datasets like SemEval on a machine equipped with Intel(R) Xeon(R) $3.50$GHz CPU with $12$ cores,
and for large datasets it usually takes less than $1$ hour with $10$ distributed machines. 

Both the SemEval-STS and SICK datasets contain training instances that come with pairs of sentences.
Whereas sentence reembedding only needs instances of sentences that reflect the distribution of interested text corpus.
Hence we simply assume each training pair gives us two independent sentences, which then doubles the size of training instances from the original dataset.

For the pre-trained embedding dictionary as input to our algorithm, we employ the publicly available GloVe embedding~\cite{emb_pennington2014glove} with dimension size $200$.
To evaluate the similarity between two sentences, we compute the cosine distance between their sentence embeddings.
Similar to~\cite{sem_arora2017}, we compare with the ground truth score based on the Pearson's coefficient $\times$ $100$ (Pearson's $r \times 100$).

\subsubsection{SemEval-sts dataset}
\begin{figure}[t]
\centering
\begin{tabular}{cc}
    \mbox{\epsfig{figure= 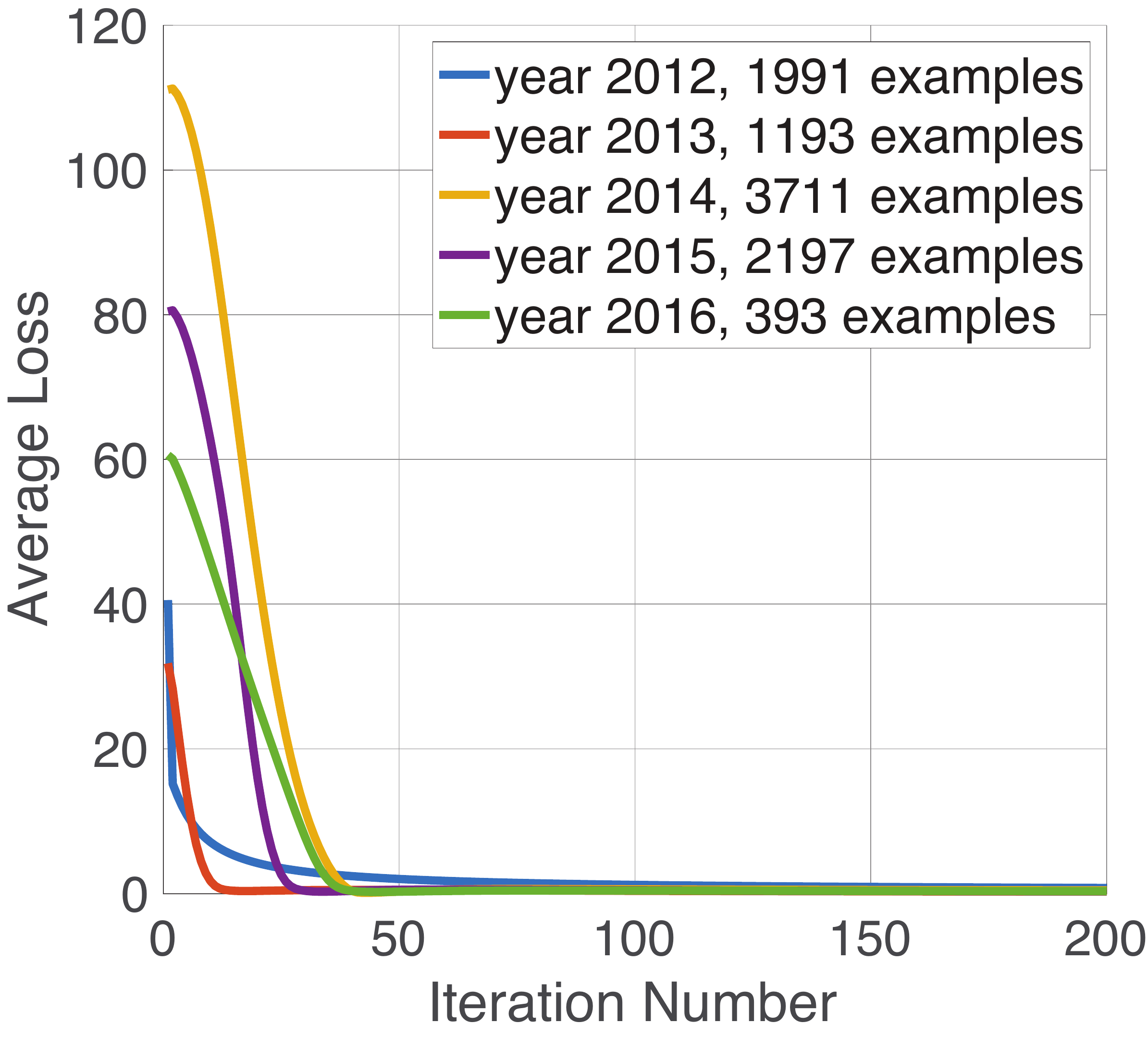, height = 4.5cm}} & \mbox{\epsfig{figure= 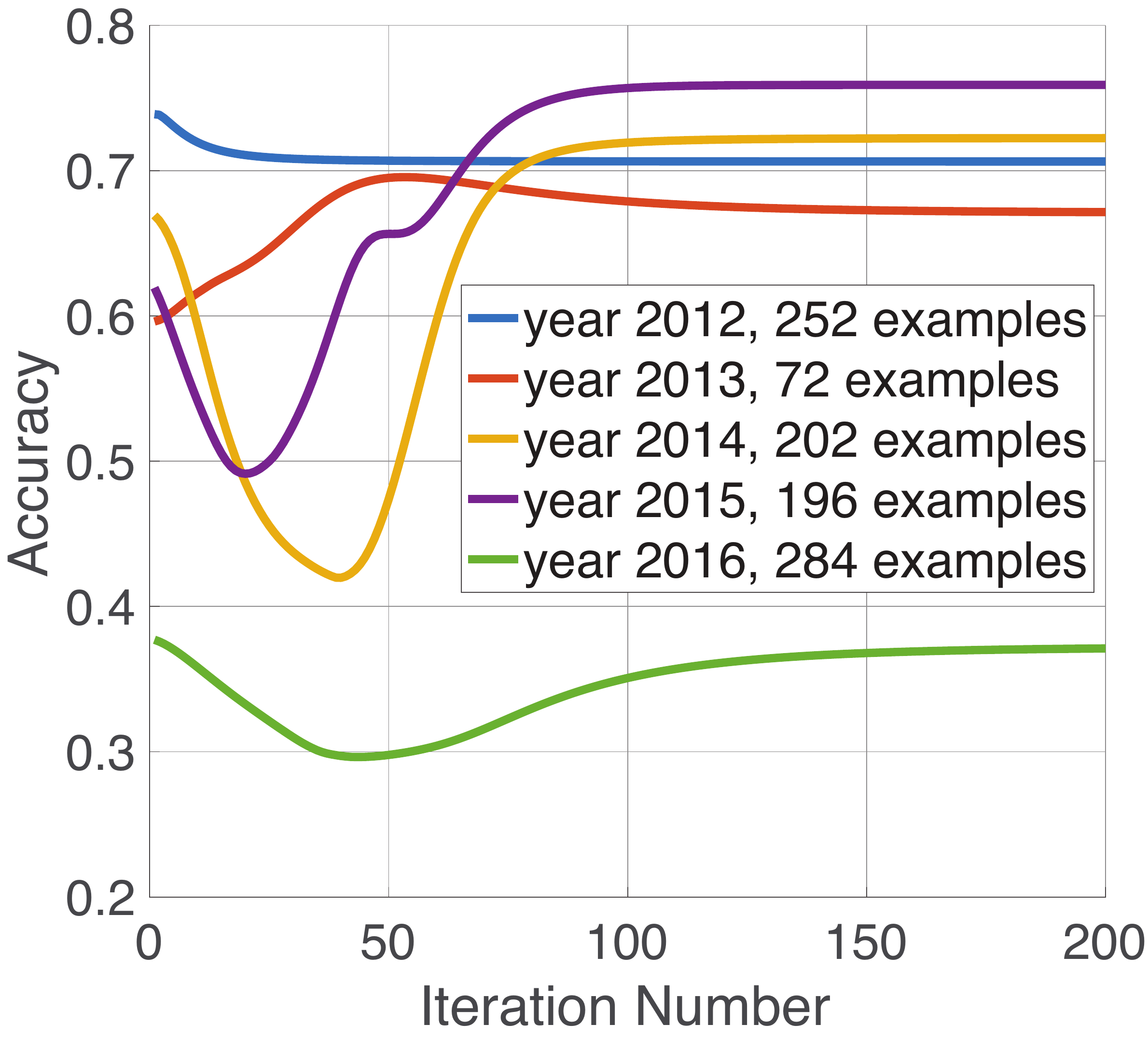, height = 4.5cm}} \\
    {\scriptsize (a) Average loss on training data.} & {\scriptsize (b) Accuracy (Pearson's $r \times 100$) on testing data.}
\end{tabular}
\caption{(Best viewed in color) Loss and accuracy on STS dataset.} \label{fig:sts_training}
\end{figure}

Here we run Alg.~\ref{alg:ca_sem_algorithm} on the training data for each year and apply it to the corresponding testing data.
Fig.~\ref{fig:sts_training} (a) shows the change in average loss (total energy of Eq.~\ref{equ:sem_reembedding_regroup} / number of training examples) in each iteration.
Fig.~\ref{fig:sts_training} (b) shows the corresponding accuracy computed over the corresponding testing data, which does not always increase but stabilizes after $100$ iterations.

Table~\ref{table:sem_sts_results} summarizes the comparisons among different approaches.





\begin{table}[ht]
\centering
\begin{tabular}{c|c|c|c}
\hline
  &    \small{GloVe} $+$ \small{Average}  &  \small{GloVe} $+$ \small{PCA} & \small{GloVe} $+$ \small{CA-SEM} (our approach) \\
\hline
\small{STS'} 12  &    $54.8$($52.5$)      & $72.3$($56.2$)  & $\mathbf{73.9}$ \\
\hline
\small{STS'} 13  &    $60.5$($42.3$)      & $66.1$($56.6$) &  $\mathbf{69.6}$ \\
\hline
\small{STS'} 14  &    $42.2$($54.2$)      & $68.4$($68.5$)  & $\mathbf{72.2}$  \\
\hline
\small{STS' }15  &     $50.4$($52.7$)     & $67.6$($71.7$) & $\mathbf{75.9}$ \\
\hline
\small{STS' }16  &     $30.1$     & $34.7$ & $\mathbf{37.2}$ \\
\hline
\end{tabular}
\caption{Results on SemEval 2017 dataset. The numbers inside the parentheses are reported by \cite{sem_arora2017}. }
\label{table:sem_sts_results}
\end{table}

\subsubsection{SICK similarity dataset}

\begin{figure}[t]
\centering
\begin{tabular}{cc}
    \mbox{\epsfig{figure= 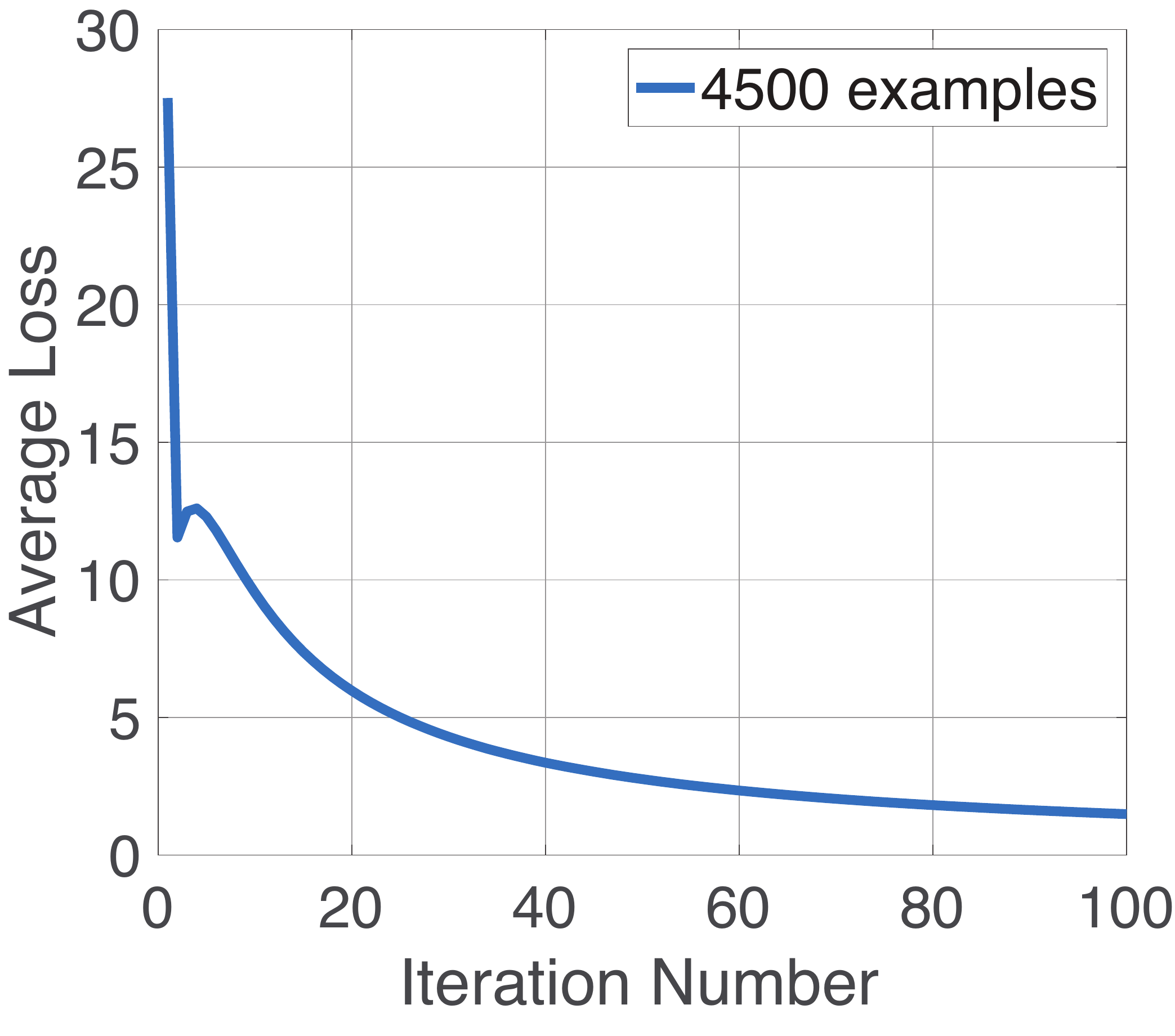, height = 4.5cm}} & \mbox{\epsfig{figure= 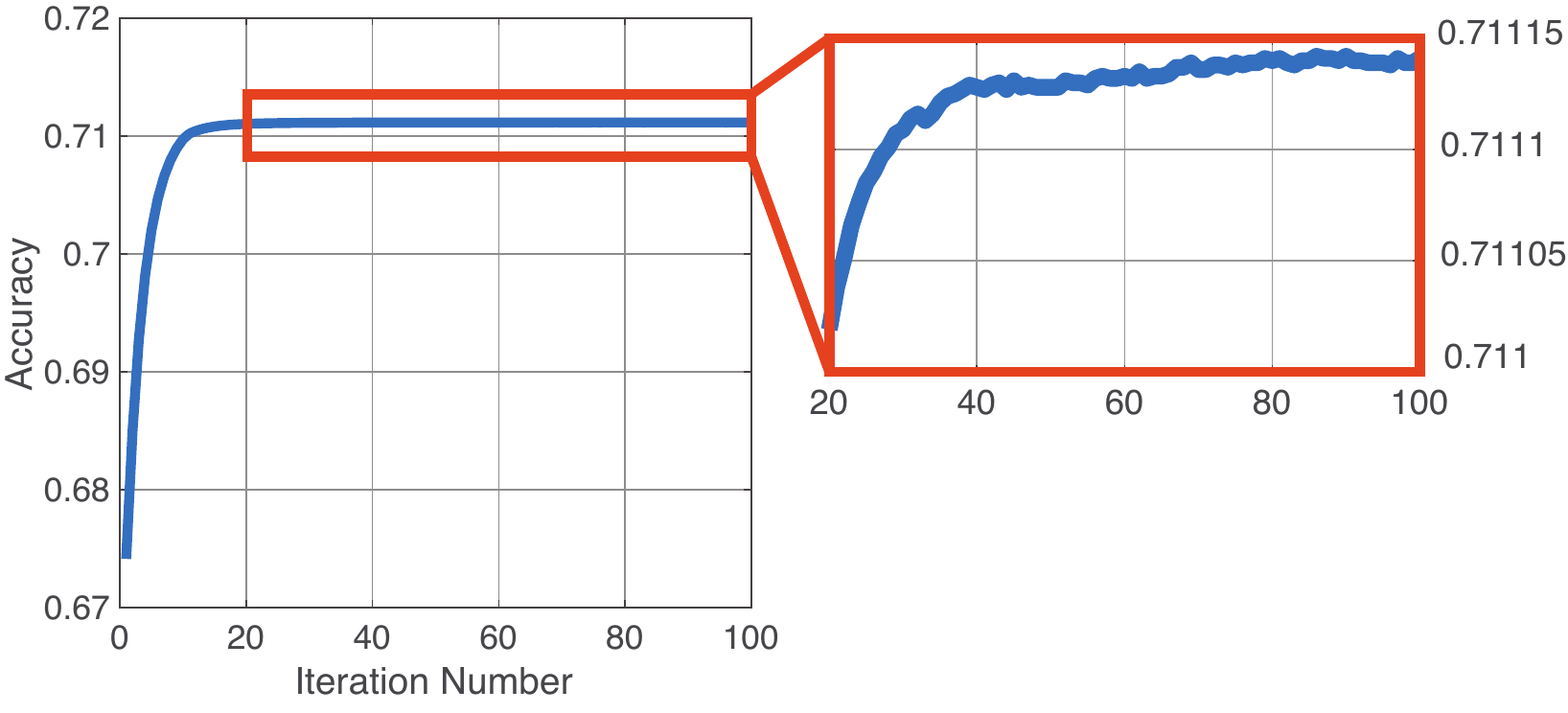, height = 4.5cm}} \\
    {\scriptsize (a) Average loss on training data.} & {\scriptsize (b) Accuracy (Pearson's $r \times 100$) on testing data.}
\end{tabular}
\caption{Training results on SICK dataset.} \label{fig:sick_training}
\end{figure}

Fig.~\ref{fig:sick_training} illustrates the loss and accuracy for each block coordinate iteration. 
In this case, the loss keeps decreasing after $20$ iterations (Fig.~\ref{fig:sick_training} (a)) but the accuracy remains near $0.71$ (Fig.~\ref{fig:sick_training} (b).
Nevertheless, this is a significant improvement over the original paper (Table~\ref{table:sem_sick_results}).

\begin{table}[ht]
\centering
\begin{tabular}{c|c|c|c}
\hline
  &    \small{GloVe} $+$ \small{Average}  &  \small{GloVe} $+$ \small{PCA} & \small{GloVe} $+$ \small{CA-SEM} (our approach) \\
\hline
\small{SICK'} 14 &     $62.4$($65.9$) &  $64.7$($66.2$) & $\mathbf{71.1}$ \\
\hline
\end{tabular}
\caption{Results on SICK' 14 dataset. The numbers inside the parentheses are reported by \cite{sem_arora2017}. }
\label{table:sem_sick_results}
\end{table}

\subsection{Results for CA-BSFE}
While CA-SEM is useful for unsupervised learning, CA-BSFE is designed for supervised learning, which provides a solution for handling any input containing bag of sparse features that are not limited to text.
In this section, we verify that its intrinsic support for computing the importance of each input (through $\chi$-function) can help
i) improve model training accuracy, 
ii) provide us with additional information from the input data and 
iii) provide us the flexibility in controlling generalization errors in training, especially when training data size is small.

\subsubsection{Implementation details}\label{subsubsec:cabsfe_impl}
The architecture of Fig.~\ref{fig:ca_bsfe} can be implemented using Tensorflow APIs, where the embedding of each word $w$ is sliced into two parts: word embedding $\vec{w'}$ and $\chi$-function input $\vec{w}_{\sigma}$ (Eq.~\ref{equ:bsfe}). A global vector $\vec{v}_0$ should also be defined as a trainable variable.
In the following we use $dim(\vec{w'})$ and $dim(\vec{w}_{\sigma})$ to denote the embedding dimensions of $\vec{w'}$ and $\vec{w}_{\sigma}$, respectively. 

We evaluate our CA-BSFE layer by plugging it into a sequence-to-binary, logistic regression problem (\emph{e.g.}, positive/negative sentiment prediction for movie reviews). 
Here the input words $\{w | w\in s\}$ are mapped into an embedding $\vec{s}$ via our CA-BSFE layer, then it is directly fed into a sigmoid output layer.
A standard sigmoid cross entropy is employed as the loss function.

\begin{figure}[t]
\centering
\begin{tabular}{ccc}
    \mbox{\epsfig{figure= 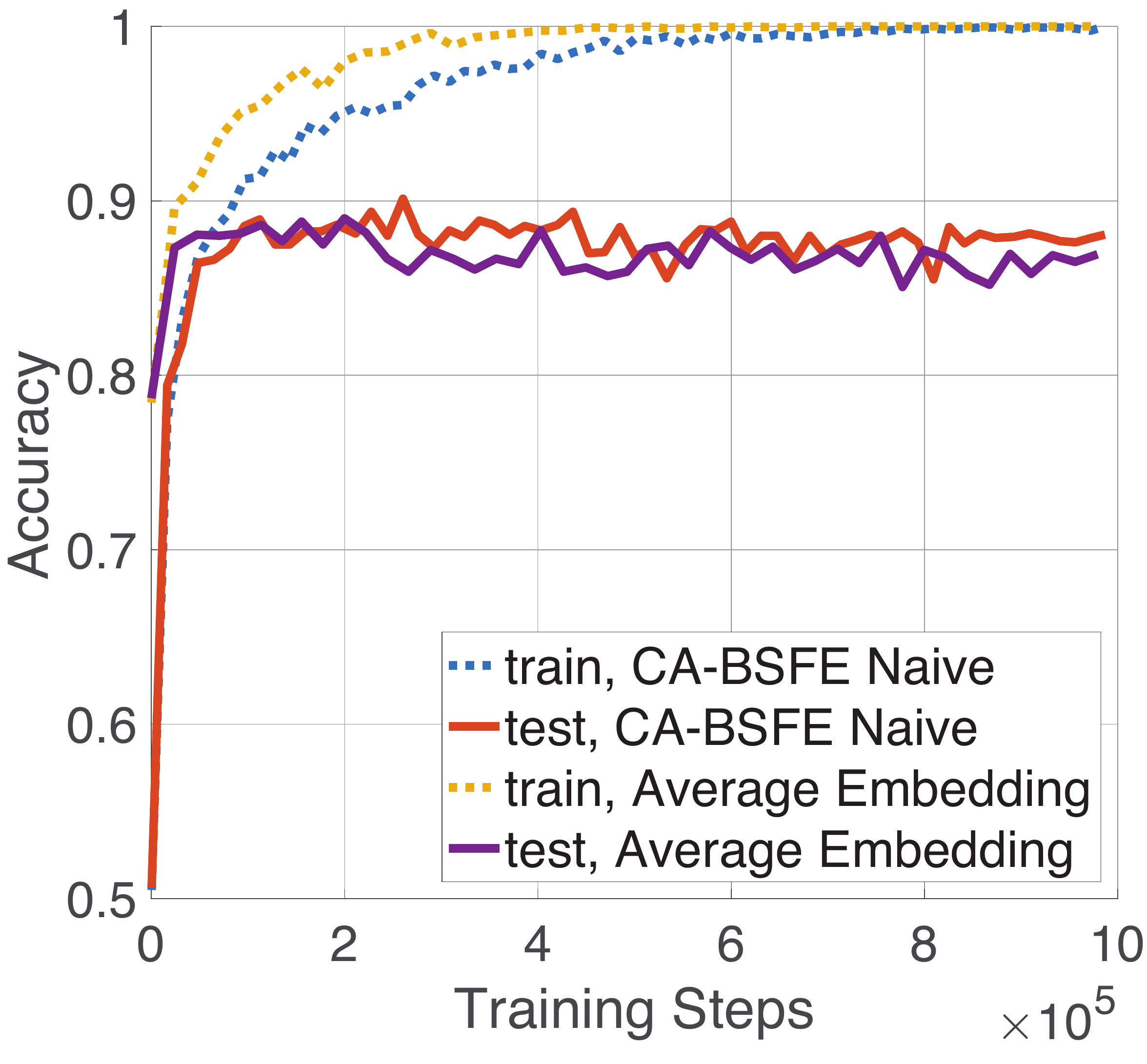, height = 4.2cm}} 
    & \mbox{\epsfig{figure= 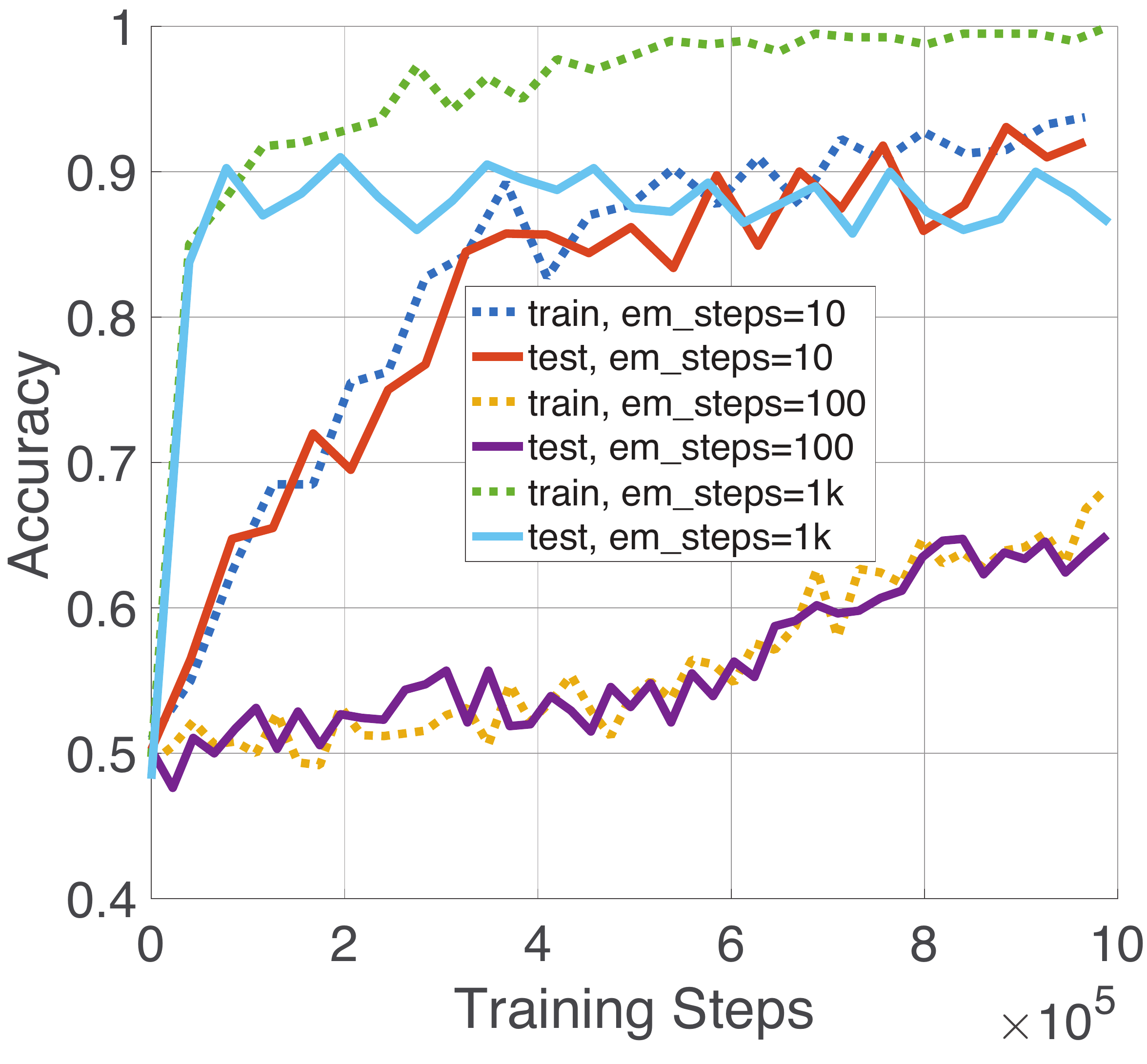, height = 4.2cm}} 
    & \mbox{\epsfig{figure= 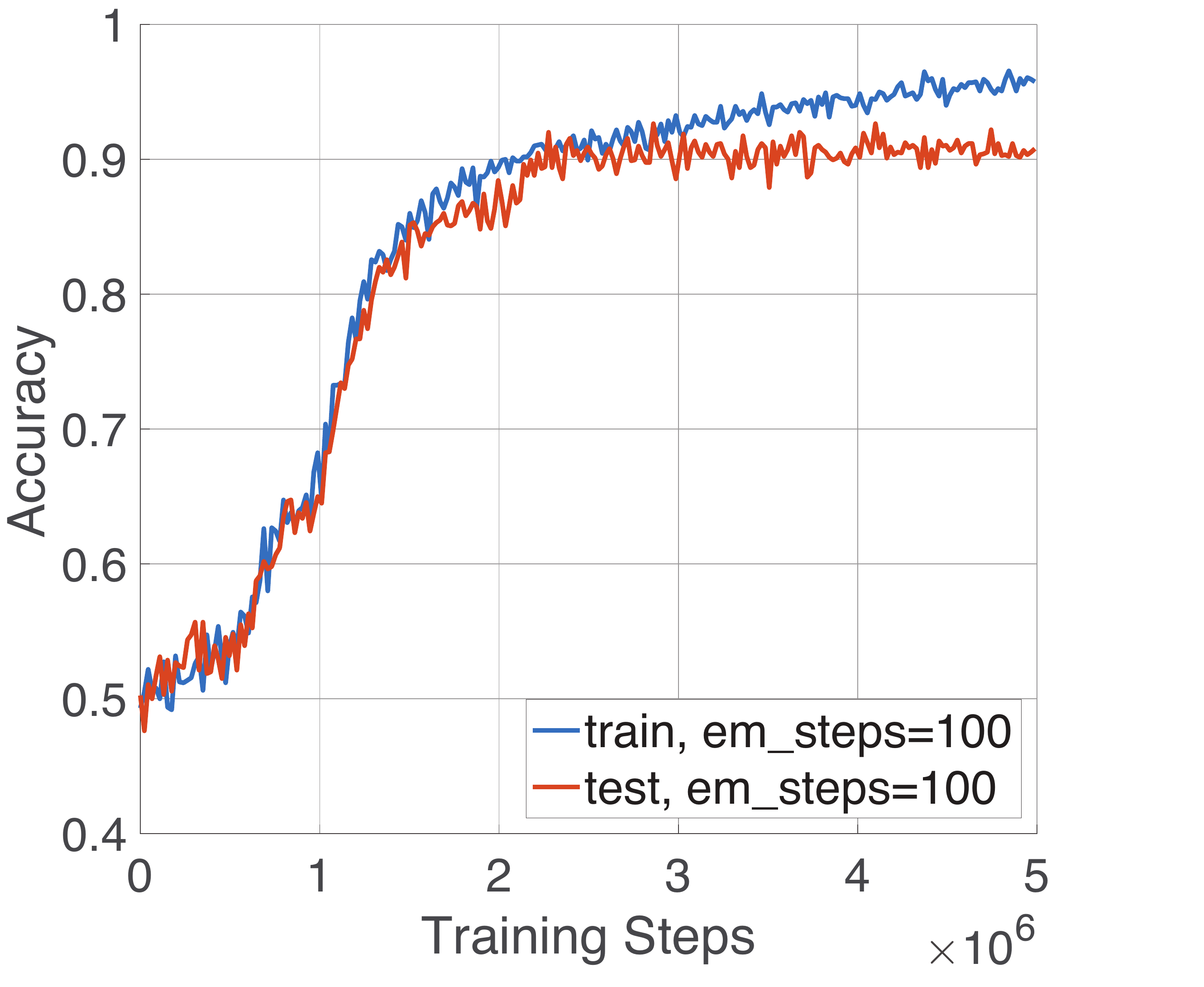, height = 4.2cm}} \\
    {\scriptsize (a)} & {\scriptsize (b)} & {\scriptsize (c)}
\end{tabular}
\caption{(Best viewed in color) Training results on IMDB dataset. Our block coordinate gradient update scheme can help improve the model's ability in generalization by preventing the model from memorizing the whole training dataset too quickly.} \label{fig:imdb_result}
\end{figure}

\textbf{Block-coordinate gradient update scheme}. However, in our first attempt to apply a naive implementation of CA-BSFE to the IMDB dataset, the gain in accuracy over average-based composition is quite marginal ($\sim 1$\%) as shown in Fig.~\ref{fig:imdb_result} (a).
One possible explanation is that the data size is too small that the model can quickly memorize the whole training dataset with zero error (see the training accuracies in Fig.~\ref{fig:imdb_result} (a)), \emph{i.e.}, the model can easily stuck at one of many possible minima that happens to not generalize well.
Fortunately, our model construction allows additional freedom in minima search that prevents the energy from converging too quickly. 
The idea is similar to Alg.~\ref{alg:ca_sem_algorithm} for CA-SEM: only update a subset of all the variables in each training step.
Specifically, we can alternate between fixing the values of $\{\chi(\vec{w}_{\sigma})| w\in \mathcal{W}\}$ or the values of $\{\vec{w'} | w \in \mathcal{W}\}$ every \emph{em\_steps} (here we employ the name ``em" as our algorithm resembles the expectation maximization, or EM algorithm) during the back-propagation update\footnote{This can be achieved using Tensorflow $1.0$ by two lines of code: \\
$should\_update = tf.equal(tf.mod( tf.train.get\_global\_step() / em\_steps, 2), 0)$ \\ $value = tf.cond(should\_update, lambda: value, lambda: tf.stop\_gradient(value))$.}.
In the following we will provide detailed analysis on how changing em\_steps can help reduce generalization error.

For all our experiments in this section, we use a batch size of $16$, a learning rate of $0.1$, $dim(\vec{w'}) = 5$ and an optimizer of Adagrad~\cite{tf_adagrad} as the basic training config.
For all our CA-BSFE models, we set $dim(\vec{w}_{\sigma}) = 5$.

\subsubsection{IMDB dataset}
Fig.~\ref{fig:imdb_result} shows the training and testing accuracy for different hyper-parameters.
Directly updating the gradients of $\chi(\vec{w}_{\sigma})$ and $\vec{w'}$ simultaneously only gains a small margin over embedding composition schema based on averaging (Fig.~\ref{fig:imdb_result}(a)). One can see that both methods can memorize the whole training dataset after $600$K steps. 
Applying our block-coordinate style gradient update makes obvious differences in narrowing the gap between training and test accuracy (Fig.~\ref{fig:imdb_result}(b)).
However, too large em\_steps (\emph{e.g.}, $1000$) still makes the model remember the training data quickly and leads to poor generalization. 
On the other hand, smaller em steps can lead to slower convergence but better accuracy (Fig.~\ref{fig:imdb_result}(c)).
For em\_steps $= 100$, the testing accuracy is stabilized above $0.91$ after $6$M steps, which is a significant improvement over the averaging scheme that achieves $0.88$ at best during the first $200$K training steps and degrades gradually afterwards.

\subsubsection{SST dataset}
\begin{figure}[t]
\centering
\begin{tabular}{ccc}
    \mbox{\epsfig{figure= 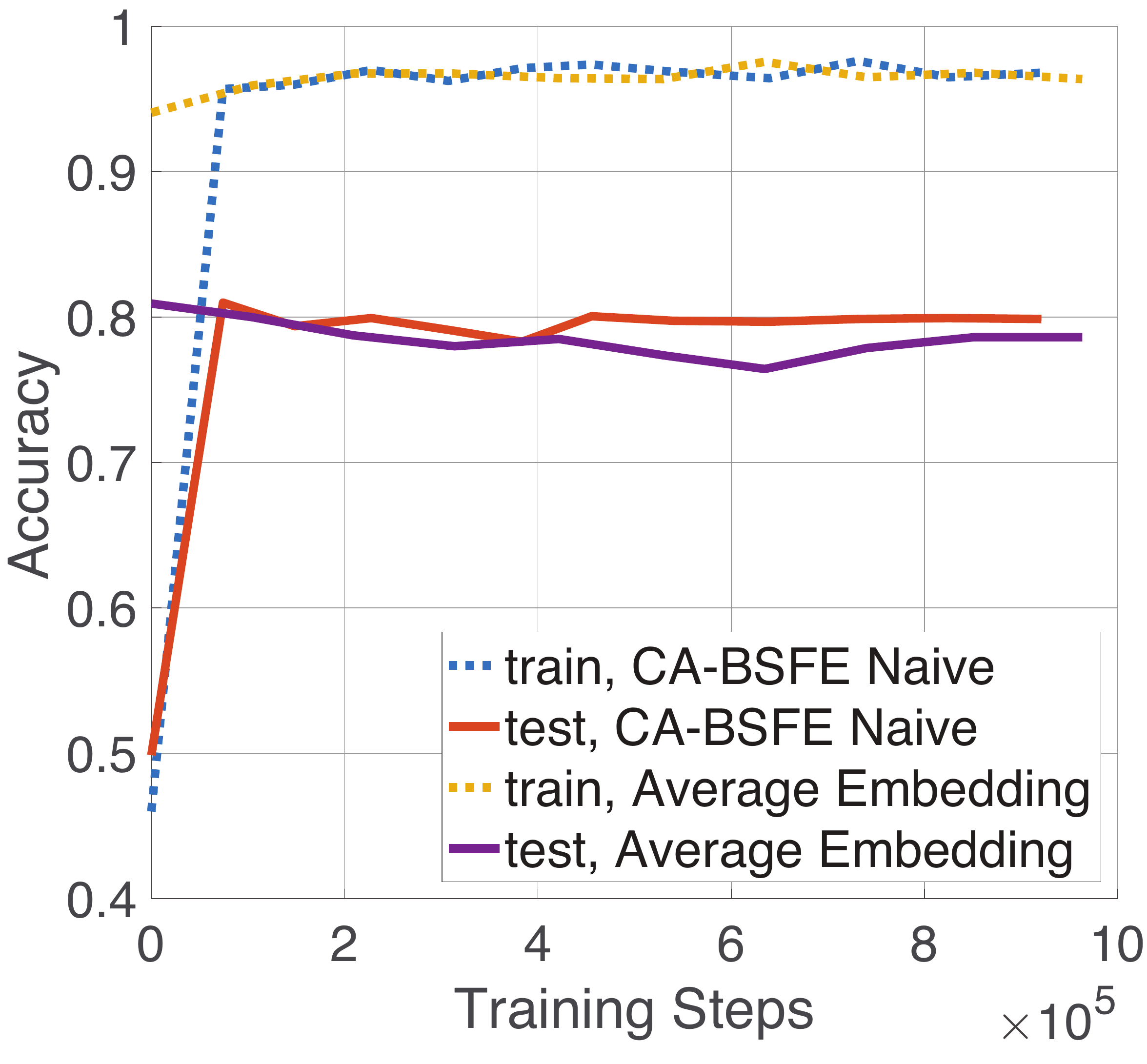, height = 4.2cm}} 
    & \mbox{\epsfig{figure= 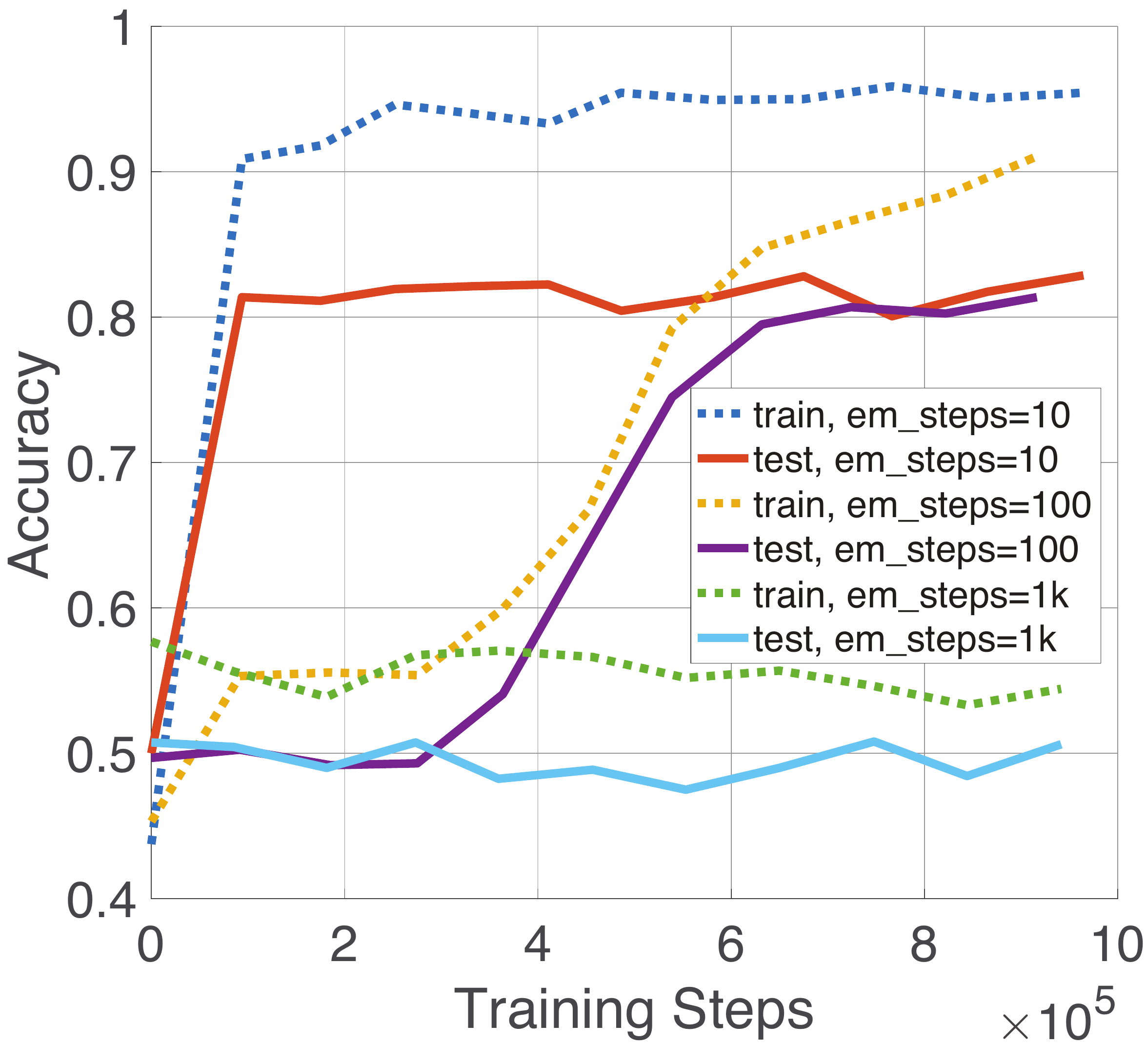, height = 4.2cm}} 
    & \mbox{\epsfig{figure= 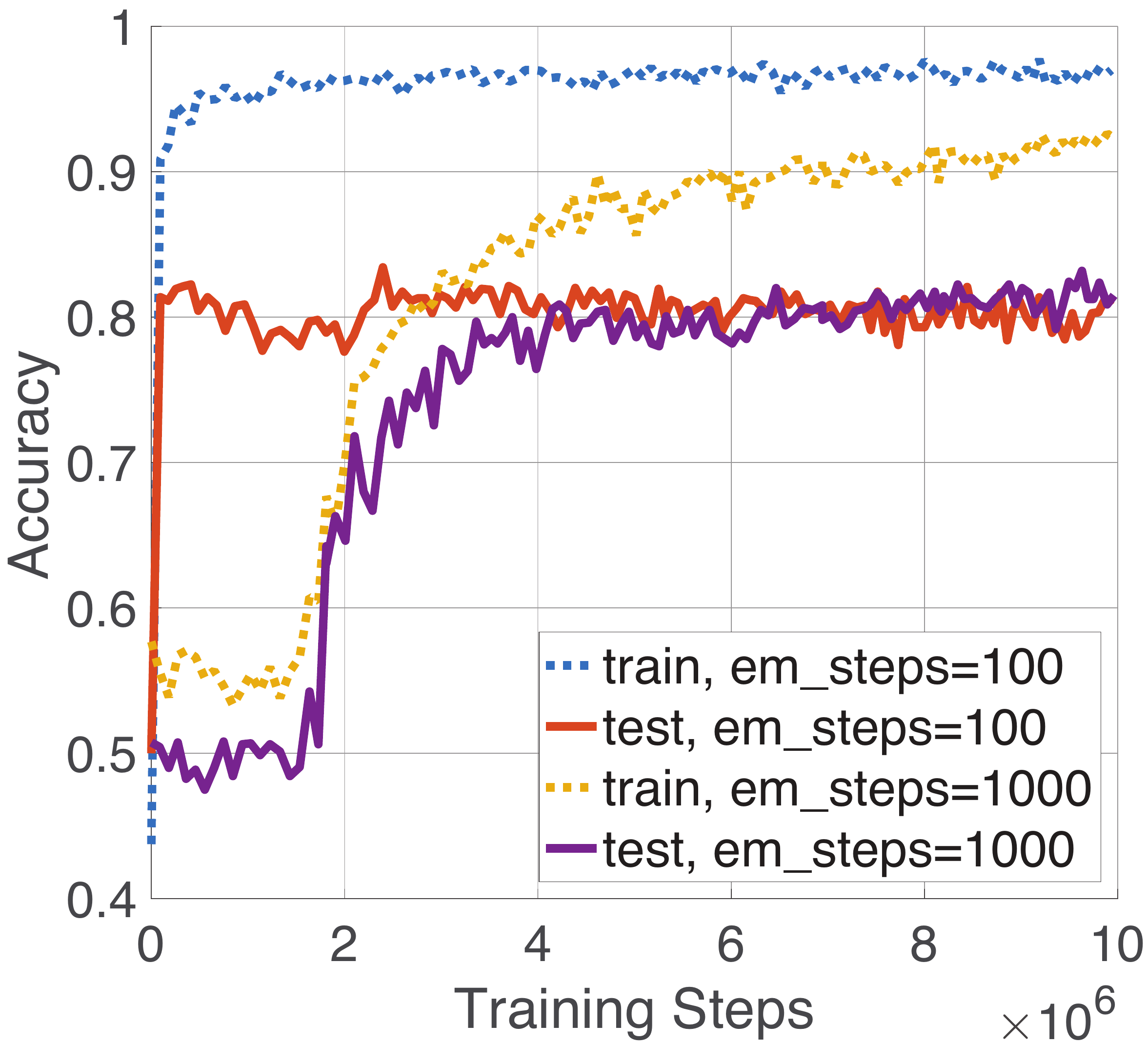, height = 4.2cm}} \\
    {\scriptsize (a)} & {\scriptsize (b)} & {\scriptsize (c)}
\end{tabular}
\caption{(Best viewed in color) Training results on SST dataset.} \label{fig:sst_result}
\end{figure}

Results on the SST dataset further confirms our observation on the benefits of our block-coordinate gradient update scheme.
In this case there are relatively a large number of training instances ($67,349$ \emph{v.s.} $25,000$ for IMDB) and a small number of testing cases ($1,821$ \emph{v.s.} $25,000$ for IMDB),
and the same model cannot memorize all the training data.
Nonetheless, using an em\_steps $>0$ still helps the generalization as shown in Fig.~\ref{fig:sst_result}.
It still hold true that CA-BSFE without em steps outperforms average-based scheme by a small margin (Fig.~\ref{fig:sst_result}(a)), but with em steps the accuracy can easily stabilize above $0.82$ (Fig.~\ref{fig:sst_result}(b) and (c)). However, this time for em\_steps $= 1000$ it performs poorly for the first $1$M steps (Fig.~\ref{fig:sst_result}(b)), but it starts to catch up after $2$M training steps (Fig.~\ref{fig:sst_result}(c)), which means the choice of em\_steps strongly depends on the data.

\subsection{Results for CA-ATT}
\begin{figure}[t]
\centering
\begin{tabular}{c}
    \mbox{\epsfig{figure= 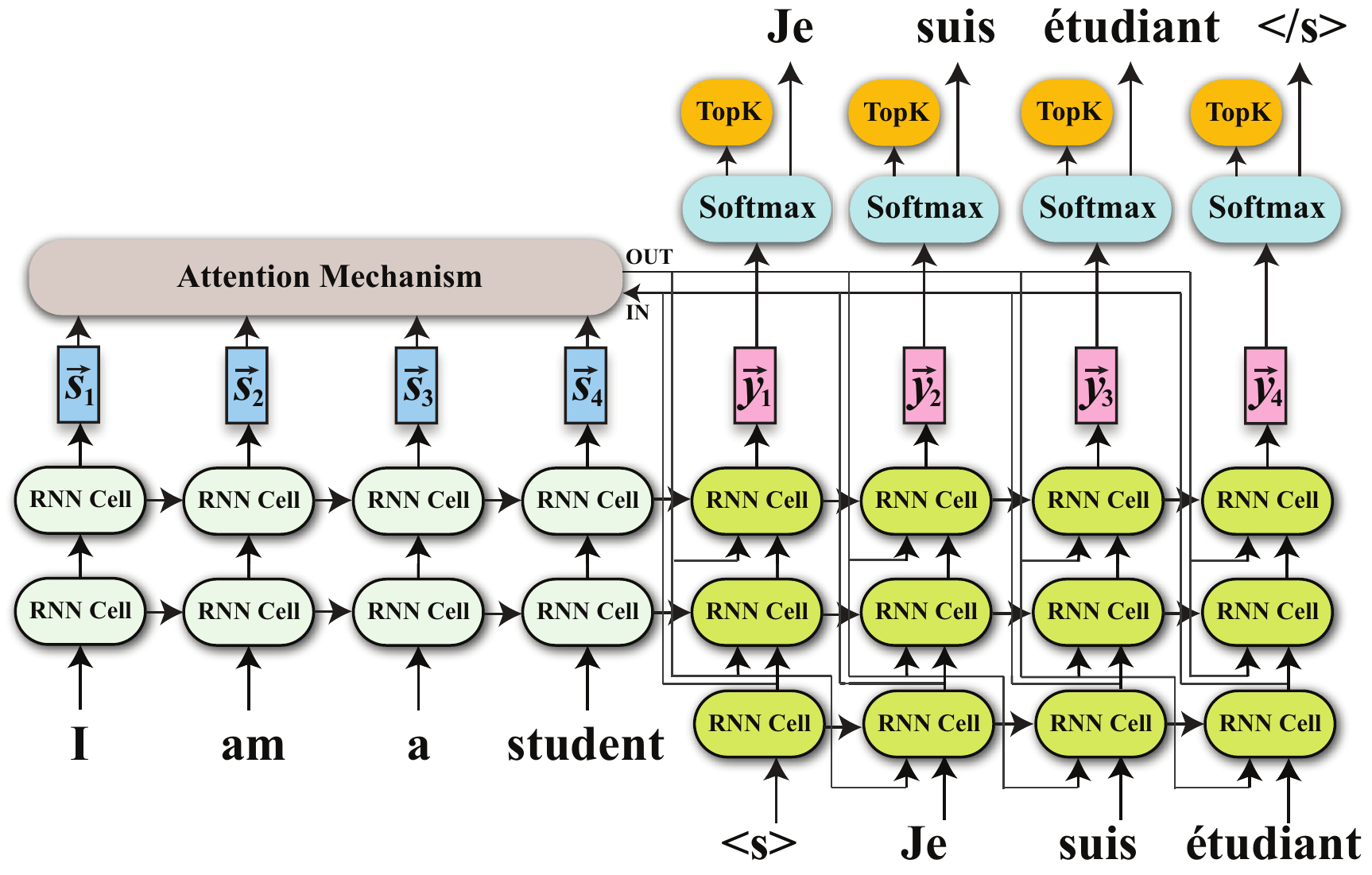, height = 7.8cm}}
\end{tabular}
\caption{The architecture of NMT model~\cite{attention_nmt14}. } \label{fig:nmt_model}
\end{figure}

Although our attention model share similar forms with existing ones (\emph{e.g.}, \cite{attention_nmt14,attention_luong15}), the assumptions behind them are quite different.
Recall that in Sec.~\ref{subsection:caatt}, we have connected attention to the distribution $P(s|c_{t - 1})$, where $s$ is the input sequence (\emph{a.k.a} memory in Fig.~\ref{fig:ca_att}) and $c_{t - 1}$ is the cell state that summarizes the target output up to time $t - 1$, or query.
This is surprisingly consistent with existing implementation of NMT model~\cite{nmt_tutorial_luong17}, which uses the cell state from previous time to query the attention vector (Fig.~\ref{fig:nmt_model}). 

We compare our attention model with two popular ones: \emph{Bahdanau attention}~\cite{attention_nmt14} and \emph{Luong attention}~\cite{attention_luong15}.
The major difference between the above mentioned two attentions is how the scoring function $f(\cdot, \cdot)$ in Eq.~\ref{equ:attention_original} is defined:
\begin{align}
f(\vec{s_i}, \vec{c}) =
\begin{cases}
\mathbf{v}^T \tanh (\mathbf{W}_1 \vec{c} + \mathbf{W}_2 \vec{s_i}) & \text{Bahdanau attention} \\
\vec{c}^T \mathbf{W} \vec{s_i} & \text{Luong attention}
\end{cases}
\end{align}
In fact, our $\chi$-function in Eq.~\ref{equ:sigma_for_caatt} can also be defined using one of the above two forms.
In this section, we show detailed analysis on the performance of these different implementations.

\subsubsection{Implementation details}
\label{subsubsec:caatt_impl}
Our new attention model is implemented using Tensorflow by making some slight changes to existing \emph{tf.contrib.seq2seq.AttentionWrapper} class\footnote{\url{https://www.tensorflow.org/api_docs/python/tf/contrib/seq2seq/AttentionWrapper}} to allow for customized attention function and implementing a new subclass for \emph{tf.contrib.seq2seq.AttentionMechanism}\footnote{\url{https://www.tensorflow.org/api_docs/python/tf/contrib/seq2seq/AttentionMechanism}}.
This way, we can easily compare different attention mechanisms side by side.

Our model is based on the publicly available implementation of the neural machine translation (NMT) architecture~\cite{attention_nmt14}, which involves a sequence to sequence model with attention mechanism~\cite{nmt_tutorial_luong17}. As mentioned above, unlike any existing implementations, we do not use any pre-defined vocabulary for token embedding.
The hyper-parameters for our model are selected as follows: batch size is $16$; learning rate is $0.01$; embedding dimension for all tokens in source, target and output sequences is $100$; optimizer for gradient descent is Adagrad. Our training is distributed into $100$ worker machines without any hardware accelerations and it can achieve $20 - 30$ global steps per second.

\subsubsection{WMT' 16 dataset}

\begin{figure}[t]
\centering
\begin{tabular}{ccc}
    \mbox{\epsfig{figure= 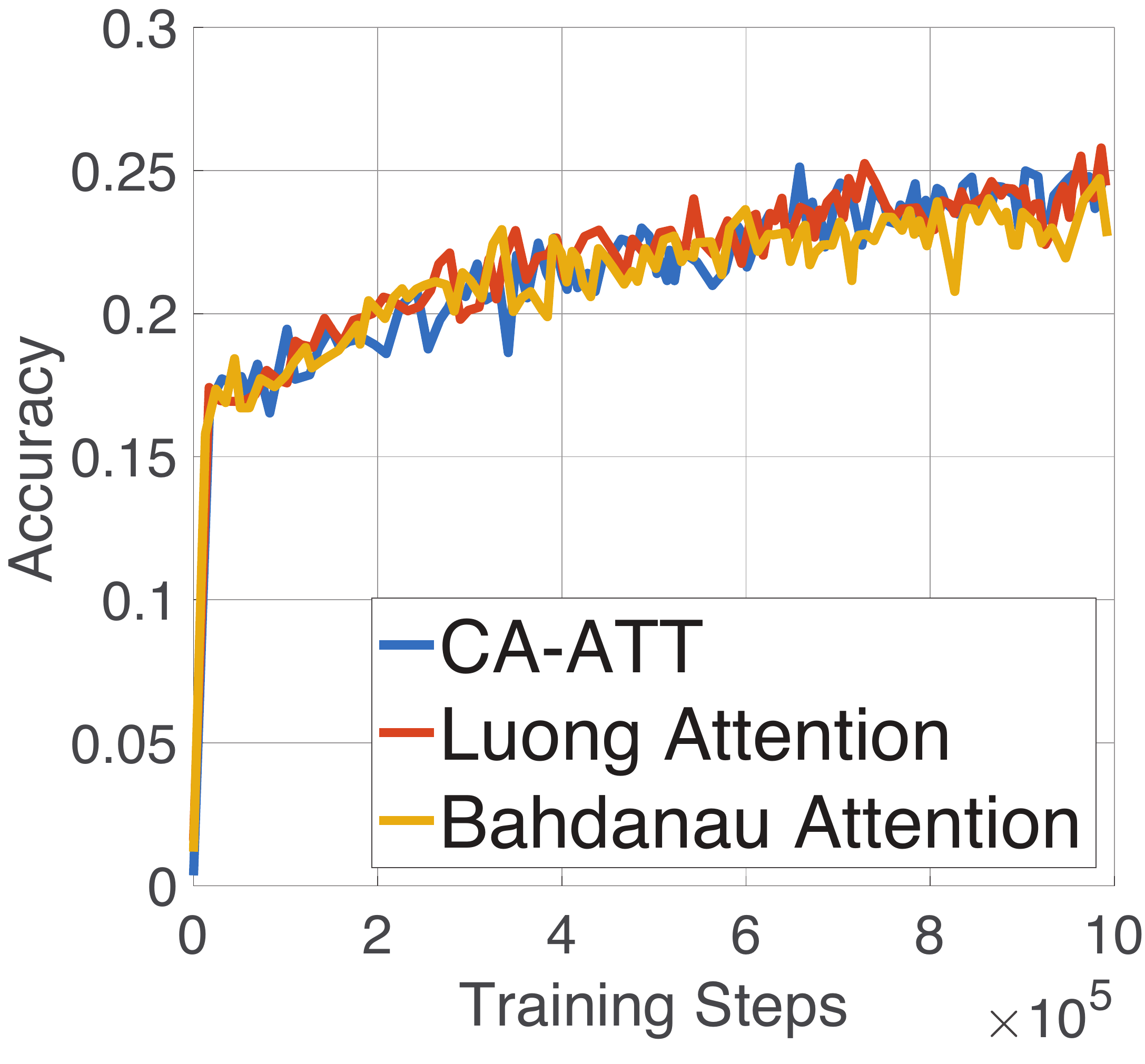, height = 4.2cm}} 
    & \mbox{\epsfig{figure= 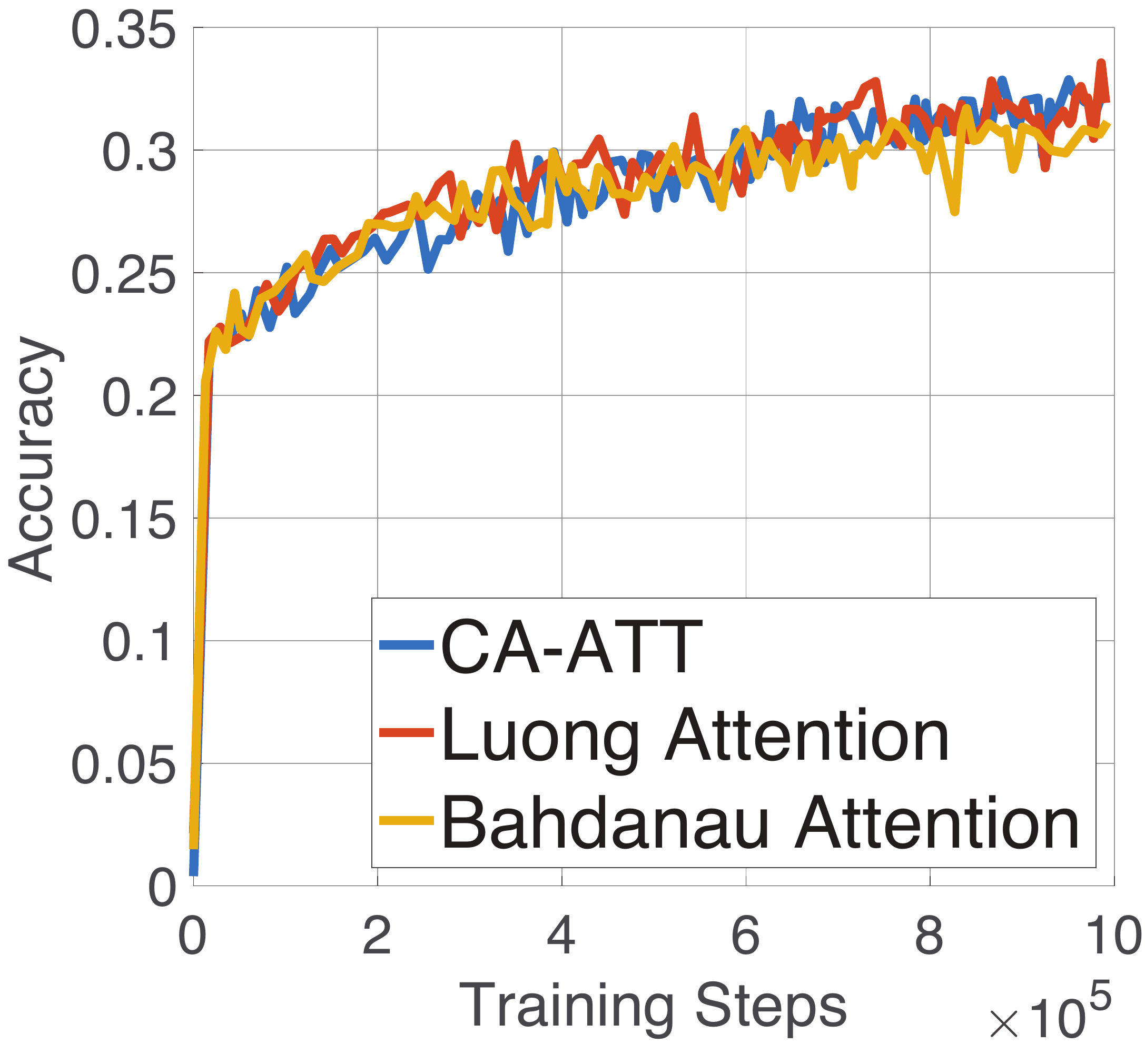, height = 4.2cm}} 
    & \mbox{\epsfig{figure= 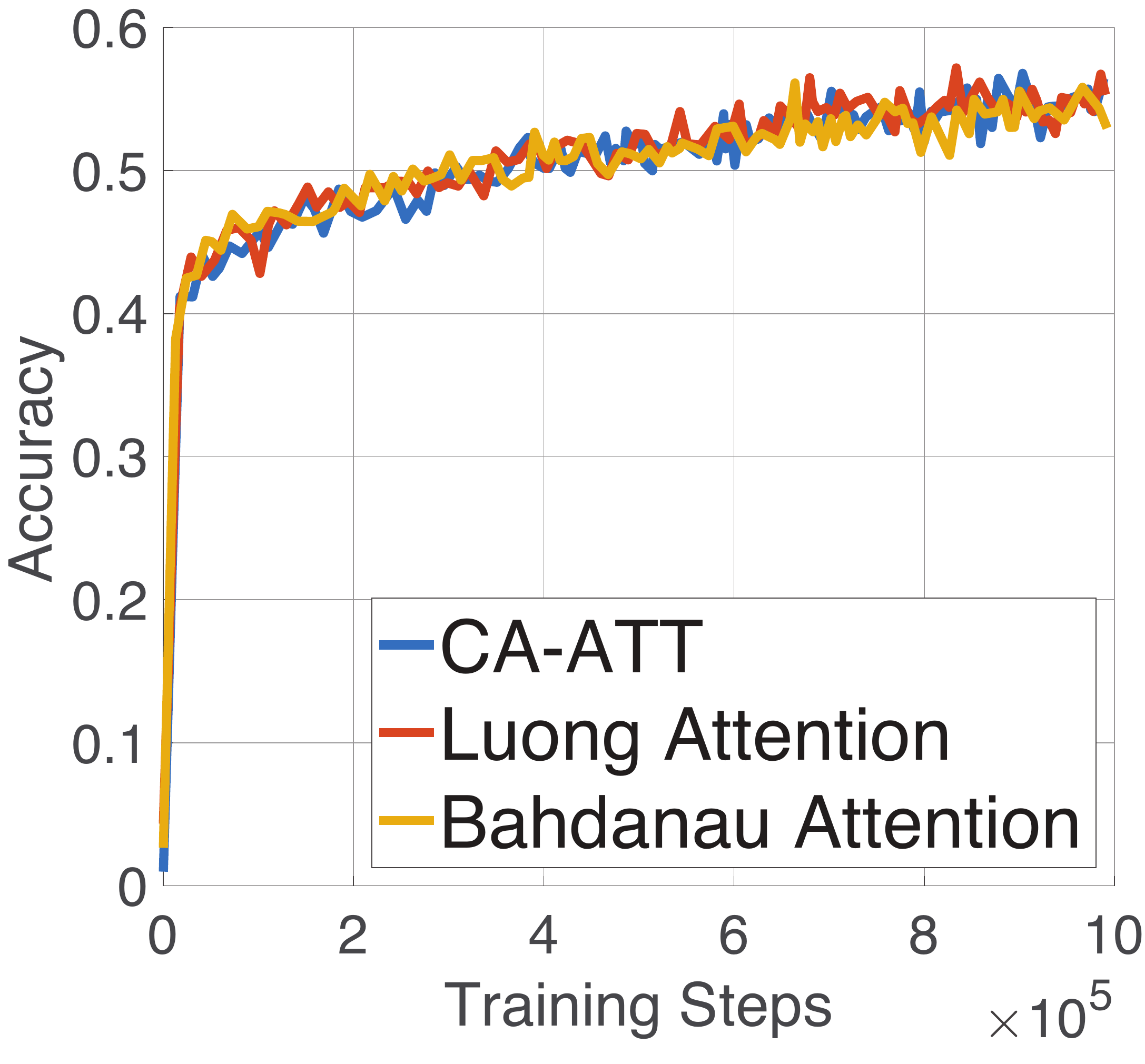, height = 4.2cm}} \\
    {\scriptsize (a) Top $5$} & {\scriptsize (b) Top $10$} & {\scriptsize (c) Top $100$}
\end{tabular}
\caption{(Best viewed in color) Comparison among different attention mechanisms in the NMT architecture Fig.~\ref{fig:nmt_model} (a) for WMT' 16 dataset.} \label{fig:attention_result}
\end{figure}

When comparing our new attention model with existing ones, unfortunately, we find that there is no significant performance gain: all three attention mechanisms are quite comparable.
Fig.~\ref{fig:attention_result} illustrates the training accuracy computed by looking at if the top k (k $=5, 10, 100$) tokens returned by the TopK layer in Fig.~\ref{fig:nmt_model} contain the target token.
We have also tried to rewire the model in Fig.~\ref{fig:nmt_model}, \emph{e.g.}, by using the cell state from the top RNN cell as query and feed the returned attention vector to the output.
Still none of these efforts result in any better performance gain; most of them perform significantly worse than the original design.

Hence, the lesson learned from these experiments is that designing a model based on the factorization can lead to the `correct' solution, as we will see in the next section.

\subsection{Results for CA-RNN}

\subsubsection{Implementation details}
\label{subsubsec:carnn_impl}
In Tensorflow each sequence model (\emph{e.g.}, RNN, GRU, LSTM) is implemented as a subclass of  \emph{tf.contrib.rnn.LayerRNNCell}\footnote{\url{https://www.tensorflow.org/api_docs/python/tf/contrib/rnn/LayerRNNCell}}. Therefore we only need to implement our CA-RNN model as a new RNN cell under the same interface, whose output is a tuple of (output embedding, cell state embedding).

Because our EDF only provides a form of the CA-RNN model (Eq.~\ref{equ:carnn_cellstate} and~\ref{equ:carnn_output}), choosing the right implementation is a trial-and-error process that often depends on the data of interest.
Nevertheless, we require that our model should at least pass some very basic test to be considered valid.
This said, we design a very simple model, which take a sequence as input and it is connect to any RNN cell (CA-RNN, GRU or LSTM), whose cell state embedding is further connect to a logistic regression output as a $0/1$ classifier.
Our synthetic training data is also simple.
For training we let the model learn the mappings \{'I', 'am', 'happy'\} $\rightarrow 1$ and \{'You', 'are', 'very', 'angry'\} $\rightarrow 0$ for $100$ iterations.
For testing we expect the model to predict \{'I', 'am', 'very', 'happy'\} $\rightarrow 1$ and \{'You', 'look', 'angry'\} $\rightarrow 0$ (our model predicts $1$ if its logistic output is greater than $0.5$, otherwise it predicts $0$).
This model validation process can help us quickly rule out impossible implementations.
For example, we found the peephole parameter $\mathbf{p}_v$ in Eq.~\ref{equ:carnn_cell_state_impl} cannot be replaced by matrix as linear mapping. 
Otherwise the model would not predict the correct result for the testing data. 

\begin{figure}[t]
\centering
\begin{tabular}{cc}
   \hspace{10mm} \mbox{\epsfig{figure= 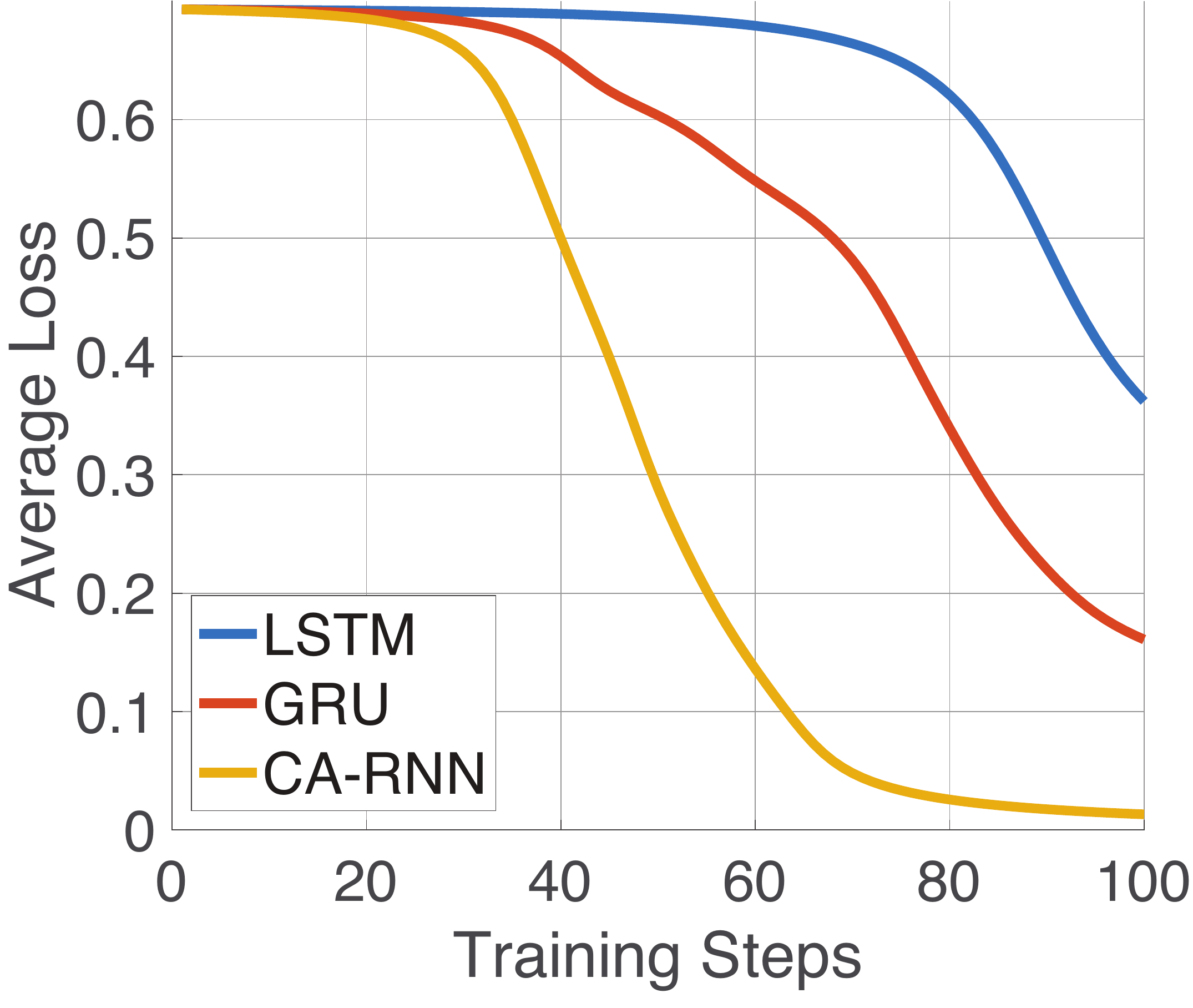, height = 4.2cm}}
    & \hspace{-30mm}\vbox{
\begin{tabular}{c|c|c}
\hline
 LSTM &  GRU   &  CA-RNN \\
\hline
$0.318$ &  $0.172$ & $\mathbf{0.126}$ \\
\hline
\end{tabular}
\vspace{15mm} }\\
    \hspace{10mm} {\scriptsize (a) Average training loss.}
    & \hspace{-30mm}{\scriptsize (b) Average prediction error.}
\end{tabular}
\caption{(Best viewed in color) Results on synthetic dataset for model validation described in Sec.~\ref{subsubsec:carnn_impl}.} \label{fig:carnn_synthetic_result}
\end{figure}

Fig.~\ref{fig:carnn_synthetic_result} shows the average loss and prediction error from running our model $10$ times on the above mentioned testing data.
It can be seen that CA-RNN model is better at both memorizing training data and generalizing to novel input.

\subsubsection{WMT' 16 dataset}
Here we employ the same model and hyper-parameters mentioned in Sec.~\ref{subsubsec:caatt_impl} with a fixed attention mechanism (\cite{attention_luong15}), only to change the RNN cell for both encoder and decoder for comparisons.

To evaluate the accuracy of the model on testing data, a BLEU score~\cite{bleu_score} metric is implemented for evaluation during training: for each target output token, simply feed it into the decoder cell and use the most likely output tokens (\emph{i.e.}, the top 1 token) as the corresponding translation.
We argue that this is a better metric to reflect the performance of a model than the BLEU score computed based on translating the whole sentence, since the later may vary with the choice of post-processing such as beam search.

\begin{figure}[t]
\centering
\begin{tabular}{cccc}
    \mbox{\epsfig{figure= 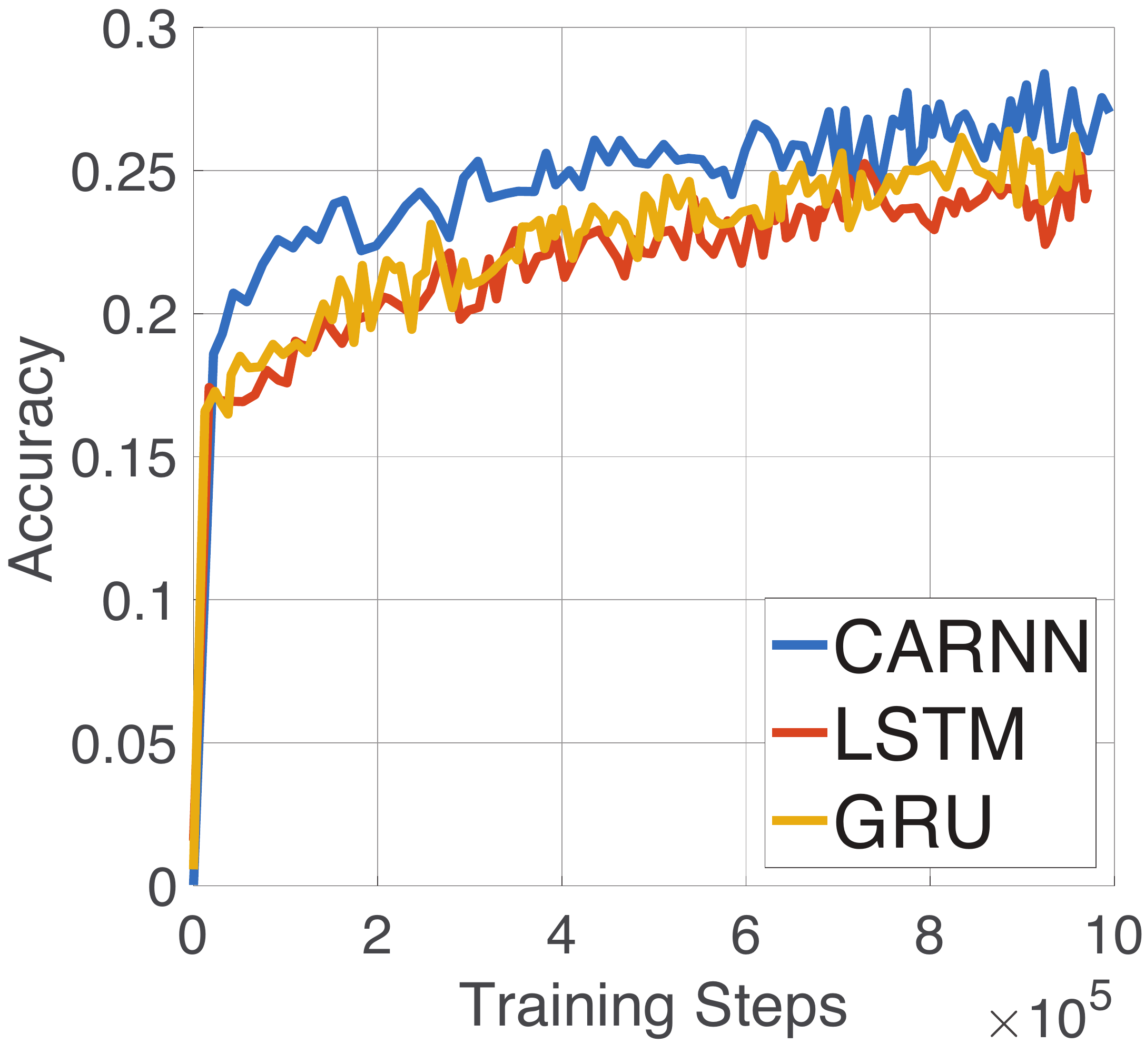, height = 3.2cm}}
    & \hspace{-4mm}\mbox{\epsfig{figure= 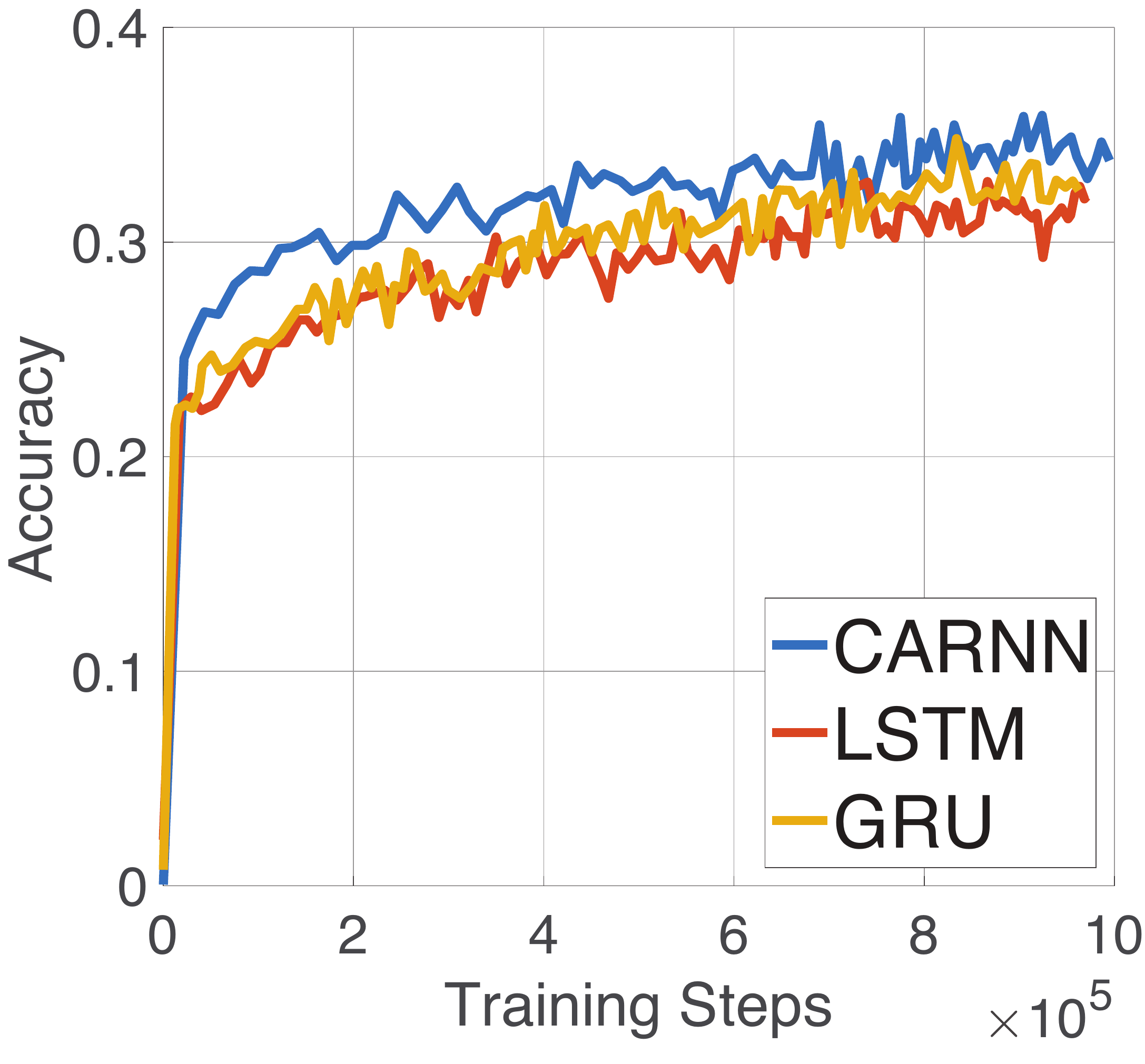, height = 3.2cm}}
    &  \hspace{-4mm}\mbox{\epsfig{figure= 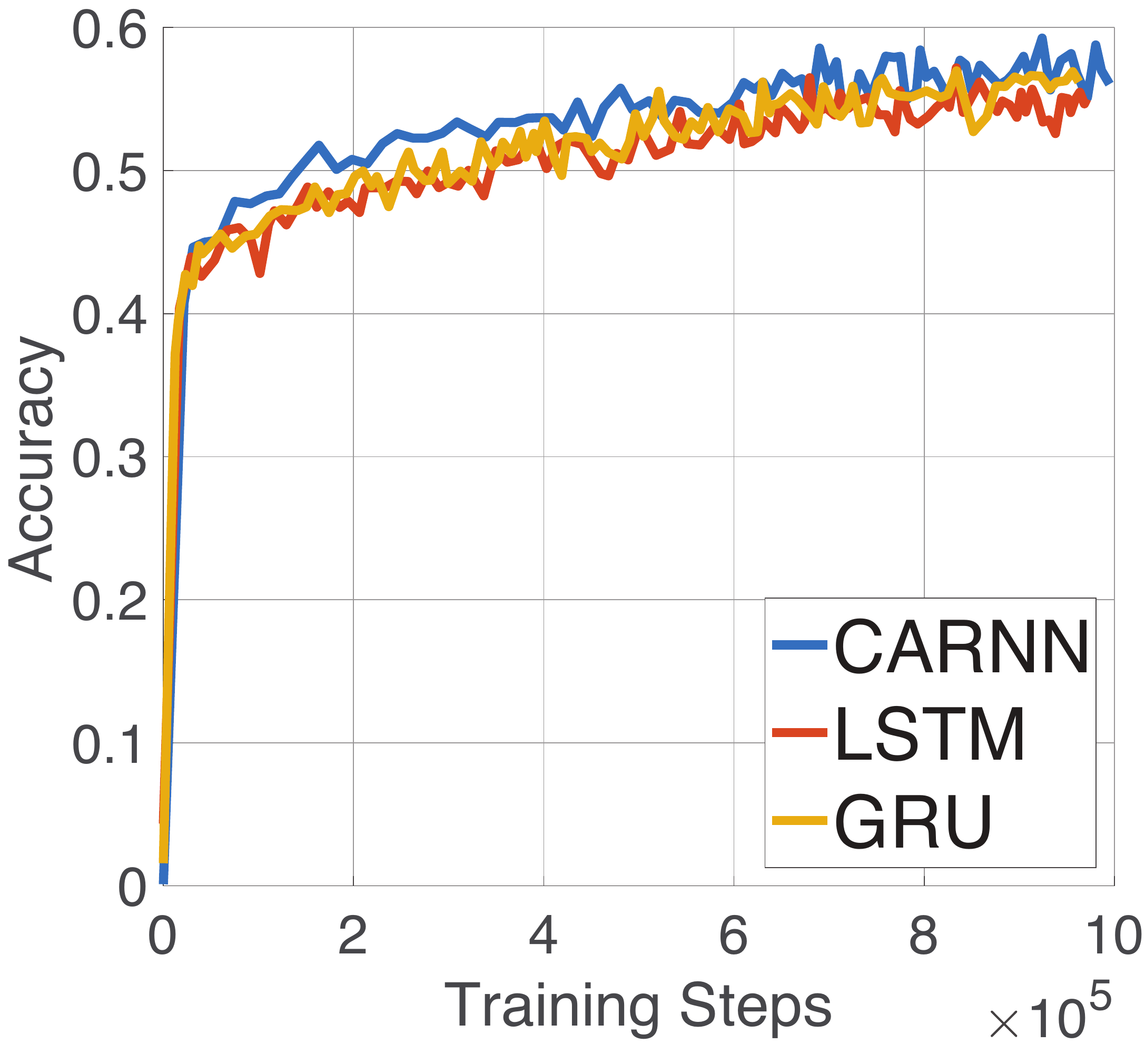, height = 3.2cm}} 
    &  \hspace{-4mm}\mbox{\epsfig{figure= 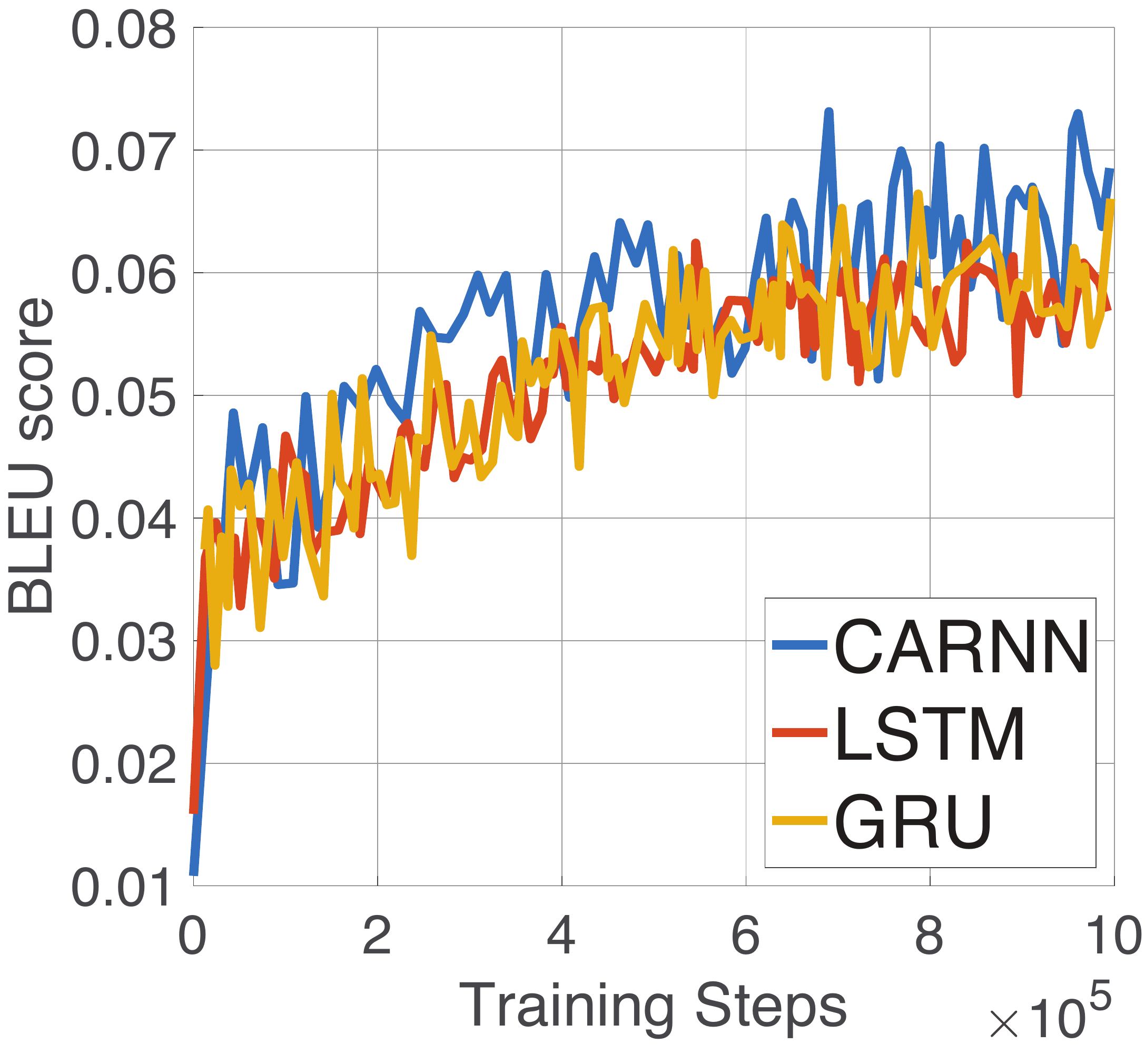, height = 3.2cm}} \\
    {\scriptsize (a) Top $5$}
    & {\scriptsize (b) Top $10$.}
    & {\scriptsize (c) Top $100$}
    & {\scriptsize (d) BLEU score}
\end{tabular}
\caption{(Best viewed in color) Training accuracy on WMT' 16 English to German dataset for first 1M training steps.} \label{fig:carnn_result}
\end{figure}

From the results of first $1$M training steps, one can see that CA-RNN consistently outperforms LSTM and GRU in top-k accuracy (Fig.~\ref{fig:carnn_result}(a), (b) and (c)) and BLEU score for prediction (Fig.~\ref{fig:carnn_result}(d)).

\subsection{Results for CA-NN}
A single layer CA-NN model can be considered as the context-aware counterpart of a NN hidden layer, whose applications is mainly focused on regression.
It is already established that a NN with a single hidden layer can approximate arbitrary function~\cite{nn_approximation} given sufficient amount of training data and training steps.
However, a simple NN hidden layer's ability to generalize may be poor given its simple structure.
In contrast, our CA-NN model allows additional flexibility to improve the model's generalization power as we will show in this section.

\subsubsection{Surface Fitting}

\begin{figure}[t]
\centering
\begin{tabular}{cccc}
    \mbox{\epsfig{figure= 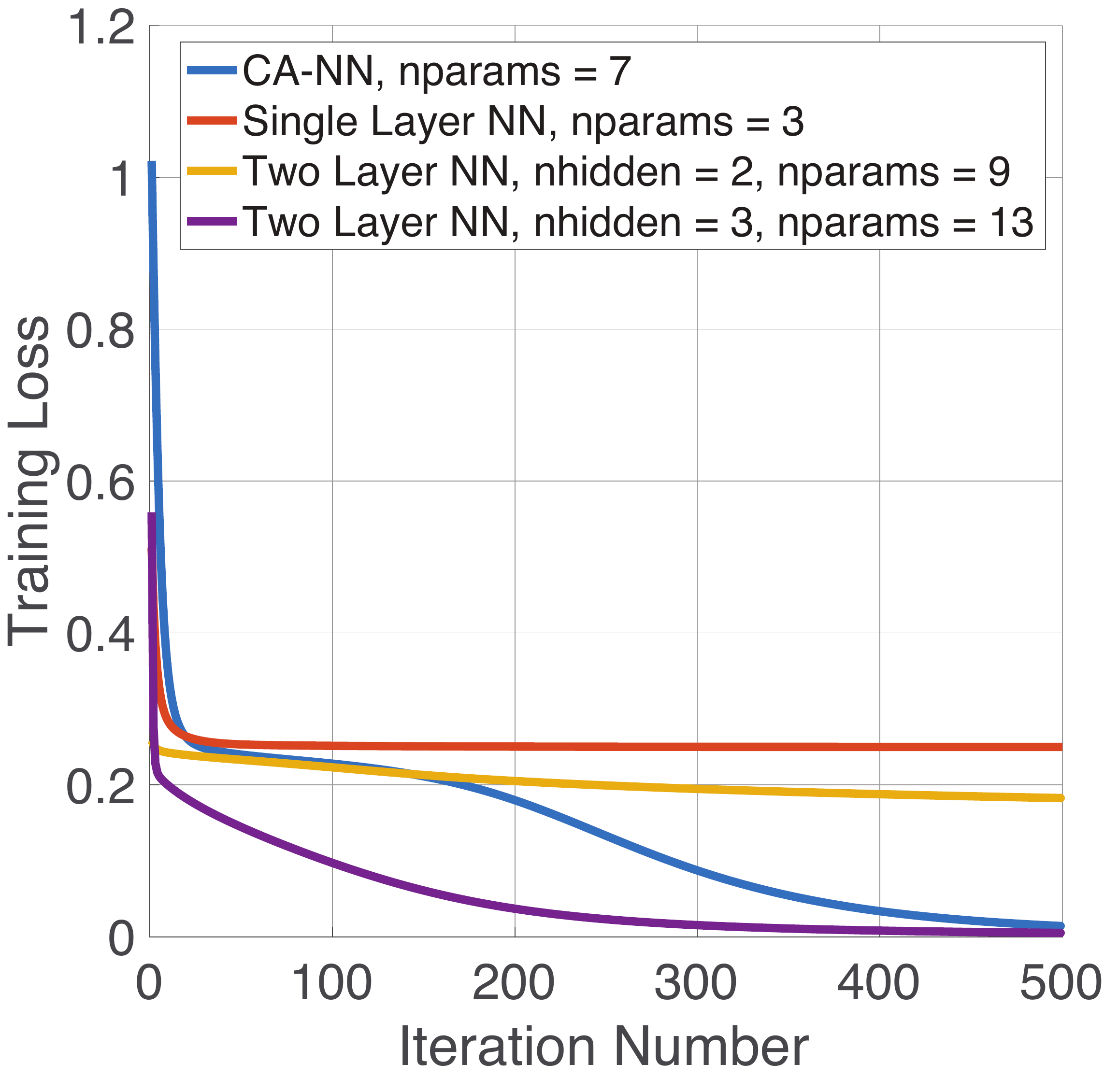, height = 2.8cm}} 
    & \mbox{\epsfig{figure= 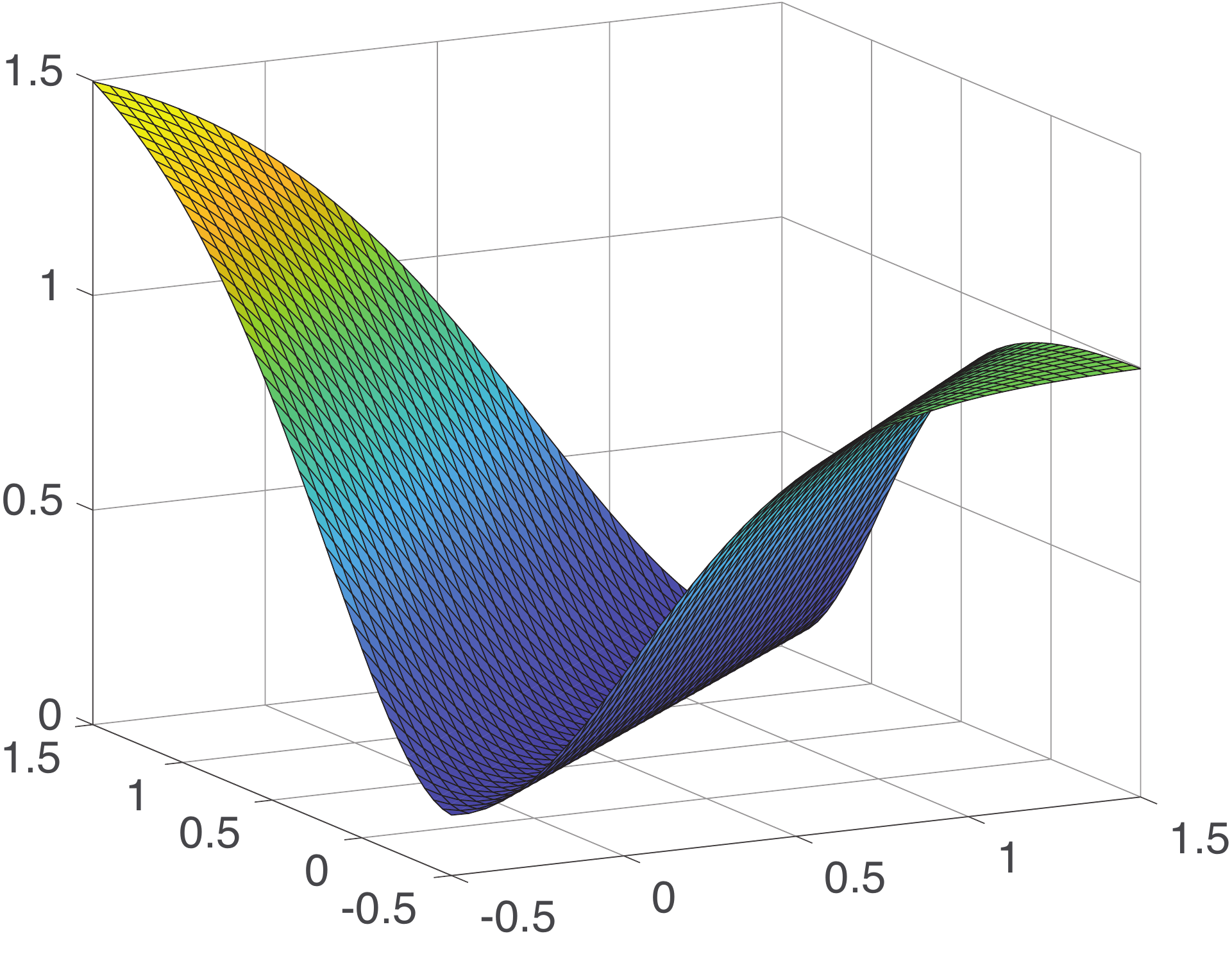, height = 2.8cm}} 
    & \mbox{\epsfig{figure= 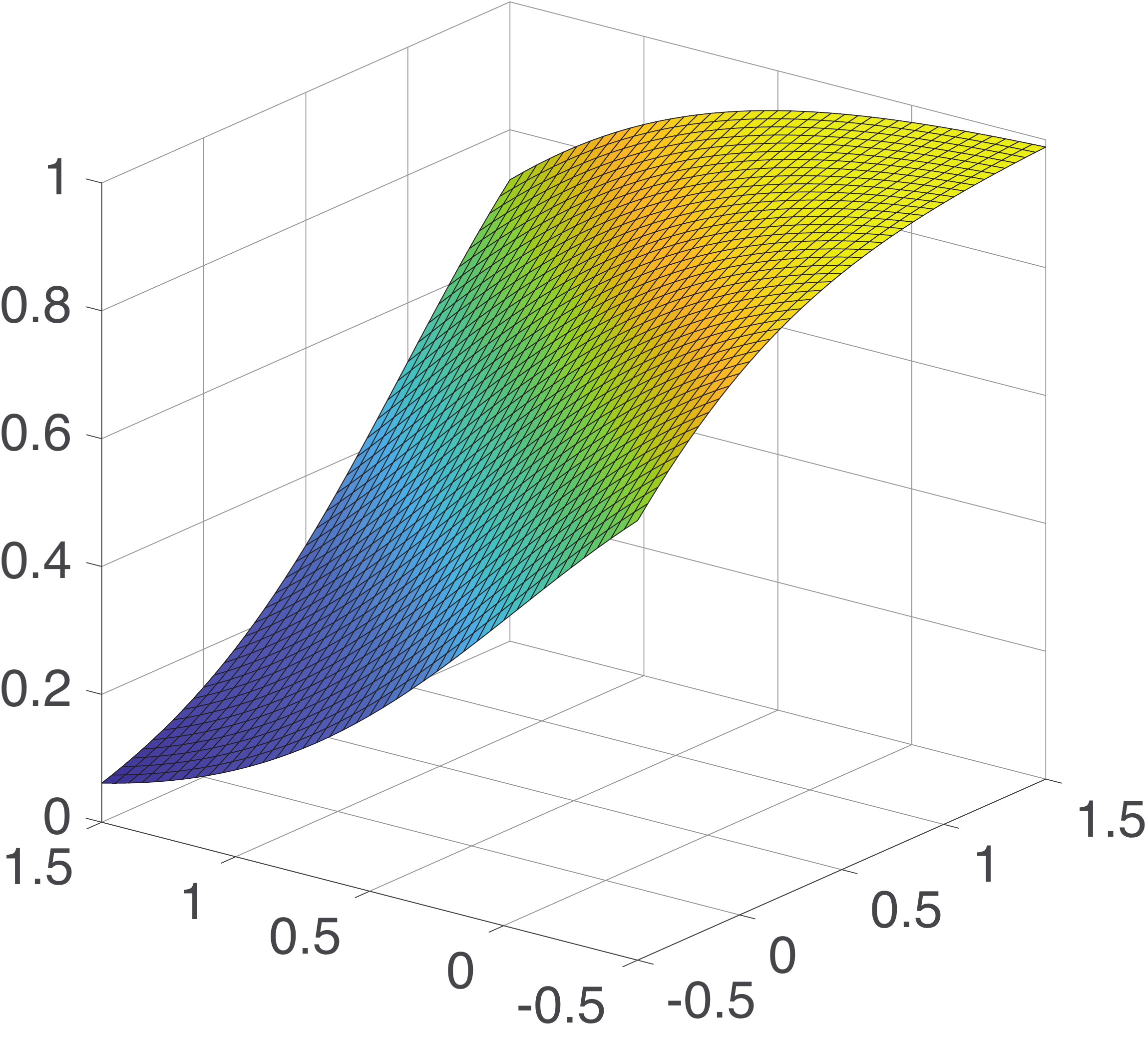, height = 2.2cm}} 
    & \mbox{\epsfig{figure= 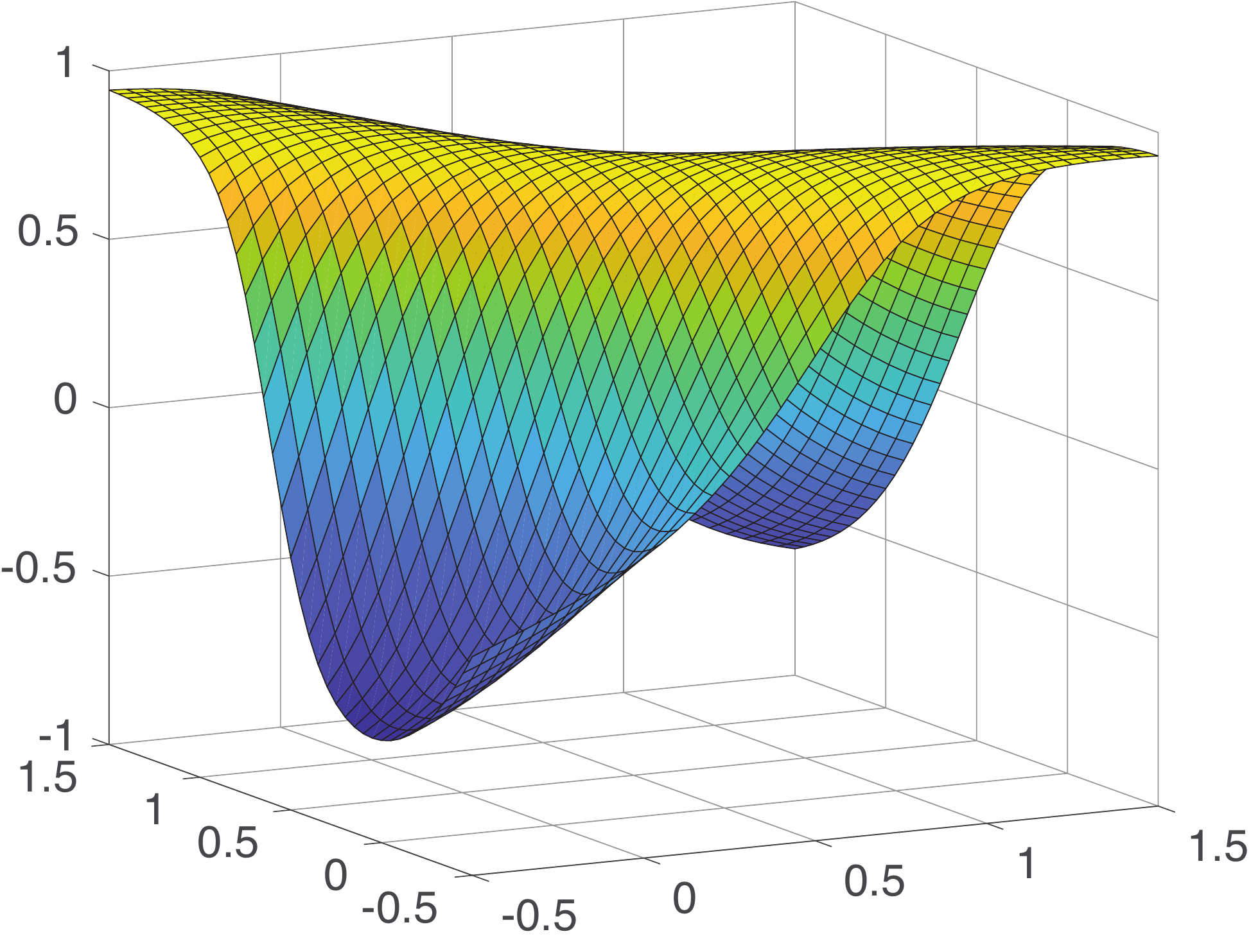, height = 2.8cm}} \\
    {\scriptsize (a) Training loss} & {\scriptsize (b) CA-NN} & {\scriptsize (c) $\chi$-function} & {\scriptsize (d) Multilayer NN, hidden size $= 3$}
\end{tabular}
\caption{(Best viewed in color) Surfaces generated by using different models to fit four point mappings: $(1, 0)\mapsto 1$, $(0, 1)\mapsto 1$, $(0, 0)\mapsto 0$, $(1, 1)\mapsto 0$. } \label{fig:cann_surface_fitting}
\end{figure}

\textbf{Synthetic data}. In this case, we simply let the model learn a mapping from $2$D to $1$D height from four training data: $(1, 0)\mapsto 1$, $(0, 1)\mapsto 1$, $(0, 0)\mapsto 0$, $(1, 1)\mapsto 0$. The loss function is defined by the L2 distance. Note that our CA-NN model allows a subnetwork to define $v(\vec{c})$ (Fig.~\ref{fig:ca_nn}).
Here we use a simple linear mapping with $\tanh$ activation. 
Fig.~\ref{fig:cann_surface_fitting} shows the comparison between CA-NN and common single/multiple layer neural networks. 
CA-NN tends to fit the data with fewer parameters (Fig.~\ref{fig:cann_surface_fitting} (a)).
Also the fitted surfaces are very different from NN models when we plot the learned network in the extended domain $[-0.5, 1.5] \times [-0.5, 1.5]$ (Fig.~\ref{fig:cann_surface_fitting} (b) and (d)).
The surface learned by CA-NN tends to be less symmetric than that learned by NN-model (adding more hidden units for the NN model would result in very similar shapes as Fig.~\ref{fig:cann_surface_fitting} (d)).
The shape of the surface fitted by CA-NN model can be explained by looking at the $\chi$-function (Fig.~\ref{fig:cann_surface_fitting} (c)), which leads the shape into two different modes.

\textbf{Smooth surface}. In another example, we test CA-NN's ability to fit a smooth surface with a small number of discrete samples.
The surface we choose can be represented in a closed form as $(x, y, xe^{-x^2 - y^2}), x, y \in[-2, 2]$ (Fig.~\ref{fig:cann_surface_fitting} (a)).
Here we discretize the surface on a $81\times 81$ grid and use $1\%$ of them for training and the rest for testing. 
For the CA-NN model, we choose the output/hidden layer size to be $dim(\vec{w_0}) = 5$ in both single and multiple layer cases, and the model is trained $1000$ steps with a learning rate of $0.1$.
Fig.~\ref{fig:cann_surface_fitting} (b) illustrates one side effect of the $\chi$-function, namely it may introduce undesired artifacts when the model is not fully learned.
This can be remedied by stacking another layer of CA-NN to it, which is able to better approximate the original shape.
The average predication error for single layer CA-NN is $0.0063$, and for multi-layer case it is $0.0029$.

\begin{figure}[t]
\centering
\begin{tabular}{ccc}
    \mbox{\epsfig{figure= 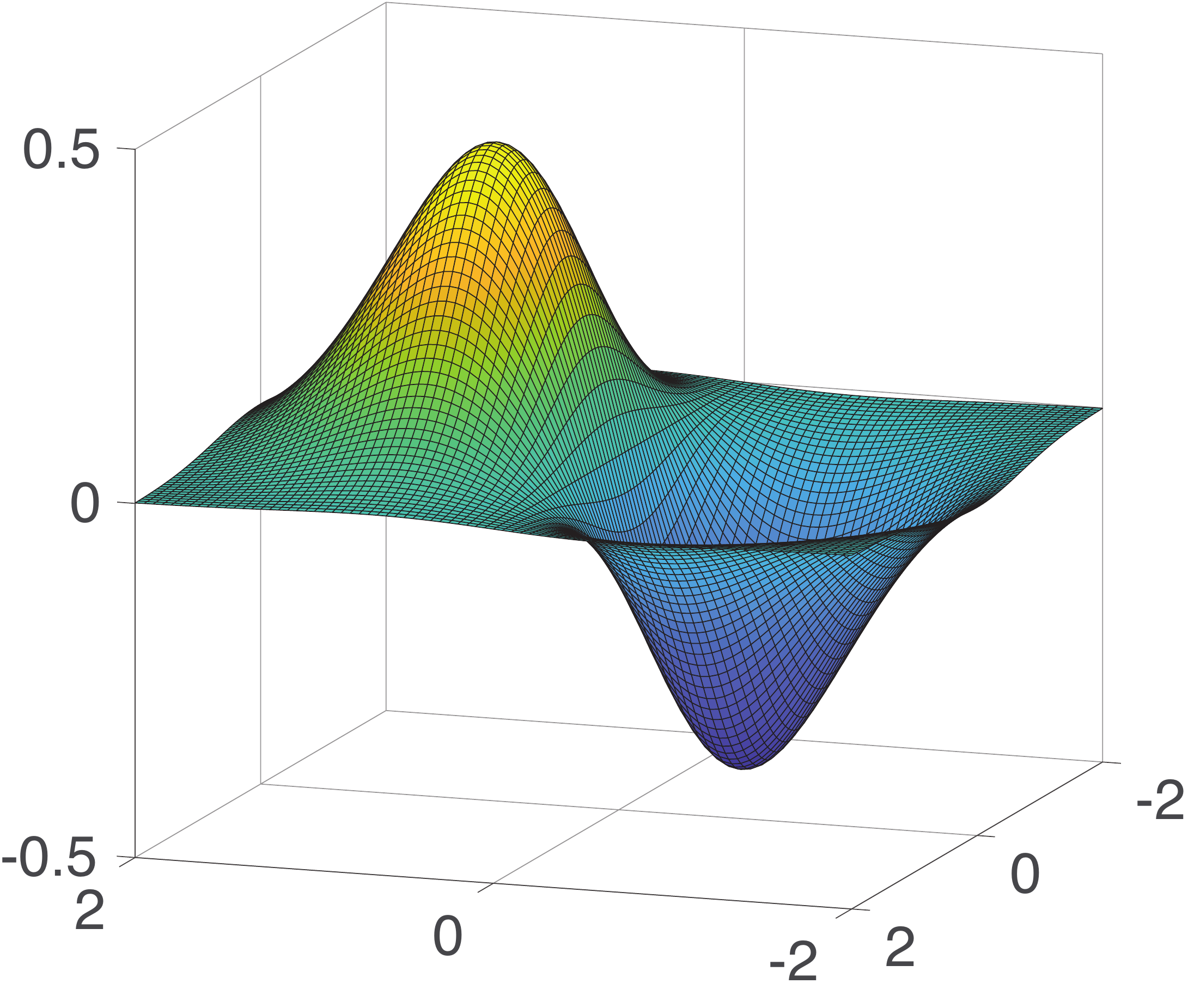, height = 3.5cm}} 
    & \mbox{\epsfig{figure= 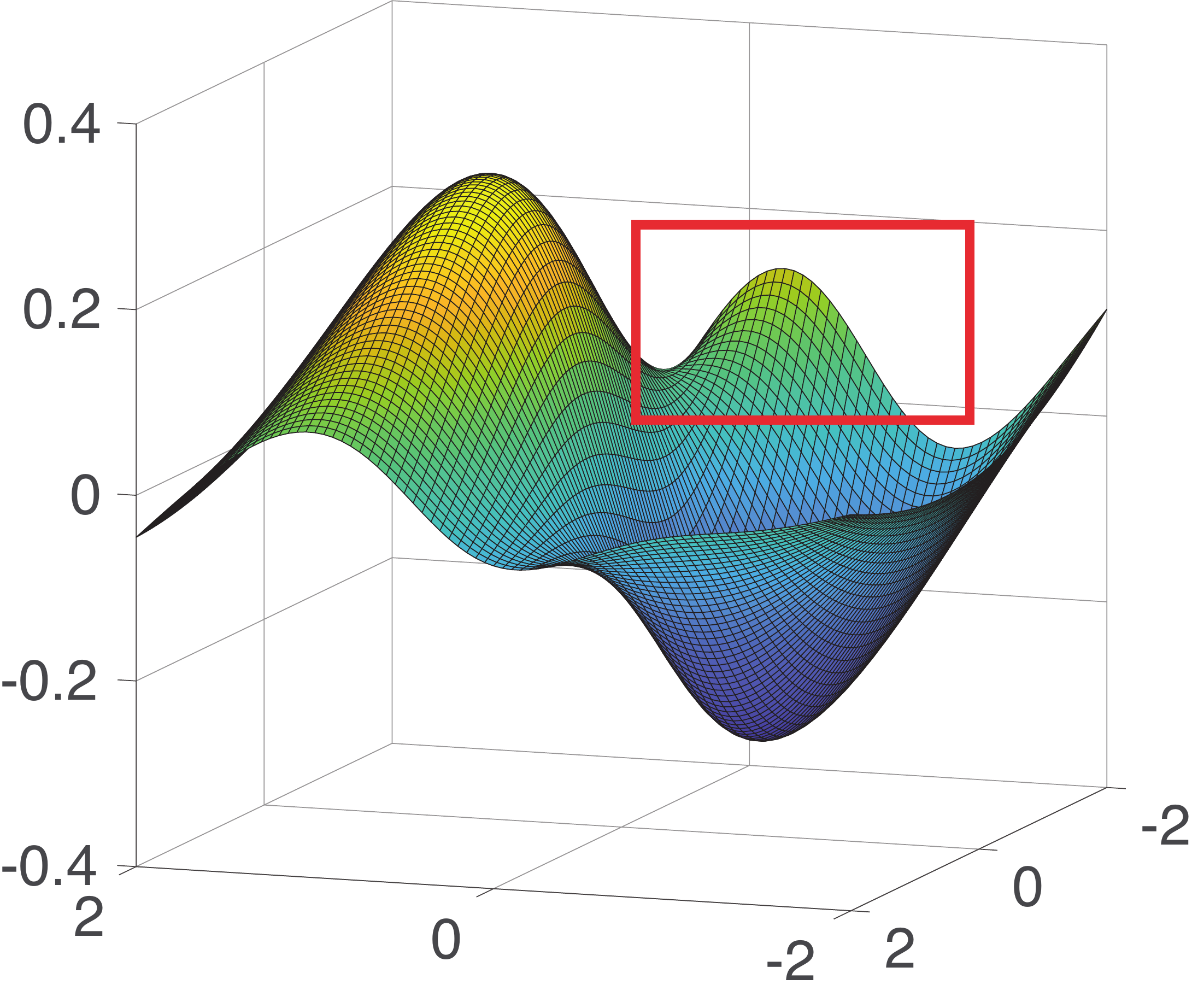, height = 3.5cm}} 
    & \mbox{\epsfig{figure= 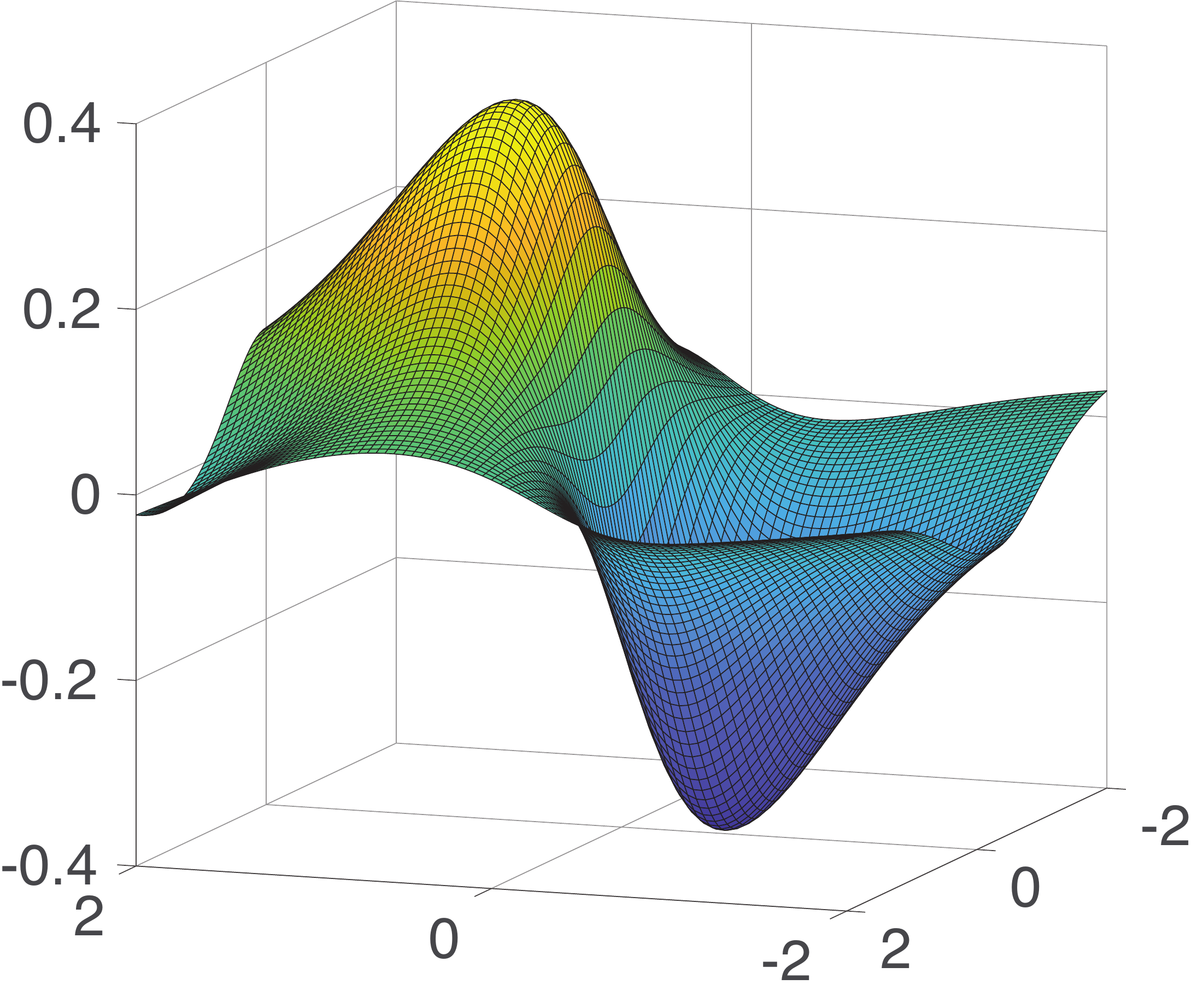, height = 3.5cm}}  \\
    {\scriptsize (a) Original} & {\scriptsize (b) Single layer CA-NN} & {\scriptsize (c)  Two layer CA-NN}
\end{tabular}
\caption{(Best viewed in color) Fitting a smooth surface (a) using a single layer CA-NN can introduce artifacts (b). Simply stacking multiple layers of CA-NN can sufficiently reduce them using the same training steps. } \label{fig:cann_smooth_surface_fitting}
\end{figure}

\subsection{Results for CA-CNN}
Here we focus only on the application of convolutional layer in image related applications.
We use the MNIST dataset to develop an intuition of the performance for a single CA-CNN layer, which can guide us build more complex model on more challenging dataset such as CIFAR~\cite{dataset_cifar}.

\subsubsection{Implementation details}
Our model is implemented by making some modifications to Tensorflow's $conv2d$ function\footnote{\url{https://www.tensorflow.org/api_docs/python/tf/nn/conv2d}}, which takes a high dimension tensor (\emph{e.g.}, an image), an integer specifying output depth and a tuple representing kernel size, etc, as input and outputs a tensor representing the result of convolution.
For example, in the $2$D case, if the kernel size is $(3, 3)$ and the depth is $128$, we can simply represent the CNN model as ``$(3, 3), 128$".
In our training, we select a batch size of $100$, a stride size of $1$, a learning rate of $0.01$ and the same Adagrad optimizer for gradient descent update.

\subsubsection{MNIST dataset}
\begin{figure}[t]
\centering
\begin{tabular}{cccc}
    \mbox{\epsfig{figure= 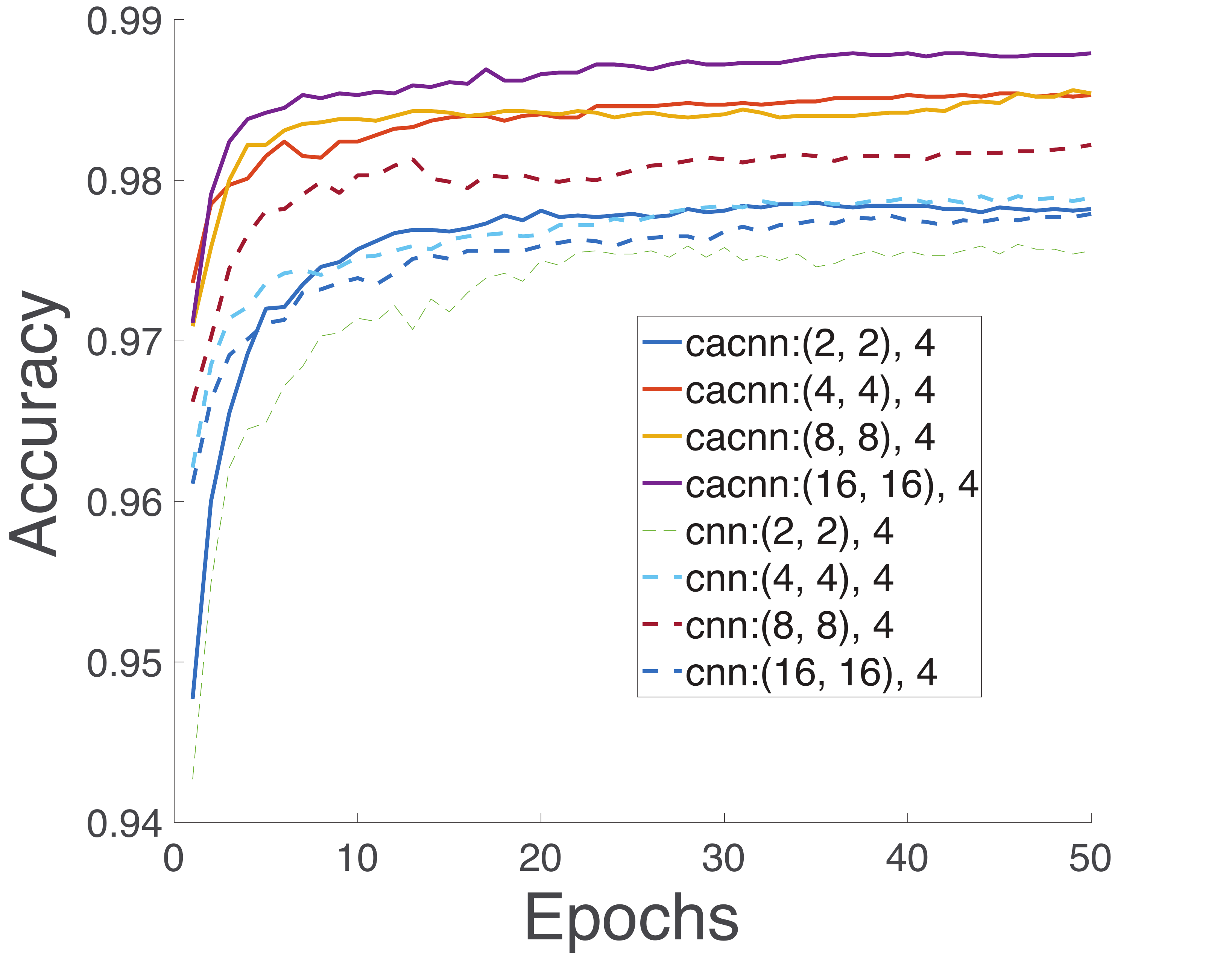, height = 3.2cm}}
    & \hspace{-7mm}\mbox{\epsfig{figure= 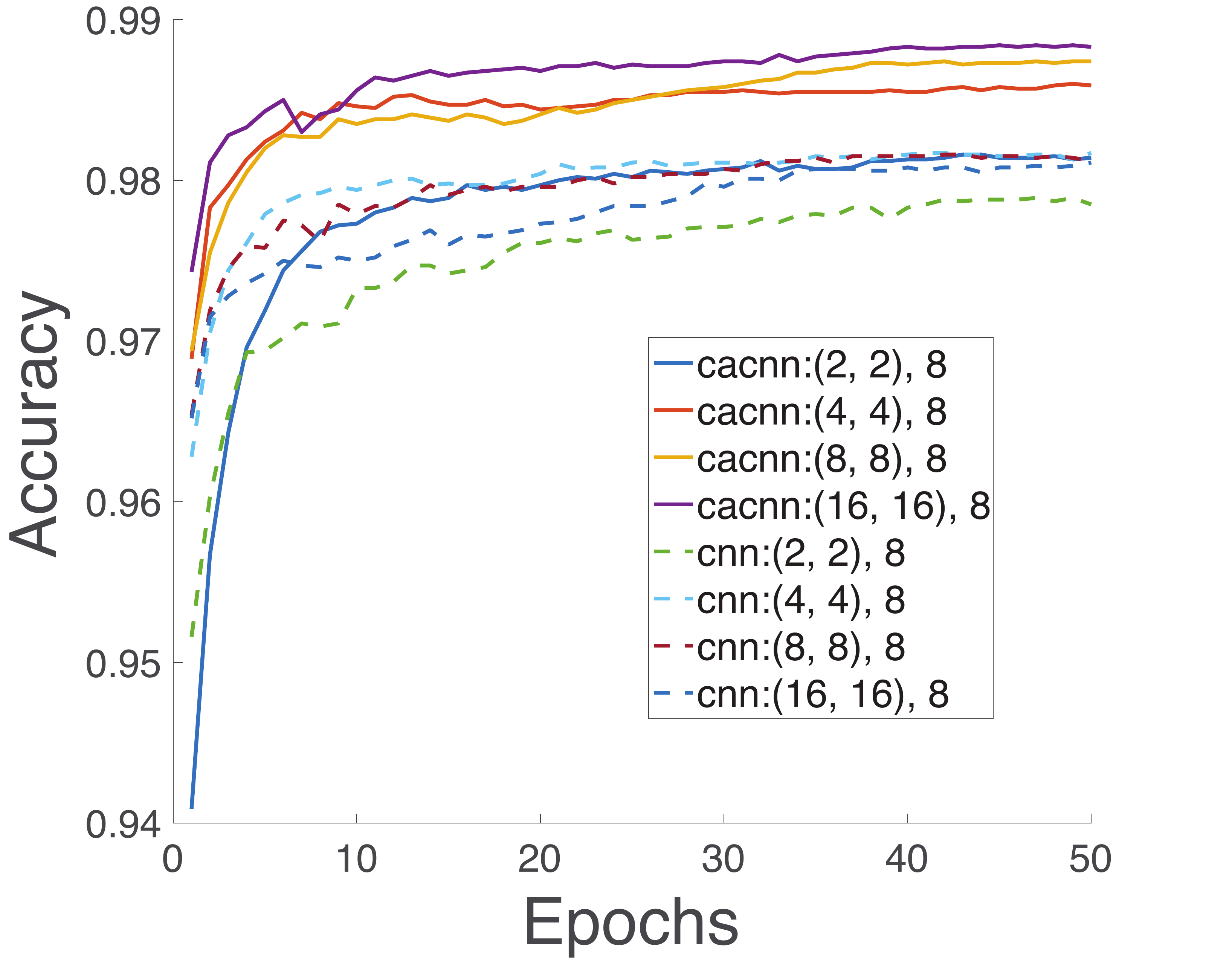, height = 3.2cm}}
    &  \hspace{-7mm}\mbox{\epsfig{figure= 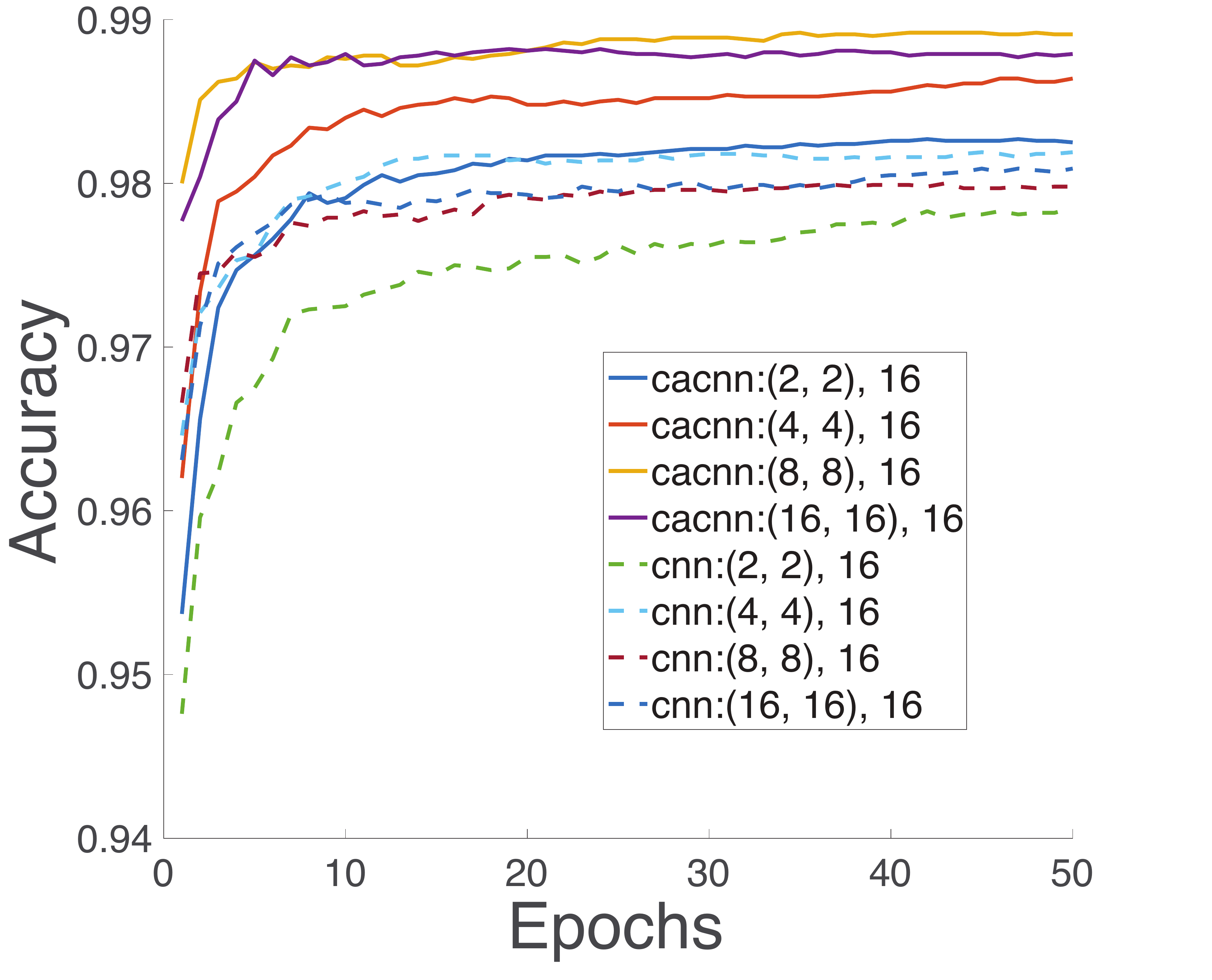, height = 3.2cm}} 
    &  \hspace{-7mm}\mbox{\epsfig{figure= 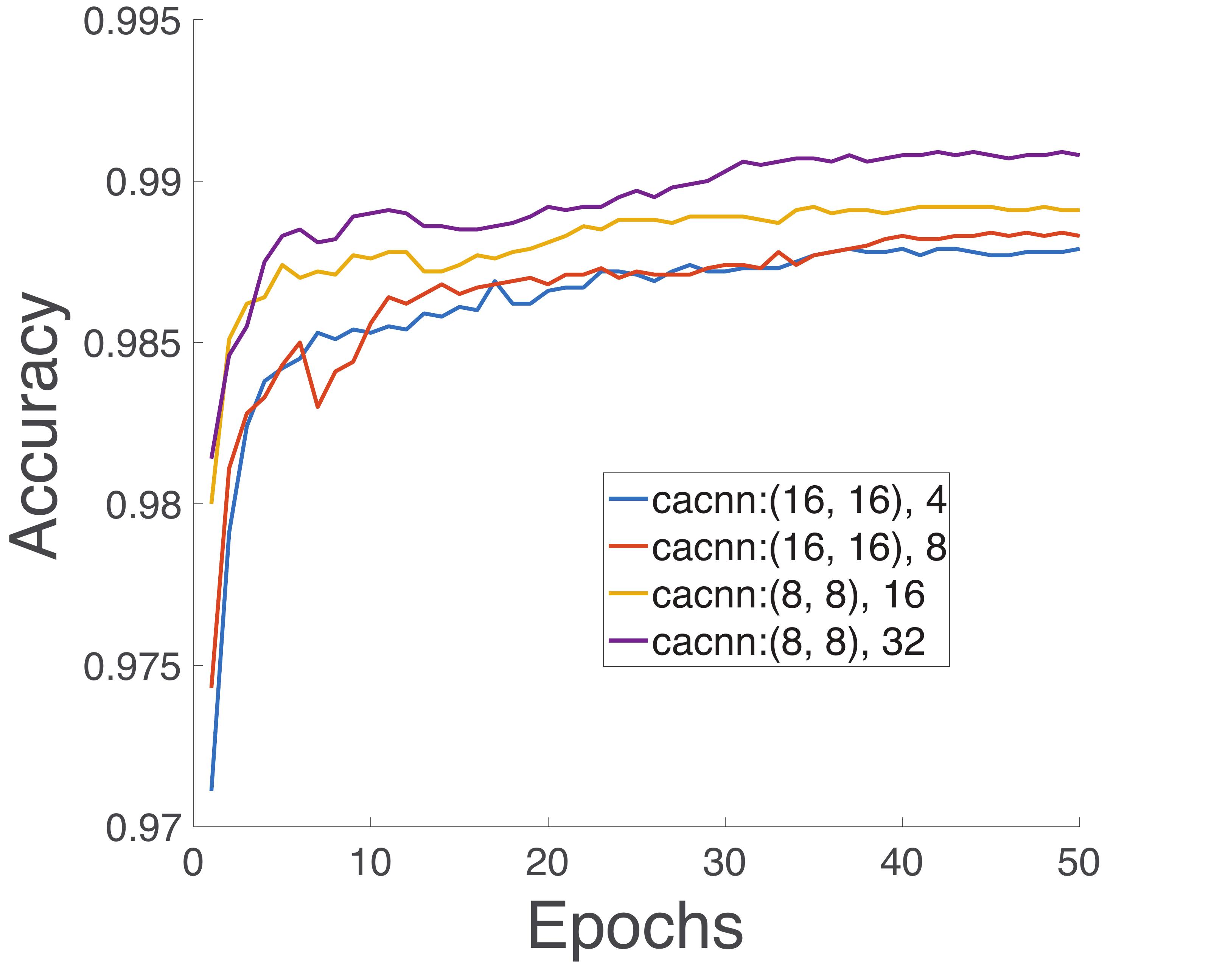, height = 3.2cm}} \\
    {\scriptsize (a) Depth = $4$.}
    & {\scriptsize (b) Depth = $8$.}
    & {\scriptsize (c) Depth = $16$.}
    & {\scriptsize (d) Best of (a), (b), (c).}
\end{tabular}
\caption{(Best viewed in color) Comparisons between CNN and CA-CNN on MINST dataset with different hyper-parameters.} \label{fig:mnist_results}
\end{figure}

For the relatively small MNIST dataset, we find a single layer CA-CNN model with one additional hidden layer (a fixed size of $128$ is used in all our experiments) to be sufficient for achieving $>99\%$ accuracy.
Compared to its CNN counterpart, the same configuration (kernel size and depth) always achieves significantly lower accuracy as shown in Fig.~\ref{fig:mnist_results}.
Fig.~\ref{fig:mnist_results} shows the influence of different kernel sizes and output depths to the accuracy, and it turns our that for CA-CNN model, a larger kernel size usually leads to a better performance.
Intuitively, a larger kernel size helps better estimate the foreground/background segmentation represented by $\chi$-function.
In contrast, CNN model performs the best with a relatively small kernel size ($(4, 4)$ or $(8, 8)$).

\begin{figure}[t]
\centering
\begin{tabular}{cc}
    \mbox{\epsfig{figure= 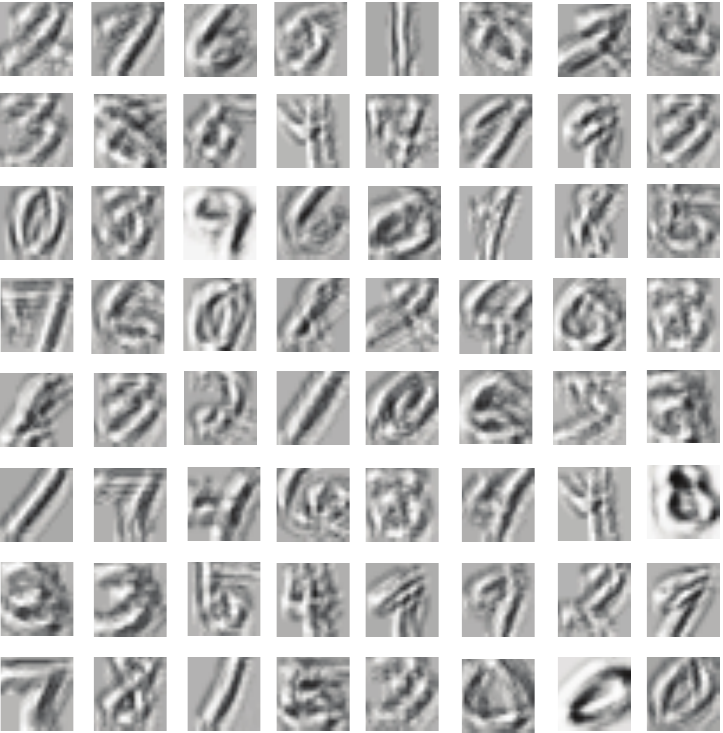, height = 4.2cm}} 
    & \hspace{8mm}\mbox{\epsfig{figure= 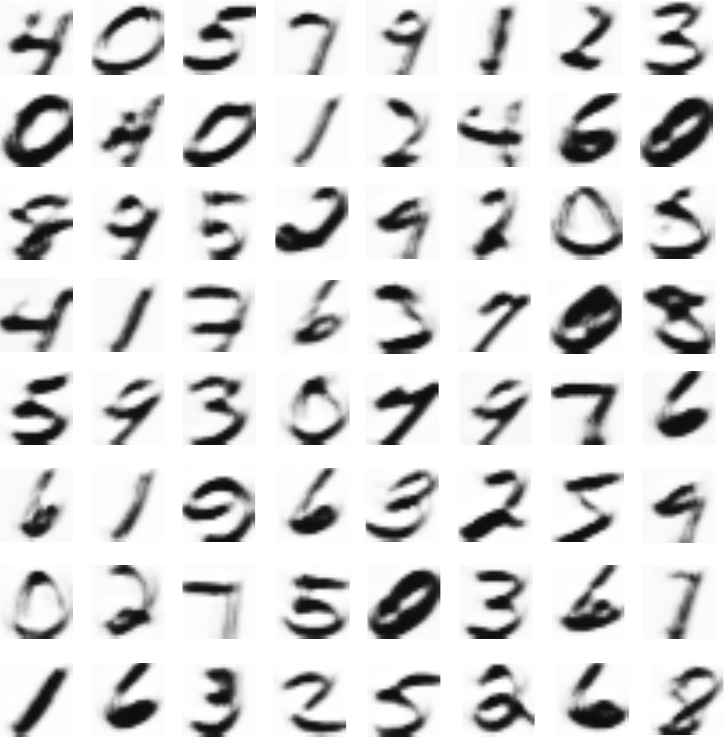, height = 4.2cm}} \\
    {\scriptsize (a) Results from initial 1000 training steps.} 
    & \hspace{8mm}{\scriptsize (b) Results after 10,000 training steps.}
\end{tabular}
\caption{The $\sigma$-map (Fig.~\ref{fig:ca_cnn}) of the $\chi$-function in Eq.~\ref{equ:sigma_function}, where $\sigma\in [0, 1]$.} \label{fig:sigma_map}
\end{figure}

Fig.~\ref{fig:sigma_map} further validates the role of $\chi$-function in helping divide relevant/irrelevant information from the input.
Here values of the $\chi$-function for every pixel are visualized as grey-scale images, or also called $\sigma$-maps.
Samples of these $\sigma$-maps from different training steps are shown in Fig.~\ref{fig:sigma_map} (a) and (b).
After $10,000$ training steps, it can be seen that the foreground and background are very well separated (Fig.~\ref{fig:sigma_map} (b)).

\subsubsection{CIFAR-10 dataset}\label{subsubsec:cacnn_cifar}
\begin{figure}[t]
\centering
\begin{tabular}{ccc}
    \mbox{\epsfig{figure= 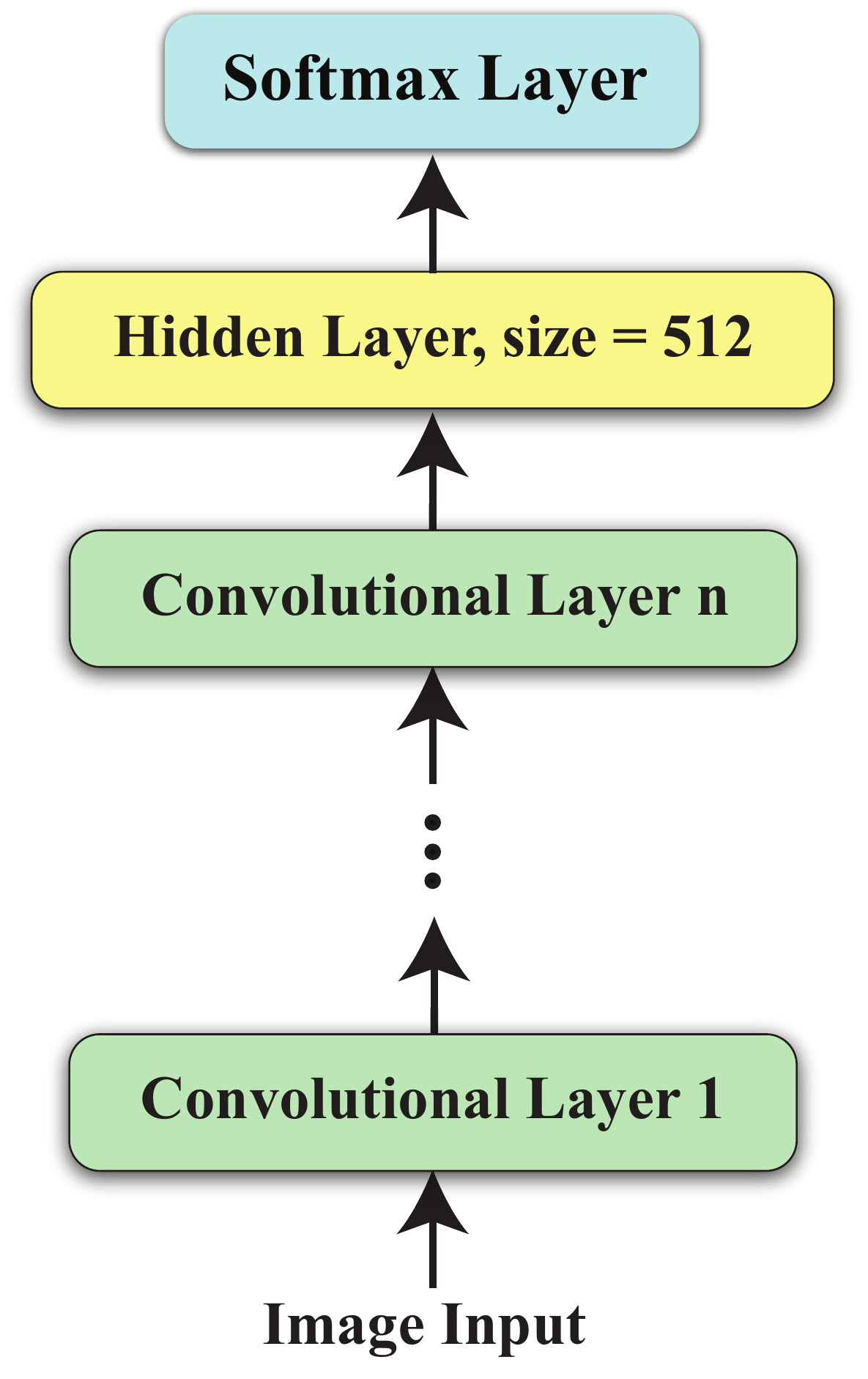, height = 4.2cm}}  
    & \mbox{\epsfig{figure= 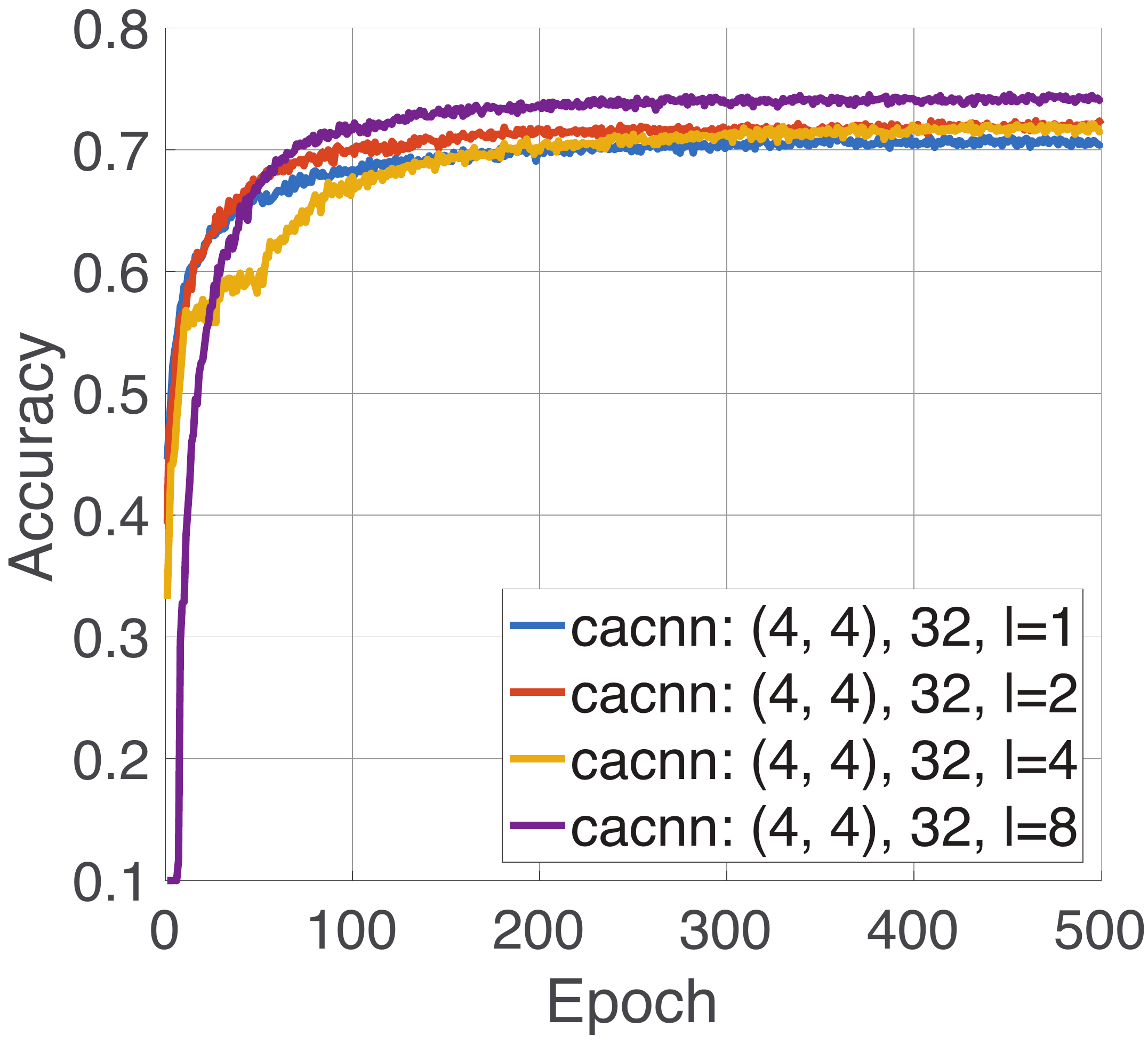, height = 4.2cm}} 
    & \mbox{\epsfig{figure= 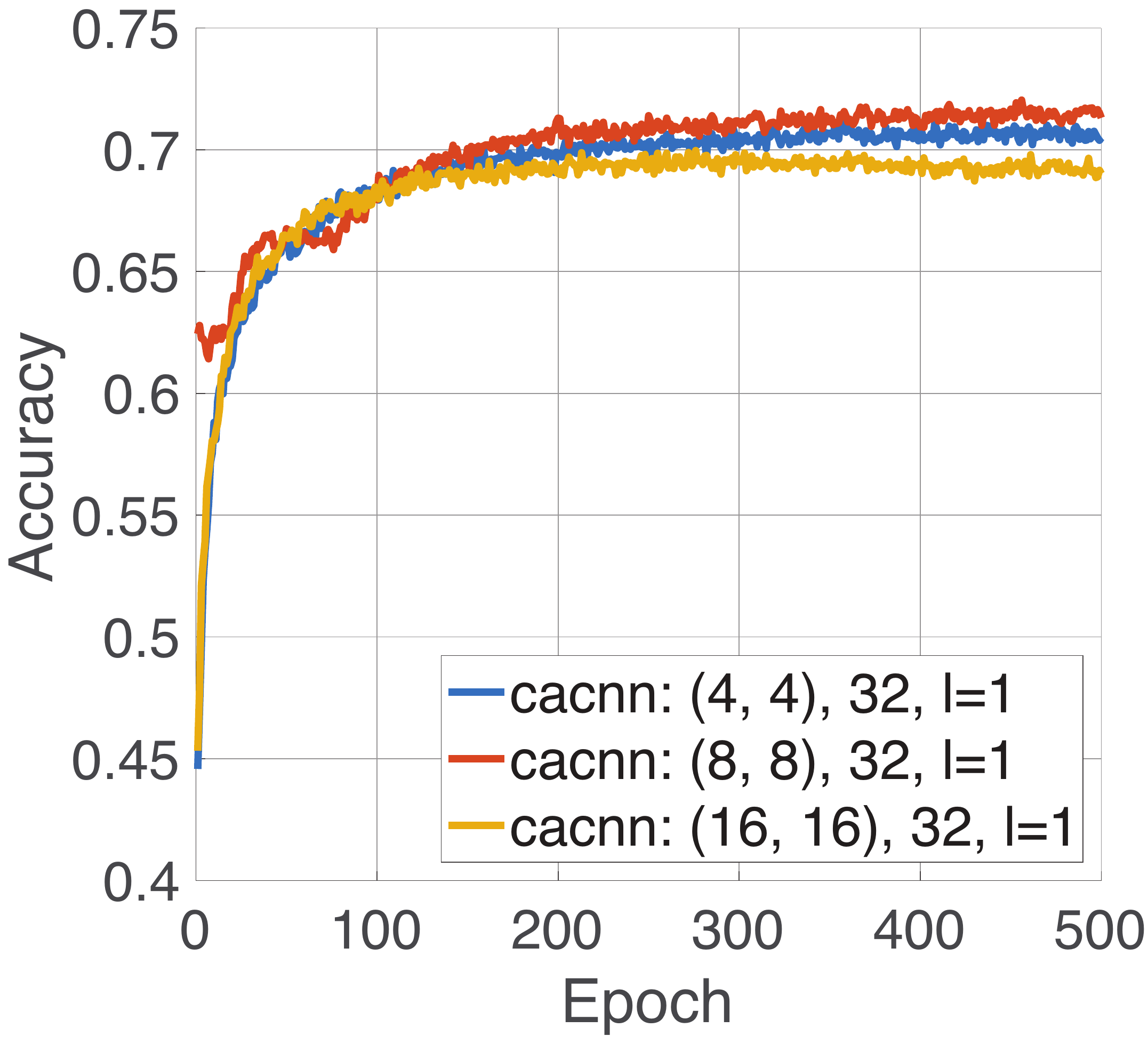, height = 4.2cm}}  \\
    {\scriptsize (a) Model architecture.} & {\scriptsize (b) Layer number $=1, 2, 4, 8$.} & {\scriptsize (c) Kernel size $=4, 8, 16$.}
\end{tabular}
\caption{(Best viewed in color) Model architecture (a) and some initial results from various configurations ((b) and (c)).} \label{fig:cacnn_cifar_result}
\end{figure}

For the CIFAR dataset, we employ a very simple model architecture to test the performance of different configurations: let the input image first go through multiple layers of convolutional networks, then feed the result to a hidden layer of size $512$ before the final softmax output (Fig.~\ref{fig:cacnn_cifar_result}(a)).
Here the convolutional network layer can be a CNN, a CA-CNN, or a CA-RES (to be discussed in the next section).

Fig.~\ref{fig:cacnn_cifar_result} shows some initial results on running CA-CNN with different configurations.
From Fig.~\ref{fig:cacnn_cifar_result}(b) it can be seen that increasing CA-CNN layer sizes can help improve the accuracy.
But when the layer size reaches $16$, the model would not learn anything without other strategies like the block-coordinate gradient update scheme mentioned in Sec.~\ref{subsubsec:cabsfe_impl}.
It is also consistent with the MNIST results that for CA-CNN, a moderate kernel size (\emph{e.g.}, $(8, 8)$) works better than smaller or larger ones (Fig.~\ref{fig:cacnn_cifar_result}(c)).

\subsection{Results for CA-RES}
Our CA-RES model can be used as a replacement to a regular deep network with multiple hidden layers, or a deep residual network.
Our design is flexible enough to subsume them as our special cases, while providing a lot more possible new architectures controlled by a few hyper-parameters.
In this section, we only demonstrate a few aspects of our model and leave it to future work to explore its many more properties.

\subsubsection{Implementation details}
The CA-RES model is implemented in a way that its complexity can be controlled by a tuple $(l, n_v, n_{\sigma}, b)$ where $l, n_v, n_{\sigma} \in \mathbb{N}^+, b\in\{0, 1\}$.
Here $l$ is the total number of layers; $n_v$ specifies the maximal number of previous layers whose activation $\vec{w}_i$ are used for computing $\vec{v}_l(\cdot)$ in Eq.~\ref{equ:cares_decomposition}, \emph{i.e.}, $\vec{v}_l(\vec{c}, \vec{w}_{l-1}, \ldots, \vec{w}_{l-n_v})$;
similarly, $n_{\sigma}$ is used for computing $\chi(\cdot)$ in Eq.~\ref{equ:cares_decomposition}, \emph{i.e.}, $\chi(\vec{c}, \vec{w}_{l-1}, \ldots, \vec{w}_{l-n_{\sigma}})$;
$b$ indicates if $\chi(\cdot)$ should always be on ($b=1$), therefore bypassing $\vec{w_{l0}}$.
This way, it is easy to see that a configuration of $(2, 1, \_, 1)$\footnote{When $b=1$, the value of $n_{\sigma}$ can be ignored as $\chi(\cdot)\equiv 1$.} with $\vec{v}_2(\vec{c}, \vec{w}_1) = \vec{c} + \vec{w}_1$ and $\vec{w}_1 = \vec{v}_1(\vec{c})$ where $\vec{v}_1$ is any activation function, would reduce our model to the residual network~\cite{resnet_2015}.
To use CA-RES in a deep neural network, one can either specify a large number of $l$, or stack multiple blocks of CA-RES layers.
For gradient update, the block-coordinate gradient update scheme proposed in Sec.~\ref{subsubsec:cabsfe_impl} can also be employed to improve the accuracy.

\subsubsection{CIFAR-10 dataset}
Here we use the same model mentioned in Sec.~\ref{subsubsec:cacnn_cifar}, with the difference that each convolutional layer is a CA-RES network that may contain multiple sublayers (\emph{e.g.}, Fig.~\ref{fig:ca_nn}(b) and (c)).
Therefore in addition to testing the impacts of layer number and kernel size to the final accuracy, we also should test the effect of sublayers in our CA-RES model. 

\begin{figure}[t]
\centering
\begin{tabular}{ccc}
    \mbox{\epsfig{figure= 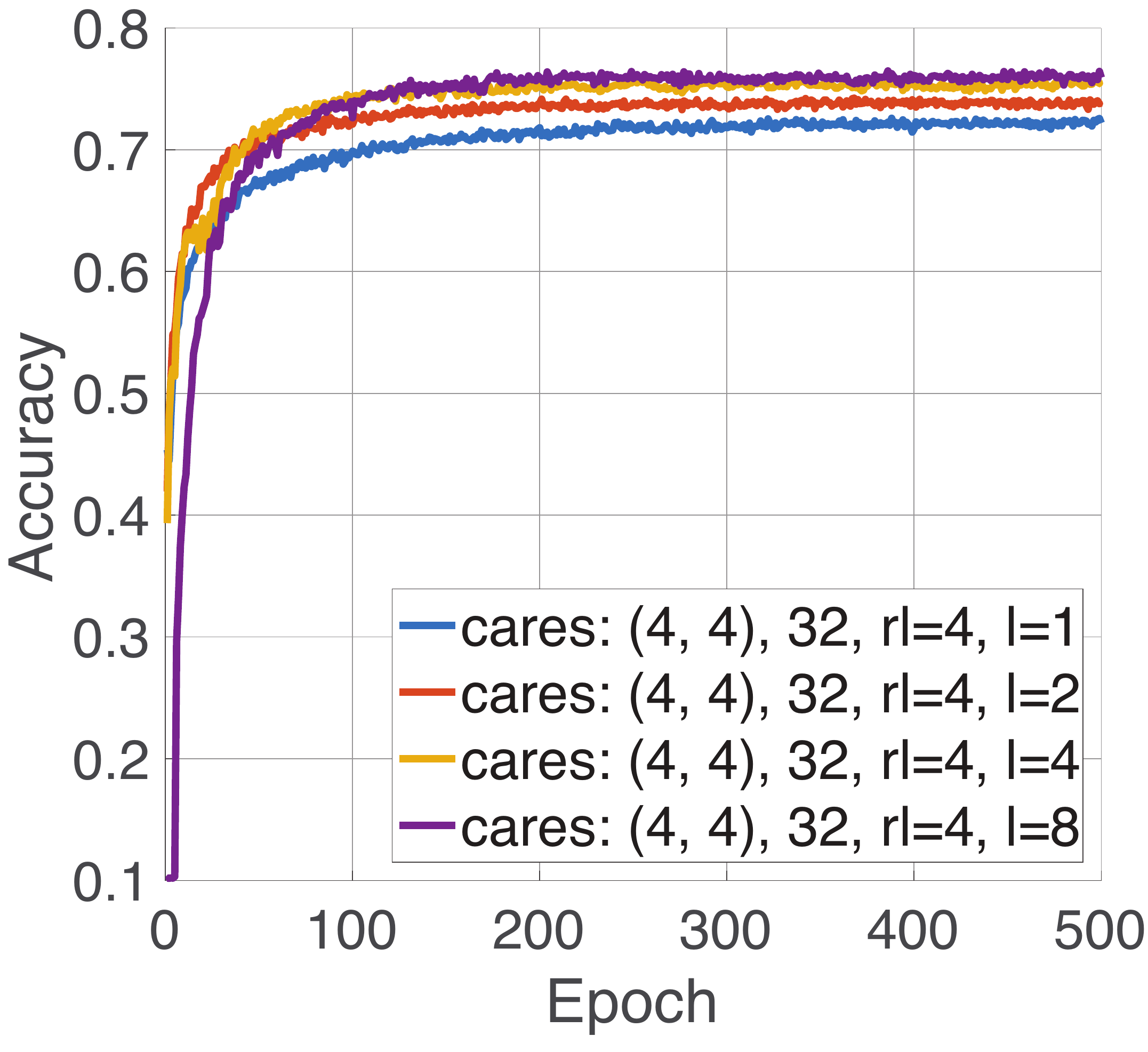, height = 4.2cm}}  
    & \mbox{\epsfig{figure= 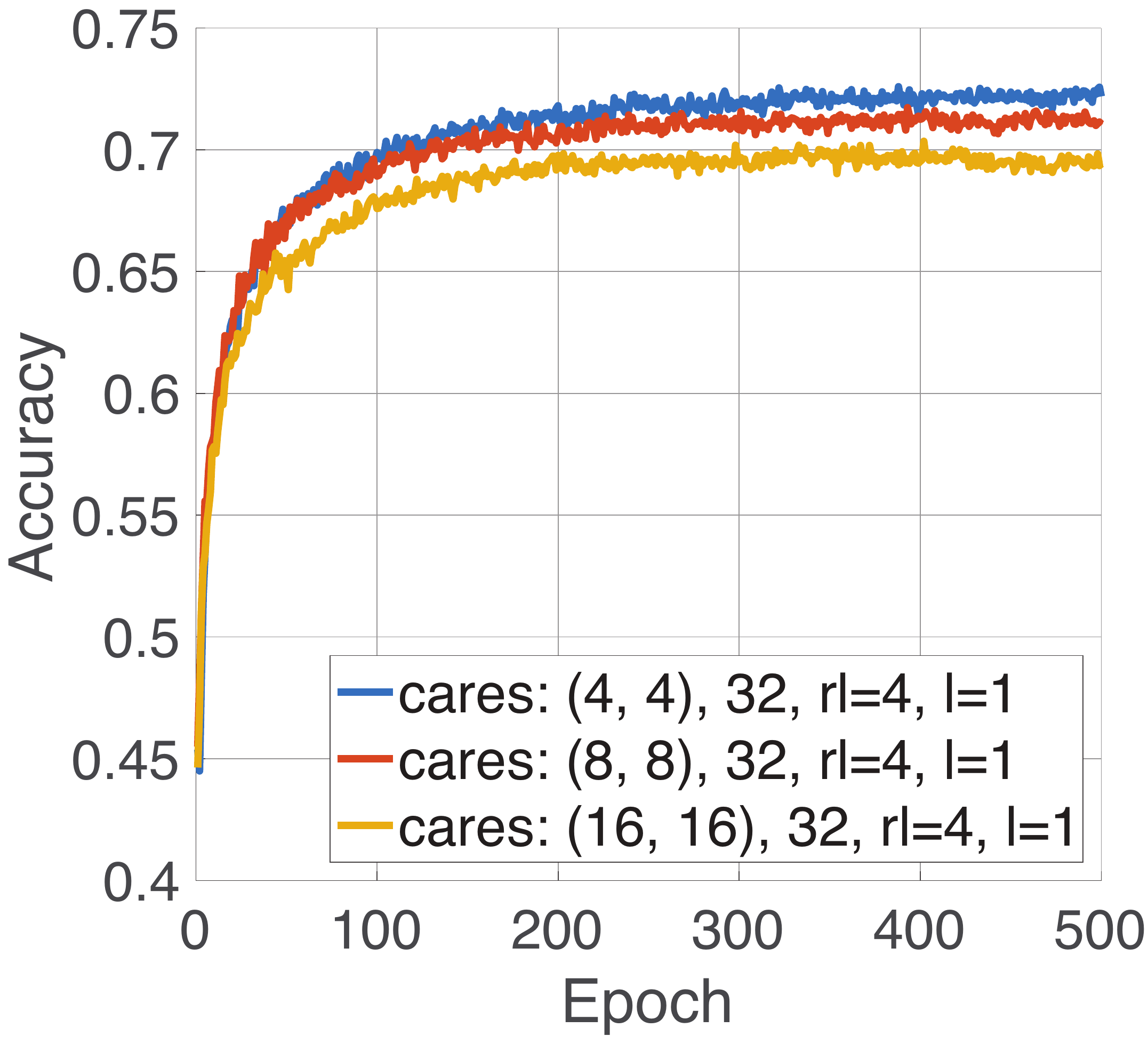, height = 4.2cm}} 
    & \mbox{\epsfig{figure= 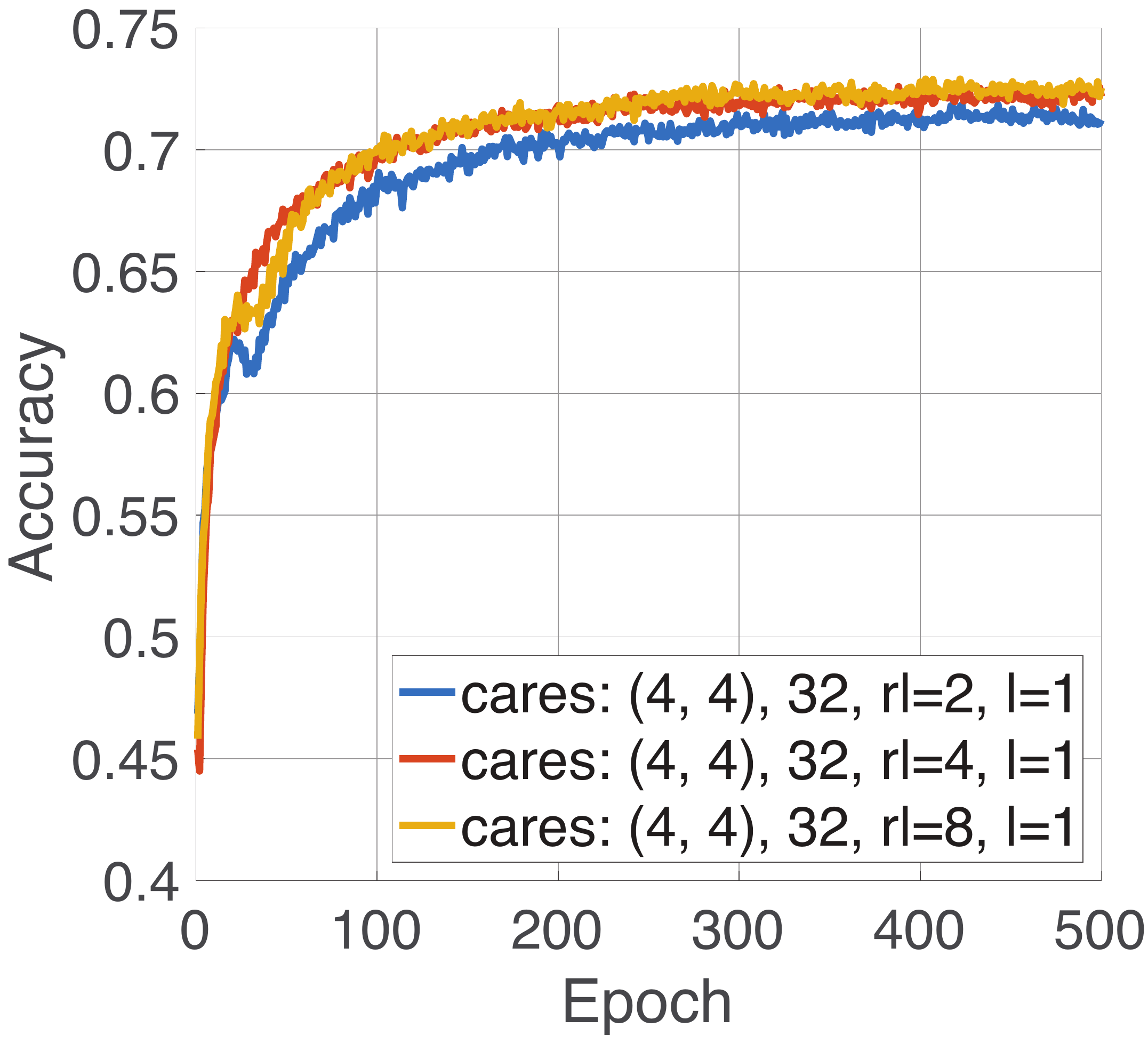, height = 4.2cm}}  \\
    {\scriptsize (a) Layer number $=1, 2, 4, 8$.} & {\scriptsize (b) Kernel size $=4, 8, 16$.} & {\scriptsize (c)Sublayer size $= 2, 4, 8$. }
\end{tabular}
\caption{(Best viewed in color) Results for various configurations of using CA-RES in a multilayer network for CIFAR-10 dataset. } \label{fig:cares_cifar_result}
\end{figure}

Similar to our CA-CNN model, more layers always leads to higher accuracy as shown in Fig.~\ref{fig:cares_cifar_result}(a).
However, unlike CA-CNN, a bigger kernel size does not help in improving the accuracy (Fig.~\ref{fig:cares_cifar_result}(b)).
Unsurprisingly, more number of sublayers in each CA-RES layer also drives better results (Fig.~\ref{fig:cares_cifar_result}(c)).  

\section{Related work}
\label{section:related_work}
Loosely speaking, existing supervised machine learning models can be categorized based on the type of input they take.
Sparse models (\emph{e.g.}, CA-SEM and CA-BSFE) handle input with discrete values such as text or categorical features, etc.
Sequence models (\emph{e.g.}, CA-ATT and CA-RNN) handle input with an order and also varying in length.
Image models (\emph{e.g.}, CA-RES and CA-RNN) take images as input that are usually represented by $n\times m$ array.
Such a taxonomy is only helpful for us to review related works; they are not necessarily exclusive to each other.
Because our new framework provides a unified view of many of these models, at the end of this section, we also review related efforts in building theoretical foundations for deep learning.

\subsection{Sparse Models}
While a single sparse feature can be represented as a vector, there is no consensus on how to represent a bag of features in the embedding space.
Although not many work has been done on the general problem of representing a bag of sparse features using a single vector~\citep{sem_mitchell2010composition,sem_mutilinual14},  its special case, phrase/sentence embedding in language modeling, has already drawn increasing attention due to the fast development of distributed representation using neural networks~\citep{emb_bengio2003neural,emb_mikolov2013word2vec,emb_pennington2014glove,emb_arora2016latent}. Existing approaches either fall into bag of words (BOW) based, or sequence based. 
Extensive comparative studies have shown that neither approach has obvious advantage over the other and their performance highly depends on the specific metric~\citep{sem_wieting2015towards,sem_hill2016learning}.  

Among bag of words based models, early attempts including~\cite{sem_le2014distributed} directly extended the idea of Word2Vec~\citep{emb_mikolov2013word2vec} by treating each sentence as a whole with a unique feature id. Later, \cite{sem_hill2016learning} showed that a more flexible approach, which simply computes the sentence embedding by summing up the embeddings of individual words, would achieve significant improvement. \cite{sem_wieting2015towards} also showed that starting from an existing word embedding model, training them towards a specific task, then simply using the average of new word embeddings in a sentence would achieve satisfactory results. Recently, ~\cite{sem_arora2017} proposed a TF-IDF like weighting scheme to composite the embedding of words in a sentence and showed that the performance can be improved simply by removing the first principle component of all the sentence vectors. 
More recently, the transformer architecture using self-attention proposed by~\cite{attention_all_you_need} provides further evidences that bag of words model can be more preferable than sequence models, at least in certain tasks.

As for sequence-based approaches, ~\cite{sem_kiros2015skipthought} proposed SkipThought algorithm that employs a sequence based model to encode a sentence and let it decode its pre and post sentences during training. By including attention mechanism in a sequence-to-sequence model~\citep{sem_lin2017structured}, a better weighting scheme is achieved in compositing the final embedding of a sentence. Besides attention, ~\cite{sem_wang2016cse} also considered concept (context) as input, and~\cite{sem_conneau2017supervised} included convolutional and max-pool layer in sentence embedding. Our model includes both attention and context as its ingredients.

In parallel to neural network based models, another popular approach to feature embedding is based on matrix factorization~\citep{cf_koren2015advances}, where a column represents a feature (\emph{e.g.}, a movie) and each row represents a cooccurrence of features (\emph{e.g.}, list of movies watched by a user). In this setting, one can compute the embeddings for both the features and cooccurrences, and the embedding for a new cooccurring case (bag of features) can be computed in a closed form~\citep{cf_hu2008collaborative}. 
Recently, the concept of cooccurrence is generalized to n-grams by reformulating the embeddings of bag-of-features as a compressed sensing problem~\cite{sem_arora2018}.
Nevertheless, these methods are mostly limited to linear systems with closed-form solutions. 

\subsection{Sequence Models}
Modeling sequential data (\emph{e.g.}, sentences or sound) requires inferring current output from both past observation and current input.
Recurrent neural network (RNN) assumes past observations can be summarized in its hidden state, which can then be learned recurrently (recursive in time).
Despite its promising modeling power, its basic form is proven to be difficult to train~\cite{seq_rnnlearn}.
Hence various improvements are proposed based on different assumptions (\emph{e.g.}, ~\cite{seq_rnndifficult}), among which the LSTM model~\cite{lstm_original} might be among the most successful one that are widely used.
Encouraged by its success, various models are proposed to modify its design but very few are game-changing~\cite{lstm_space_odyssey}.
One of the very few successful challengers of LSTM is the GRU model~\cite{seq_gru}, which is proven to perform better than LSTM in many cases. 

More recently, rather than improving the basic RNN-like structures, it was found that changing how they are stacked and connected can help solve very challenging problems like language translation.
The sequence-to-sequence model~\cite{seq_seq2seq} is one of such successful stories.
Then the attention mechanism was proposed able to significantly improve the performance~\cite{attention_nmt14}.
Recently, it was practically validated that even the basic RNN structure can be rid of by using attention mechanism alone~\cite{attention_all_you_need}.

\subsection{Image Models}
The visual pathway is one of the most well understood components in human brain, and the convolutional neural network model~\cite{cnn_nips1990} has been proven to model its hierarchical, multi-layer processing functionality very closely.
A series of upgrades have been added to the basic CNN structure to improve its performance, such as pooling~\cite{cnn_maxpooling}, dropout~\cite{cnn_dropout}, or normalization~\cite{batch_normalization,layer_normalization}.
Recently, the residual network model~\cite{resnet_2015} and its variants (\emph{e.g.}, ~\cite{resnet_wideres,image_training_very_deep_networks}) becomes very popular due to their superior performance. 
Among them, the highway networks~\cite{image_training_very_deep_networks} shares certain similarity with our CA-NN model in its use of a gate (maps to our $\chi$-function) to switch between an identity activation (we use a default value $\vec{w_0}$ instead) and a nonlinear activation (maps to $v(\vec{c})$).
Each year, combinations of these building blocks are proposed to reach higher accuracies on standard benchmarks for image classification (\emph{e.g.}, \cite{cnn_imagenet,cnn_lenet15}) or other tasks.

However, human vision is also known to be sensitive to changes in time and attend only to contents that have evolutionary value.
\cite{cnn_visualattention} proposed a reinforcement learning model to explain the dynamic process of fixation in human vision.
\cite{cnn_attention15} directly generalizes the text attention model in the encoder-decoder structure~\cite{attention_nmt14} to image caption generation.
\cite{cnn_attention18} proposed a mutilayer deep architecture to predict the saliency of each pixel. 
Note that our CA-CNN architecture also allows explicit computation of saliency maps for input images.

\subsection{Theoretical Foundations for Deep Learning}
Initially, a biological neuron is characterized by its binary active/inactive states~\cite{theory_neuronmodel} and hence inspired the perceptron model~\cite{theory_perceptron}.
This was then extended to multiple neurons that are divided into visible or hidden, whose states can be modeled by a Boltzmann machine~\cite{theory_bm}, or its simplified version, \emph{i.e.}, restricted Boltzmann machine (RBM)~\cite{theory_rbm}. 
Later on, this framework was further extended to the exponential family~\cite{exponential_family_harmoniums}\footnote{Again, its major difference with our use of exponential family is that we do not deal with the number of visible/hidden units and instead consider a single layer as a variable, whereas the work of~\cite{exponential_family_harmoniums} assumes the sufficient statistics (embedding) is applied to each visible/hidden unit. }.
Despite their plausibility in explaining certain brain functions (\emph{e.g.}, ~\cite{theory_wakesleep}), and deep connections to statistical physics and graphical models, these models are often inefficient to learn and limited in solving large scale problems compared to those heuristically defined, back propagation based neural network models discussed in this paper.

Connecting data-driven machine learning problems to physics-driven geometric approaches can help us understand the problem from a different angle.
Intuitively, geometry concerns the shape of the data (topology + metrics), which translates to the embeddings of the data and their distance measures~\cite{thesis_wainwright,thesis_geometryML}.
Another direction is to study the process of optimization (gradient descent) as a continuous, dynamical system such that one can easily analyze its optimality properties (\emph{e.g.}, ~\cite{theory_meanfieldview,theory_optimaltransport}). 
These analyses are often based on simplified assumptions of the energy or network structure and therefore may not scale well if the complexity of the system increases.

\section{Conclusion}
\label{section:conclusion}

We demonstrated that it is beneficial both in theory and in practice to study neural network models using probability theory with exponential family assumption -- it helps us better understand existing models and provides us with the principle to design new architectures.
The simple yet fundamental concept of context-awareness enables us to find a mathematical ground for (re)designing various neural network models, based on an equally simple EDF.
Its intrinsic connections to many existing models indicates that EDF reveals some \emph{natural} properties of neural networks that is universal.
It is foreseeable that many new models would be proposed to challenge the status-quo, based on principles developed in this paper, as we plan to do in the near future.




\newpage








\vskip 0.2in
\bibliography{sem}

\end{document}